\documentclass[final,a4paper,12pt,english]{UnicaPhdThesis-updated}

\usepackage[nottoc]{tocbibind}

\usepackage{amsmath} 
\usepackage{graphicx}

\usepackage[nomargin,inline,index]{fixme} 
\fxusetheme{color}
\usepackage{ifdraft}                
\usepackage{fontspec}

\usepackage{hyperref} 
\usepackage{tabularx}
\usepackage{multirow}
\usepackage{booktabs}
\usepackage{enumitem}
\usepackage[title]{appendix}
\usepackage[most]{tcolorbox}
\usepackage{xcolor}
\usepackage{subcaption}
\usepackage{amsmath} 
\usepackage{amssymb} 
\definecolor{mediumgreen}{RGB}{168, 208, 141} %
\definecolor{darkgreen}{RGB}{34, 139, 34} %
\hypersetup{final} 
\newcounter{example}
\newenvironment{example}{%
  \refstepcounter{example}%
  \noindent(\theexample)\,%
}{}

\title{Methods for Knowledge Graph Construction from Text Collections: Development and Applications}

\author{Vanni Zavarella
}

\supervisor{ Prof. Gianni Fenu}
\cosupervisor{Prof. Diego Reforgiato Recupero}
\coordinator{Prof. Antonio Iannizzotto}

\phd{Computer Science}

\ssd{INF/01}

\cycle{XXXVIII}

\date{Final exam. Academic year 2024/2025 \\ 
Thesis defense session: February 2026}

\clearpage
\begin{document}

\maketitle

\frontmatter
\pagenumbering{roman}

\chapter*{Statement of Authorship}

I declare that this thesis entitled “Methods for Knowledge Graph Construction from Text Collections: Development and Applications” and the work presented in it are my own. I confirm that:
\begin{itemize}
    \item this work was done while in candidature for this PhD degree;
\item when I consulted the work published by others, this is always clearly attributed;
\item when I quoted the work of others, the source is always given;
\item I have acknowledged all main sources of help;
\item with the exception of the above references, this thesis is entirely my own work;
\item appropriate ethics guidelines were followed to conduct this research;
\item for work done jointly with others, my contribution is clearly specified.
\end{itemize}

\addcontentsline{toc}{chapter}{Statement of Authorship}
\clearpage

\begin{abstract}
Virtually every sector of society is experiencing a dramatic growth in the volume of
unstructured textual data that is generated and published, from news and social media online interactions, through open access scholarly communications and observational data in the form of digital health records and online drug reviews. The volume and variety of data across all this range of domains has created both unprecedented opportunities and pressing challenges for extracting actionable knowledge for several application scenarios. However, the extraction of rich semantic knowledge demands the deployment of scalable and flexible automatic
methods adaptable across text genres and schema specifications. Moreover, the full potential of these data can only be unlocked by coupling information extraction methods with Semantic Web techniques for the construction of full-fledged Knowledge Graphs, that are semantically transparent, explainable by design and interoperable. 

In this thesis, we experiment with the application of Natural Language Processing, Machine Learning and Generative AI methods, powered by Semantic Web best practices, to the automatic construction of Knowledge Graphs from large text corpora, in three use case applications: the analysis of the Digital Transformation discourse in the global news and social media platforms; the mapping and trend analysis of recent research in the Architecture, Engineering, Construction and Operations domain from a large corpus of publications; the generation of causal relation graphs of biomedical entities from electronic health records and patient-authored drug reviews. 

The contributions of this thesis to the research community are in terms of benchmark evaluation results, the design of customized algorithms and the creation of data resources in the form of Knowledge Graphs, together with data analysis results built on top of them. Most of the material presented in this thesis originates from
research publications in international journals or conference proceedings.

\end{abstract}
\clearpage

\chapter*{Biography}
\addcontentsline{toc}{chapter}{Biography}

Vanni Zavarella was born on June 23, 1978 in Sulmona (Italy). He is a PhD Candidate in Computer Science at the Department of Mathematics and Computer Science, University of Cagliari (Italy), under the supervision of Prof. Gianni Fenu and co-supervision of Prof. Diego Reforgiato Recupero. He received a MSc Degree in ``Computer Science, Cognitive Science and Applications" from the 
University of Lorraine, France (formerly University of Nancy 2), with a specialization in Natural Language Processing.

Currently working as a freelance data scientist and NLP developer, he has served for more than 14 years as a Scientific Officer and consultant at the European Commission’s Joint Research Centre (JRC) in the implementation and management of Natural Language Processing projects, being responsible for supporting JRC text analysis and media monitoring services in domains such as open source intelligence (OSINT), Global Health Surveillance and social media mining for Disaster Management. He is a former core developer of the popular news monitoring platform Europe Media Monitor (EMM). 

He has co-authored over 40 research publications in international conferences and journals and has given talks and poster presentations at several conferences and workshops, including  FSMNLP 2008, EACL 2014, ESWC 2014, ISCRAM 2017, LREC 2020, ECIR 2020, TEXT2KG 2024, UMAP 2024, LOD 2024 and 2025, etc. He has been program and organizing committee member of the CASE (Challenges and Applications of Automated Extraction of Socio-political Events from Text) workshops series.
\clearpage

\chapter*{Dissemination}
\label{dissemination}
\addcontentsline{toc}{chapter}{Dissemination}

The topics, techniques and resources presented in this Ph.D. thesis are the product of research efforts the resulted in scientific publications in international journals, conference proceedings and workshop proceedings. I express sincere gratitude to my co-authors for their invaluable contributions, which I acknowledge through the inclusive use of the scientific ’we’
throughout this thesis. Moreover, during my nine month long research stay abroad at the Institute of Data Science and Artificial Intelligence of the Universidad of Navarra (DATAI), I had the privilege of collaborating with PhD Juan Carlos Gamero Salinas. The work described in Chapter~\ref{sec:aeco} is the result of this collaboration.

I conceived the research concepts outlined in this thesis and undertook the majority of the research, implementation, testing and evaluation work. I conceptualized the methodologies, determined the research trajectories, and collected and analyzed the necessary datasets. The responsibility for script implementation also fell within my purview. Furthermore, I undertook the authorship of the papers, expertly navigating the peer-review process and iteratively refining them. My interactions with the co-authors were characterized by close collaboration and consultation. Their input encompassed offering insights into methodologies, providing technical assistance, engaging in the exploration of techniques, and contributing to the refinement of submitted work. Additionally, I assumed the role of presenter for 3 of the papers at conferences and workshops. 

The detailed references to the produced papers are provided below.
\\

\textit{Peer-reviewed Publications in International Journals:}

\begin{enumerate}[label=\roman*.]
    \item \textbf{Vanni Zavarella} \& Sergio Consoli, Diego Reforgiato Recupero, Gianni Fenu, Simone Angioni, Davide Buscaldi, Danilo Dessì, Francesco Osborne (2024). \textit{Triplétoile: Extraction of knowledge from microblogging text.} Heliyon, Volume 10, Issue 12, e32479 DOI: 10.1016/j.heliyon.2024.e324 (\color{blue}ISI/Scimago Q1)\color{black}
    
    \item \textbf{Vanni Zavarella} \& Juan Carlos Gamero-Salinas, Danilo Dessì, Sergio Consoli, Gianni Fenu, Diego Reforgiato Recupero \textit{Mapping the AECO Research Landscape using Topic Modeling, Bibliometrics and Information Extraction methods}. under review by IEEE ACCESS 
    
    \item  \textbf{Vanni Zavarella} \& Lorenzo Bertolini, Sergio Consoli, Gianni Fenu, Diego Reforgiato Recupero, Alessandro Zani. \textit{Leveraging Large Language Models for Causal Relation Extraction in Biomedical Texts} under review by Information Processing and Management, Special Issue on Causal Reasoning in Language Models.
\end{enumerate}

\textit{Peer-reviewed Publications in International Conference and Workshop Proceedings:}
\begin{enumerate}[label=\roman*.]
    \item \textbf{Vanni Zavarella} \& Sergio Consoli, Diego Reforgiato Recupero and Gianni Fenu (2024) \textit{Exploring Digital Health Trends in the Headlines via Knowledge Graph Analysis} proceedings of the 10th International Conference on Machine Learning, Optimization, and Data Science (LOD2024) by Springer Nature - Lecture Notes in Computer Science \url{https://www.springer.com/gp/computer-science/lncs} (\color{red} Presenter\color{black})
    
    \item \textbf{Vanni Zavarella} \& Diego Reforgiato, Sergio Consoli, Gianni Fenu (2024). \textit{Charting the Landscape of Digital Health: Towards A Knowledge Graph Approach to News Media Analysis} Adjunct Proceedings of the 32nd ACM Conference on User Modeling, Adaptation and Personalization https://dl.acm.org/doi/abs/10.1145/3631700.3665237) (\color{blue}Rank B)\color{black}

    \item \textbf{Vanni Zavarella} \& Diego Reforgiato Recupero, Sergio Consoli, Gianni Fenu, Simone Angioni, Davide Buscaldi, Danilo Dessí, Francesco Osborne (2024). \textit{Knowledge Graphs for Digital Transformation Monitoring in Social Media} CEUR Workshop Proceedings, 3rd International Workshop on Knowledge Graph Generation from Text (TEXT2KG), co-located with ESWC 2024 (\url{https://ceur-ws.org/Vol-3747/text2kg\_paper8.pdf})
    
    \item \textbf{Vanni Zavarella} \& Juan Carlos Gamero-Salinas, Sergio Consoli (2024) \textit{A Few-Shot Approach for Relation Extraction Domain Adaptation using Large Language Models} Proceedings of the Workshop on Deep Learning and Large Language Models for Knowledge Graphs (DL4KG@KDD2024) \url{https://ceur-ws.org/Vol-3894/} (\color{red} Presenter\color{black})
    
    \item \textbf{Vanni Zavarella} \& Lorenzo Bertolini, Sergio Consoli, Gianni Fenu, Diego Reforgiato Recupero, Alessandro Zani. (2025) \textit{LLM-Powered Knowledge Graph of Causal Relations in Drug Reviews} CEUR WORKSHOP PROCEEDINGS, 4th International Workshop on LLM-Integrated Knowledge Graph Generation from Text (Text2KG) \url{https://ceur-ws.org/Vol-4020/Paper_ID_9.pdf}

    \item \textbf{Vanni Zavarella} \& Lorenzo Bertolini, Sergio Consoli, Gianni Fenu, Diego Reforgiato Recupero, Alessandro Zani (2025). \textit{An Interactive Dashboard for Exploring Patient-Reported Drug-Condition Relations} under publication as conference post-proceedings of the 11th International Conference on Machine Learning, Optimization, and Data Science (LOD2025) by Springer Nature - Lecture Notes in Computer Science  (\color{red} Presenter\color{black})                                              
\end{enumerate}

\chapter*{Acknowledgments}
\addcontentsline{toc}{chapter}{Acknowledgments}

I would like to express my gratitude to my supervisor, Prof. Gianni Fenu, and to my co-supervisor Prof. Diego Reforgiato Recupero for their continuous support and encouragement and for their invaluable management and strategic guidance throughout this research. Their insight and expertise have been instrumental in shaping this thesis and achieving the scientific results we have reached.

This work has leveraged the Collaboration Agreement (CA) \#36805 between the Joint Research Centre of the European Commission and the Department of Mathematics and Computer Science of University of Cagliari aiming to develop Data Science applications for Healthcare, exploiting the value of health data by leveraging on novel cutting-edge technologies like those from Data Science and (Deep) Machine Learning, ensuring that the results obtained are used to support policy-making.

In this respect, I would like to thank the colleagues of the Digital Health Unit (JRC.F7) at the Joint Research Centre for the helpful guidance and support during the development of this research work. In particular, I would like to express my personal gratitude to Dr Sergio Consoli from JRC for his tireless support on improving the scientific rigor of this work. Nonetheless, all the views expressed in this thesis are purely mine and may not in any circumstance be regarded as stating an official position of the European Commission.

I would like to warmly thank the Institute of Data Science and Artificial Intelligence of the Universidad of Navarra and his Director, Prof. Jesús López Fidalgo, for making my research stay with them so enriching and in particular Juan Carlos Gamero Salinas, for helping making our scientific collaboration so fruitful and personally enjoyable at the same time. Moreover, I would like to collectively thank the various faculty members and researchers of the School of Architecture of the University of Navarra for their volunteer validation work on the SKG-AECO pipeline described in Chapter~\ref{sec:aeco}.

Above all, I owe my deepest gratitude to my partner Ana, for her unconditional love and faith in me throughout this journey. Her encouragement has been my constant source of motivation.

Finally, I dedicate this thesis to my parents Laura and Vittorio, their work ethic and high respect for the culture and science has been and will always be a guiding light in my professional and personal life.


\chapter*{Nomenclature}
\addcontentsline{toc}{chapter}{Nomenclature}

\section*{Abbreviations}

\begin{tabular}{ll}
NLP  & Natural Language Processing \\
AI  & Artificial Intelligence \\
KG  & Knowledge Graph \\
SW  & Semantic Web \\
RDF & Resource Description Framework \\
IRI & Internationalised Resource Identifier \\
OWL & Ontology Web Language \\
KB  & Knowledge Base \\
ANN & Artificial Neural Networks \\
DL & Deep Learning \\
RNN & Recurrent Neural Network \\
CNN & Convolutional Neural Network \\
PEFT & Parameter-Efficient Fine-Tuning \\
GNN & Graph Neural Network \\
ML  & Machine Learning \\
IE  & Information Extraction \\
NER  & Named Entity Recognition \\
RE  & Relation Extraction \\
EL & Entity Linking \\
CRE & Causal Relation Extraction \\
ADE & Adverse Drug Event \\
LLM & Large Language Model \\
DT & Digital Transformation \\
AECO & Architecture, Engineering, Construction and Operations \\
\end{tabular}

\section*{Numerical Expressions}

 \begin{tabular}{ll} 
 \{n\}k & \{n\} thousands\\
 \{n\}M & \{n\} millions\\
 \{n\}B & \{n\} billions\\
 \{n\}T & \{n\} trillions\\
 \end{tabular}

\tableofcontents
\listoffigures
\listoftables

\mainmatter
\pagenumbering{arabic}

\chapter{Introduction}
\label{sec:intro}

\section{Motivation}

Virtually every sector in society is experiencing a dramatic growth in the volume of unstructured textual data that is generated, stored and exchanged.

For example, news and social media (SM) online interactions produce  massive streams of multi-domain text content timely reflecting both viral events or long term societal trends on a local to global level.

The globalization of scientific communities and the establishment of open access standards and dissemination platforms has brought about an average 4–6\% yearly growth rate of scholarly communications~\cite{https://doi.org/10.1002/asi.23329}, which in some areas are estimated to double their volume roughly every 10 years. Despite the advancement of indexing and semantic retrieval services, the volume of scientific publications are far beyond the manageable scope of human analysis or surveying. At the same time, the knowledge encoded in the text content of such documents  remains still inaccessible to machine services.

In the medical domain, the digitization of health records and the increasing volume of patient-reported experiences with drugs and therapies in online forums, specialized websites and social media channels have opened up completely new scenarios for the passive collection of observational data~\cite{Fernainy2024}.

The volume and variety of data across all this wide range of domains has created both unprecedented opportunities and pressing challenges for extracting actionable knowledge for several application scenarios. 

First, the monitoring and analysis of large and pervasive societal processes such as Digital Transformation (DT) can be effectively carried out via the detection and tracking of its key players' interactions from fast responsive SM platforms. Secondly, the dynamics of the scientific research and innovation in a determined domain can be studied by tracking the research entities and their semantic relations, such as the \textit{tasks}, \textit{methods}, \textit{algorithms} and the \textit{metrics} they have been evaluated against, hidden within the text of scholarly publications. Finally, the knowledge extracted from clinical notes and crowdsourced patient reviews can be used to extend and update current authoritative Knowledge Bases (KB) about medical entities like drugs, therapies, conditions and their complex interactions. 

However, the extraction of rich semantic knowledge from the sheer volume and variety of these data demands the deployment of scalable and flexible automatic methods leveraging compact and easy-to-process representations, capable to be adapted to different input text genres and characteristics and to the schema specifications of the given use case application.

Within the Natural Language Processing (NLP) field, several frameworks and techniques have been developed that generalize statistical patterns over linguistics features in the text data to extract abstract concepts such as named entities, topics, relations or events. One step further, Deep Learning (DL) architectures are able to discover complex patterns from large labeled data with no or minimal feature annotation, leveraging highly contextualized representations of the tokens in a text. More recently, the explosion of decoder-only pre-trained Large Language Model (LLM) architectures has enabled to solve high-level language understanding tasks with zero-shot or few-shot inference methods, sparing the need of collecting costly training datasets.

The data potential, though, can only be fully unlocked by coupling the above mentioned methods with Semantic Web (SW) techniques for the construction of full-fledged KGs out of domain data collections. KG representations are semantically transparent, explainable by design and machine-readable. Consequently, the knowledge extracted from a given text collection and consolidated using SW techniques can be retrieved and aggregated using semantically explicit predicates. Moreover, it is interoperable with and can update or expand existing knowledge repositories modeling the same domain, via linking of uniquely identified entities and predicates. 

In this thesis, we experiment with the application of NLP, ML and Generative AI methods, powered by SW techniques and best practices, in three application domains and categories of text collections:
\begin{enumerate}
    \item the analysis of DT discourse in the global news and SM platforms;
     \item the mapping and trend analysis of the research in the Architecture, Engineering, Construction and Operations (AECO) domain from a large corpus of recent publications;
    \item the construction of causal relation graphs from health records and patient-authored drug reviews.
\end{enumerate}

\section{Challenges}
\label{challenges}
Unlocking the latent value of unstructured data for each of these use cases requires addressing several challenges, pertaining to the characteristics of the data itself and of the target domain.

Social media messages are typically short texts with little explicit context, using colloquial and often noisy language with context-dependent platform-specific expressions like hashtags. Scholarly publications contain technical discourse conventions, abbreviations, acronyms that are specific to a single domain and often ambiguous across domains. Clinical notes contain highly specialized jargon, frequent use of abbreviations for a large technical terminology of drugs, diseases, symptoms, dosages, etc. These text characteristics are challenging for standard NLP methods that need to be customized in order to extract entities and relations with acceptable recall. 

Moreover, while standard named entities of type \textit{Person}, \textit{Location} or \textit{Drug} have relatively lower variability in text, more abstract entities such as \textit{Energy Efficiency}, categorized as instance of a general type \textit{Task}, are far more ambiguous and hard to be merged to a canonical form and ultimately to link to entries in reference KBs.  

Finally, as it will be shown, some application scenarios (like DT monitoring of Chapter~\ref{sec:digitalTransformation}) are better fit for a data-driven, unsupervised generalization approach, rather than to a prior specification of the target entity-predicate schema.

The main research questions addressed in this thesis, distributed across the three use cases, are:

\textbf{Q1}. How NLP and Semantic Web technologies can be combined to extract knowledge from noisy user-generated text collections and represent it in interoperable formats?

\textbf{Q2}. Which NLP and ML techniques better fit different application scenarios?

\textbf{Q3}. How can KG representations enhance the analysis of trends within a specific domain?

\textbf{Q4}. Can Generative AI techniques support the construction of scientific knowledge graphs of very abstract research concepts in technical domains, with only limited customization?

\textbf{Q5}. Which Generative AI techniques and LLM architecture and training methods are best suited to the generation of causal graphs from medical texts?

\section{Contributions}

This thesis makes contributions to the advancement of research on NLP and KG generation methods in terms of experimental results and of generated models, data resources and data analytics. The main contributions are:

\begin{itemize}
    \item The design and evaluation of an open IE pipeline for KG extraction from micro-blogging and news articles text collections
    \item the release of KG data and visualization dashboards for the analysis of the DT discourse in news and social media
    \item the enhancement of an existing methodology for building research knowledge graphs from scholarly data and its customization to the AECO domain
    \item the release of a prototype data visualization dashboard for the analysis of AECO research trends
    \item A systematic benchmark evaluation of LLMs architectures, inference and learning techniques for a Causal Relation Extraction (CRE) task in the medical domain
    \item The release of a fine-tuned model for CRE and a demonstration of its applicability for building causal graphs of Drug-Adverse Drug Event (ADE) interactions
\end{itemize}

\section{Outline}

The thesis is organized as follows:

\begin{itemize}
    \item In Chapter~\ref{sec:kgBuilding} we introduce the technical definitions of KGs, KG properties and SW techniques that we will be using throughout this thesis. Afterwords, we will briefly define the NLP, deep learning (DL) and enerative AI methods that are backing the whole KG extraction process. The aim of this chapter is not to provide a comprehensive survey of all techniques from the research literature but rather to offer a focused overview of the methodological options suitable for our particular use cases, thereby providing a rationale for the techniques presented in the subsequent chapters;      
    \item Chapter~\ref{sec:digitalTransformation} presents the use case application of NLP-based KG generation techniques to large collections of news and social media crowdsourced data for the monitoring of DT discourse;
    \item In Chapter~\ref{sec:aeco} we describe how the integration of topic modeling, LLMs and SW techniques can be used to generate a large scale scientific KG of the AECO research landscape and thus support research trend analysis in this domain;
    \item Chapter~\ref{sec:causalRE} discusses the role of causality graph data in medical knowledge bases and pharmacovigilance and it benchmarks a large range of LLM architectures and methods for the generation of causal graphs of biomedical entities from clinical notes and drug reviews;
    \item Finally,  Chapter~\ref{sec:conclusions} presents a discussion of the limitations of the proposed methods within the studied application domains along with an overview of the directions of our ongoing research on these topics.
\end{itemize}

\chapter{Knowledge Graphs Construction Methods}
\label{sec:kgBuilding}

\section{Knowledge Graphs for domain knowledge representation}
\label{sec:kgs}

KGs are a very flexible formalism for representing data in a domain of interest as directed edge-labeled (DEL) graphs, where nodes represent domain objects and edges represent binary relationships between these objects.
Formally, they are described by a tuple $G =(V,E,L)$ where $V$ is a finite set of nodes, $L$ a finite set of labels and $E \subseteq V \times L \times V $ is a set of edges.
An edge is often conventionally represented as a triple $(h,r,t)$, where $r \in L$ is a relation predicate connecting a ``head" node $h$ to a ``tail" node $t$.

\begin{figure*}[!ht]
\centering
\begin{minipage}[b]{0.9\textwidth}
    \centering
    \includegraphics[width=0.8\linewidth]{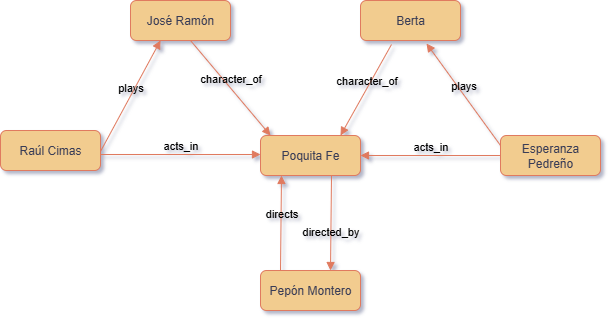}
    \\a.
\end{minipage}
\hfill
\\[1em]
\begin{minipage}[b]{0.7\textwidth}
\centering
   \begin{tcolorbox} [fontupper=\ttfamily\scriptsize,fontlower=\ttfamily\scriptsize, colback=yellow!5!white,colframe=yellow!50!black]
@prefix sdbo: <http://series-database/ontology\#> .

@prefix sdb:  <http://series-database/resource> .

@prefix rdf: <http://www.w3.org/1999/02/22-rdf-syntax-ns\#> .

@prefix rdfs: <http://www.w3.org/2000/01/rdf-schema\#> .

@prefix owl: <http://www.w3.org/2002/07/owl\#> .

@prefix dbr: <http://dbpedia.org/resource/> .

@prefix dbo: <http://dbpedia.org/ontology/> .

sdb:raúl\_cimas  rdf:type owl:NamedIndividual ;

\hspace*{2em}  rdf:type sdbo:Actor ;
                 
\hspace*{2em}   sdbo:hasName  "Raúl Cimas".

sdb:poquita\_fe rdf:type owl:NamedIndividual ;

\hspace*{2em} owl:sameAs dbr:poquita\_fe .

sdbo:Actor rdf:type owl:Class ;

\hspace*{2em}   rdfs:subClassOf sdbo:Performer .

sdbo:acts\_in rdf:type owl:ObjectProperty ;

\hspace*{2em} rdfs:subPropertyOf sdbo:performs\_in ;

\hspace*{2em} rdfs:domain sdbo:Actor .

\end{tcolorbox}
b.
\end{minipage}
\caption{Fragment of a TV series knowledge graph (a.), with corresponding RDF serialization in Turtle format (b.).}
\label{fig:SeriesDBKGExample}
\end{figure*}

The formalism is expressive enough to enable encoding n-ary relations with $n > 2$ without the need to change the whole data schema, like in a relational database. For example, in the fragment of a sample KG of Spanish TV series in Figure~\ref{fig:SeriesDBKGExample}, one can add an intermediate ``character" node \textit{Jose Ramón} to further qualify the binary \textit{acts\_in} relation between actor \textit{Raúl Cimas} and the TV series \textit{Poquita Fe}. Notice also that the basic formalism does not impose constraints on the graph topology, for example inverse relations connecting pairs of nodes (like \textit{directs} and \textit{directed\_by} in Figure) can introduce cycles.

No matter how flexible and expressive, interoperability requires KGs to explicitly define the semantics of the nodes and relation labels. A minimal formalization of DEL graph semantics is achieved by using RDF/RDF Schema. 

\paragraph{RDF}RDF is a standardized data model that enforces restrictions on node/edge identifiers. Namely, nodes can be:
\begin{itemize}
\item Internationalised Resource Identifiers (IRIs), i.e. global persistent identifiers that can be looked up by web-servers to return RDF descriptions of the entity;
\item literals, for representing strings and other XML Schema datatypes
\item blank nodes, anonymous nodes used for representing more complex data structures like lists, etc.
\end{itemize}

For example, in the RDF serialization of Figure~\ref{fig:SeriesDBKGExample}.b, the node labeled as \textit{Raúl Cimas} is identified by an IRI like \textit{http://series-database/resource/raúl\_cimas}, assuming \textit{http://series-database/resource} is the namespace of our TV series KG, and it is connected to the literal ``Raúl Cimas" via a data property (that is, a string valued predicate) \textit{hasName}. The unique namespace \textit{http://series-database/resource} can be prefixed (like in \textit{sdb:raúl\_cimas}) and prevents name clashing with resources in other KGs modeling the same domain entities.

\paragraph{RDFS} RDF Schema is a metalanguage making available predefined predicates that allow a partial definition of the semantics of the terms in a RDF KG (what is often called the KG's ontology). For example, one might further specify the information in the sample KG of Figure~\ref{fig:SeriesDBKGExample} by making explicit that the resource \textit{sdb:raúl\_cimas} is an instance of class \textit{sdbo:Actor} defined by the KG ontology, which on its turn is a subclass (\textit{rdfs:subClassOf}) of the \textit{sdbo:Performer} class. Moreover, \textit{sdbo:acts\_in} is an \textit{owl:ObjectProperty} and a subproperty of \textit{sdbo:performs\_in}, etc. Additional constraints can be enforced by using the full range of RDFS predicates or defining a set of rules using the more expressive OWL ontology language.

\paragraph{Bridging}A major mechanism for knowledge integration in the Semantic Web is bridging. It consists of referencing external resources (so called named graphs) in the namespace declarations of a KG and then using the OWL predicate \textit{owl:sameAs} to state equality of individuals, like in Figure~\ref{fig:SeriesDBKGExample} where the TV series \textit{sdb:poquita\_fe} is stated to be equivalent to the DBpedia entity \textit{dbr:poquita\_fe}.

In this way, one can access heterogeneous information about the same individuals as encoded in different KGs. This opens up a vast analytical potential when doing knowledge retrieval from KGs, as SPARQL queries can get evaluated against the union of a default RDF and the additional set of named graphs.

\paragraph{DBpedia} With many KGs linked to it and by bridging to several external open resources, DBpedia is one of the most central knowledge hubs in the Linked Open Data ecosystem.

It is a large scale, open domain and multilingual graph that has been collectively developed under the open data philosphy~\cite{dbpedia2023}. It is constantly expanded using an information extraction framework that parses Wikipedia infoboxes to identify property-value pairs and align them with a community-maintained ontology. This guarantees that DBpedia remains up-to-date to the evolving structure of Wikipedia. As of release 3.8 (2015) DBpedia features 3.77M entities just for English, but recent snapshots counted over 850M triples and a total of nearly 1.9B RDF statements.

We perform bridging to DBpedia for the KGs described in Chapters~\ref{sec:digitalTransformation} and~\ref{sec:aeco}.

\paragraph{Density Measures} KGs can be characterized by diverse density measures (or equivalently, inverse sparsity measures), which have an impact on the effectiveness of the graph generation techniques discussed in the subsequent sections~\cite{KONG2026111876}. 

\textit{Graph-theoretic density} measures how many edges are present in a graph $G =(V,E,L)$ relative to the maximum possible:
\begin{equation}
Density(G) = \frac{\vert E \vert}{\vert V\vert \cdot (\vert V\vert  - 1)}
\end{equation}
It is a raw density measure as KGs are typically extremely sparse in this respect, because not every pair of entities should be linked, depending on the relation label.

\textit{Relation Type Instantiation} defines how much of the potential space of valid triples for a relation type is instantiated in the graph. That is, if $Dom(r)$ and $Range(r)$ are respectively the domain and range of a relation label $r$ in $G$,
\begin{equation}
Density(r)=\frac{\vert\{(h,r,t)\in  G\}\vert}{\vert Dom(r)\vert \cdot \vert Range(r)\vert}
\end{equation}

\textit{Density per Entity} measures the average degree (number of edges) per entity $V$ (or per entity type) in a graph:
\begin{equation}
AvgDeg(V) = \frac{\vert 2E \vert }{\vert V \vert }
\end{equation}

Finally, \textit{Schema Coverage} (also called ontology usage density), measures the fraction of schema-defined relations that appear at least once in the KG. This is useful to quantity coverage imbalance, that is how much unequally the KG instantiates the relations in the ontology, and semantic density, that is how fully the ontology’s expressive capacity is realized in the KG facts. 

\color{black}
\section{Generating Knowledge Graphs}

While KG are a semantically transparent representation of structured data and support effective methods for retrieving and inferring knowledge from them, their symbolic nature makes them hard to be manipulatet at scale.

Therefore, recent years have seen the emergence of inductive knowledge discovery methods for KGs that are based on compact representations of the graphs on low-dimensional vector spaces, known as graph embeddings, that preserve the latent structure of the graph and at the same time can be used my standard ML algorithms for inducing new knowledge.

The general idea of graph embeddings is to apply a self-supervised approach where, after initializing a random embedding vector representation of each entity $e \in \mathbb{E}$  and relation $r\in \mathbb{R}$, a representation for each triple $(h,r,t)$ in the graph (e.g. $(AlfredHitchcock,\,DirectorOf,\,Psycho)$) is learned by optimizing on a scoring function that maximizes the plausibility of the set $\mathbb{D}^+=\{(h,r,t)\}$ of all the facts in the graph (positive triples) and minimizes the plausibility of (synthetically generated) negative triples.

In its basic version (translational models such as TransE~\cite{10.5555/2886521.2886624}), graph embedding methods represent both entities and relations in the same $d$-dimensional vector space $\mathbb{R}^d$ and interpret relations as translation vectors \textbf{r} connecting the vectors \textbf{h} and \textbf{t}, such that if $(h,r,t)$ holds, $\textbf{h} + \textbf{r} \approx \textbf{t}$~\cite{8047276}. Here, the underlying intuition, taken from the paradigm of distributional semantics~\cite{Mikolov2013EfficientEO}, is that a vector representation, say for the relation $DirectorOf$, must guarantee that both $JamesCameron + DirectorOf \approx Avatar$ and $AlfredHitchcock + DirectorOf \approx Psycho$ hold. Therefore, the scoring function is defined as the negative distance between $\textbf{h} + \textbf{r}$ and $t$, that is:
\begin{equation}
f_{r}(h,t) = -\|\textbf{h} + \textbf{r} - \textbf{t}\|_{1/2}
\end{equation}
Other variants of graph embeddings have been proposed to solve some of the shortcomings of translational models, such as for example semantic matching models that use scoring functions matching the latent semantic components of entity and relation vectors (tensor decomposition models~\cite{rabanser2017introductiontensordecompositionsapplications}), or RDF2Vec~\cite{Ristoski2016RDF2VecRG}, that linearize a graph as a set of sentences by randomly traversing it and collecting the visited paths and finally feed the collected sentences to a standard language embedding learning algorithm such as word2vec~\cite{Mikolov2013EfficientEO}.

Instead of creating dense vector representations of graphs to be fed standard ML algorithms, Graph Neural Networks (GNN) are ANN architectures that mirror the topology of the graph data~\cite{10.1145/3503043}, with network nodes representing graph nodes and node connections representing graph edges.    

The  general idea here is that each node and edge in the graph is associated with a feature vector (one-hot representation or based on node attributes), while an embedding vector $h_v$ (or \textit{state vector}) for a node $v$ is learned by iteratively applying for several layers $k$ a ``message passing" function:
\begin{equation}
h_v^{(k+1)} = f(W \cdot AGG\{h_u^{(k)}: u \in N(v) \cup \{v\}\})
\end{equation}
where $N(v)$ is the set of neighbors of node $v$, $AGG$ is an aggregation function and $W$ is a learnable weight matrix. In practice, a GNN updates node embeddings for all its nodes by aggregating information from their neighbors in the graph. Then another function $g$ is used to compute an output value for a node $v$ based on its embedding vector $h_v$, its feature vector $n_v$ and a weight matrix $W'$, that is:
\begin{equation}
o_v^k = g(W',h_v^k,n_v)
\end{equation}
The idea here is that feature vectors remain fixed across the learning process, while the desired output values are given only for a subset of supervised nodes of the graph, which represent the training set of the process. The framework is able to learn the parameters $W$ and $W'$ of the two functions that generate the expected output, and the output function $g$ is then applied to other nodes of the graph.   

What all the presented frameworks share is the capability to leverage the latent properties of an initial KG nucleus for downstream tasks of KG completion. Typical graph completion tasks involve link prediction, where a missing edge between two existing nodes is predicted based on characteristics of the involved nodes and their connectivity patterns with other nodes in the graph~\cite{10.3233/SW-160218}.

For example, for a task of predicting which entity head $h$ is connected via a relation $r$ to an entity tail $t$ (i.e. $(?,r,t)$) one can take every candidate entity $h'$ in the graph, compute the corresponding score $f_{r}(h',t) = -\|\textbf{h'} + \textbf{r} - \textbf{t}\|_{1/2}$ based on the learned embeddings and scoring function (in this case, using a translational model), ranking the scores in descending order and finally add predicted triples based on some heuristic thresholds on these ranked scored. 

However, overall these techniques best fit use case scenarios where:
\begin{itemize}
\item an existing nucleus of graph-encoded knowledge is already available and at the same time there is scarcity of other sources of information from where additional knowledge could be derived; 

\item a set of relation labels is already defined in a KG ontology;

\item the target graph that is to be built has a significant level of expected relation instantiation and entity density, as defined in Section~\ref{sec:kgs};

\item there are no strong requirement for explainability or source traceability of the automatically-induced knowledge~\cite{molnar2025}

\end{itemize}

However, the use cases we discuss in this study exhibit traits that, to varying extents, fail to comply with one or more of the conditions listed above.

In all three use case scenarios, a rich and up-to-date source of information is available in the form of a large collection of unstructured text from which the target KG is to be generated from scratch. The nucleus of pre-existing knowledge about the nodes of the target graph is not necessarily null: for example, in the DT monitoring use case in Chapter~\ref{sec:digitalTransformation} the entities we target are described in a number of reference KGs we eventually link to. However, the set of relations we aim to derive are meant to be an extension to the static body of relations already encoded in those reference KGs, so that few evidence on these new relation types is initially available. Moreover, in some use cases (such as the DT monitoring) the set of relation labels is simply not given \textit{a priori} and instead is being generalized from the relation instances extracted from the text collection.
Overall, most of the target relations we aim to induce are very sparse, therefore graph embeddings would struggle due to lack of training signal or overfit to the few represented relation types.

Finally, for sensitive domain use cases like the causal relation graph of drug and ADE entities introduced in Section~\ref{sec:causalRE}, the acquired relation instances might need to be explainable or at least traced back to the source document which generated them, while an holistic induction from global patterns of the graph without human-interpretable justification might not be acceptable.

For these reasons, in this thesis we make use of and describe exclusively extractive KG construction techniques that generate graph nodes and edges via the application of NLP/DL methods to large collections of unstructured data, i.e. text corpora.

 \section{Knowledge Graphs Construction Process}
\label{sec:KGConstructionChallenges}

The overall task of extracting a KG from a collection of unstructured text documents can be factorized into two main sub-tasks~\cite{10.3233/SW-180333,10.1016/j.websem.2015.12.004}, namely:
\begin{enumerate}
    \item Entity Extraction and Linking (EEL), which consists of detecting in the input text occurrences (mentions) of named entities and associating them with existing unique nodes in an external Knowledge Base (e.g. via DBpedia IRI identifiers), or creating novel candidate identifiers in the KB (also called emerging entities). The detected entities represent the nodes of the generated graph and can be assigned a type (typically via \textit{rdf:type} property) contextually with the extraction process or by inheriting from the linked identifier in an external KB.
    \item Relation Extraction and Linking (REL), which consists of extracting $n$-ary relations (with $n \geq 2$\footnote{In all the use cases described in this thesis, we will only deal with binary relations.}) between detected entities in the input text, linking the relation predicates to (object or data) properties in a reference ontology or to a newly defined relation predicate\footnote{As we will describe later on, this latter option characterizes the Open Information Extraction paradigm.}. When both the relation predicate and entities get linked to a KB or when identifiers are created for them, the result of the EEL and REL process can be used to populate the KB with additional facts.
\end{enumerate}

\begin{figure}[!htbp]
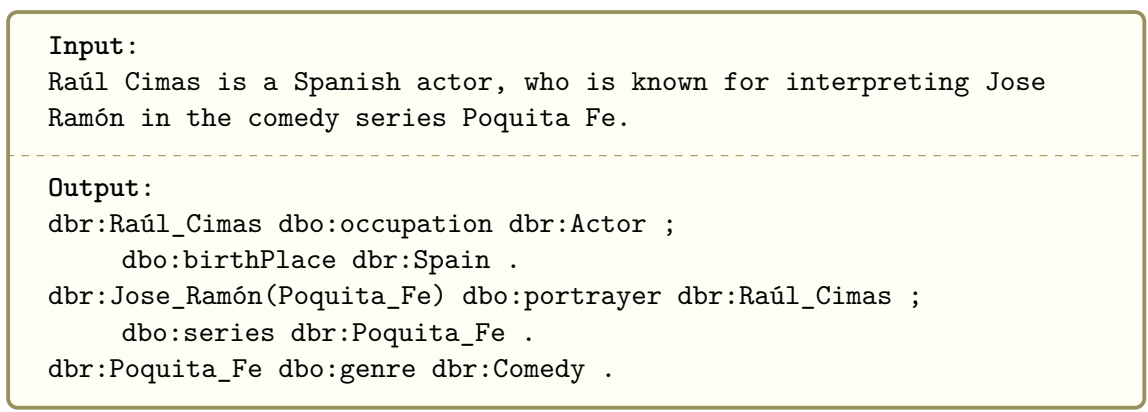

\begin{tcolorbox}[fonttitle=\ttfamily\small,fontupper=\ttfamily\small,fontlower=\ttfamily\small, colback=yellow!5!white,colframe=yellow!50!black]
\textbf{Input}:

Raúl Cimas is a Spanish actor, who is known for interpreting Jose Ramón in the comedy series Poquita Fe.
\tcblower
\textbf{Output}:

dbr:Raúl\_Cimas dbo:occupation dbr:Actor ;

\hspace*{2em} dbo:birthPlace dbr:Spain .
    
dbr:Jose\_Ramón(Poquita\_Fe) dbo:portrayer dbr:Raúl\_Cimas ;

\hspace*{2em} dbo:series dbr:Poquita\_Fe .
    
dbr:Poquita\_Fe dbo:genre dbr:Comedy .
\end{tcolorbox}
\caption{Example of applying Entity Extraction and Linking and Relation Extraction to an input text, using DBPedia as a reference KB.}
 \label{fig:nerReExample}
\end{figure}

The example in Figure~\ref{fig:nerReExample} illustrates the input and (partial) output of the combination of the two tasks, where four entities (\textit{Raúl Cimas}, \textit{Jose Ramón}, \textit{Poquita Fe} and \textit{Spain}) are detected and mapped to their corresponding DBpedia resources (namespace prefix \textit{dbr}), and five relation triples are extracted (e.g. $(Raul\_Cimas,Occupation,Actor)$) and their predicates mapped to DBpedia object properties (e.g. \textit{dbo:occupation}).

\paragraph{Entity Extraction and Linking}

Although the EEL task could be directly accomplished by creating a dictionary of entity labels out of the set of nodes of the reference KB and then applying some form of lexical lookup across the input text collection using this dictionary, this simplistic approach typically suffers from low recall, not to mention that it does not support by definition the extraction of emerging entities that are currently missing in the KB.

Therefore, a standard approach is to split the EEL task into a Named Entity Recognition (NER) step that locates mentions of (generic or typed) entities based on contextual information from the input text where they occur, and an Entity Linking step that leverages the reference KB graph information to map entities to it~\cite{8999622}. Methods here vary significantly with respect to the type of information from entity mentions and KB node candidates they use. Typically, each mention-KB node candidate pair is scored based on string similarity between mention and candidate labels, or keyword-based similarity between the context of the mentions and a context from the reference corpus mapped to the KB (e.g. Wikipedia for DBpedia), or keyword-based similarity with the context of the nodes connected to the candidate node in the KB graph, etc.~\cite{10.1145/2506182.2506198}.

\paragraph{Relation Extraction}

When a reference KB covering the target entities and relations is available, and once the EEL task is solved, a popular REL technique is ``distant supervision"~\cite{hoffmann-etal-2011-knowledge}. This is based on the hypothesis that if a given triple $(h,r,t)$ (say for example:  \textit{(dbr:Poquita\_Fe,dbo:genre,dbo:Comedy}) is found in the KB, the text of a sentence where the triple's entities are mentioned would also contain a mention of the target relation (e.g. ``one of the most innovative comedy series was Poquita Fé"). In this way one can collect a training dataset by heuristically matching the entities of the KB triple set, generalize some linguistic patterns for each target relation predicate, and then apply the pattern to new matched entity pairs, expanding the KB with new triples.

However, for those application scenarios, including the use cases dealt with in this thesis, where context knowledge provided by a pre-existing KBs is null or extremely sparse, the KG construction process is entirely carried out by standard NLP or DL enabling technologies processing solely the input data stream, prior and independently of the KB linking phase. In the remaining sections of this Chapter we outline the main families of such methods and architectures, along with their advantages and disadvantages.

\section{Natural Language Processing Methods}
\label{subsec:nlpMethods}
By NLP methods, we refer here to approaches that solve the Named Entity Recognition and Relation Extraction tasks by using, as part of their input, language structures generated by NLP processing of the raw text. This encompasses both rule-based methods, that build heuristic rules over NLP-generated structures, or learning-based approaches that apply some standard ML techniques over features generated by NLP processing.

\subsection{Dependency Parsing}

One of the central building blocks of NLP-based NER/RE pipelines is dependency parsing. Dependency parsing represents the syntactic structure of a sentence solely in terms of its words (lemmas) and a set of binary grammatical relations connecting pairs of words. 

\begin{figure*}[!ht]
 \centering
 \includegraphics[width=1.0\linewidth, height=1.1\textheight, keepaspectratio]{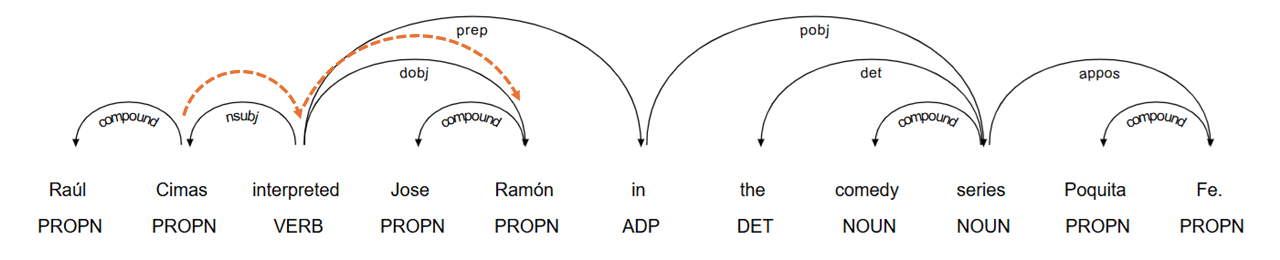}
 \caption{Dependency tree of an example sentence (root node omitted).}
 \label{fig:depTreeExample}
\end{figure*}

Grammatical relations are directed, typed edges from a \textit{head} to its \textit{dependent}. For example, in the parse tree in Figure~\ref{fig:depTreeExample} the main verb lemma ``interpreted" is the head of a \textit{dobj} (direct object) relation to the lemma ``Ramón" (black edge). An inventory of labels for these dependency relations that are valid cross-linguistically is provided by the Universal Dependency set~\cite{de-marneffe-etal-2014-universal}.

Therefore, we can formally define a dependency tree as a directed, labeled multi-relational graph $G =(\varepsilon,\mathcal{R},\mathcal{T})$ with $\varepsilon$ the set of lemmas in the vocabulary plus a \textit{root} node marking the head of the sentence dependency tree, $\mathcal{R}$ the set of Universal Dependency labels, and where the additional constraints hold that: 1) there is a single designated \textit{root} node that has no incoming edge; 2) with the exception of the \textit{root} node, each node has exactly one incoming edge and 3) there is a unique path from the \textit{root} node to each node in $\varepsilon$.

In Figure~\ref{fig:depTreeExample} we also mark in orange the Shortest Dependency Path (SDP,~\cite{bunescu-mooney-2005-shortest}) between the two lemmas ``Cimas"  and ``Ramón". We can describe those shortest paths by collecting the edge labels during path transversal. For example, in this case the path connecting the two target lemmas is of type $(nsubj,dobj)$. In Section~\ref{sec:relExtraction} we will apply constraints on the labels of the paths connecting entities and containing verb lemmas, in order to extract candidate relations.

\subsection{Open Information Extraction}
\label{openIE}

As dependency parsing algorithms are relatively efficient, dependency trees are the standard input representation for a family of RE methods known as Open Information Extraction. 

Open IE is an highly scalable framework that has been succesfully deployed to extract millions of triples from large web corpora. It best fits use cases when a curated relation schema definition (and corresponding manually annotated data) is not provided and one aims to ``discover" the most significant relation types emerging within a target text collection (similarly to the use case of Chapter~\ref{sec:digitalTransformation}). The selection of the extracted triples is purely syntactical, and its soundness is guaranteed by imposing a set of constraints on the dependency trees connecting entity tokens.
These constraints can be in the form of handcrafted rules like in the earlier systems, or they can be patterns learned by distant supervision.

\begin{figure*}[!ht]
 \centering
 \includegraphics[width=0.95\linewidth, height=1.1\textheight, keepaspectratio]{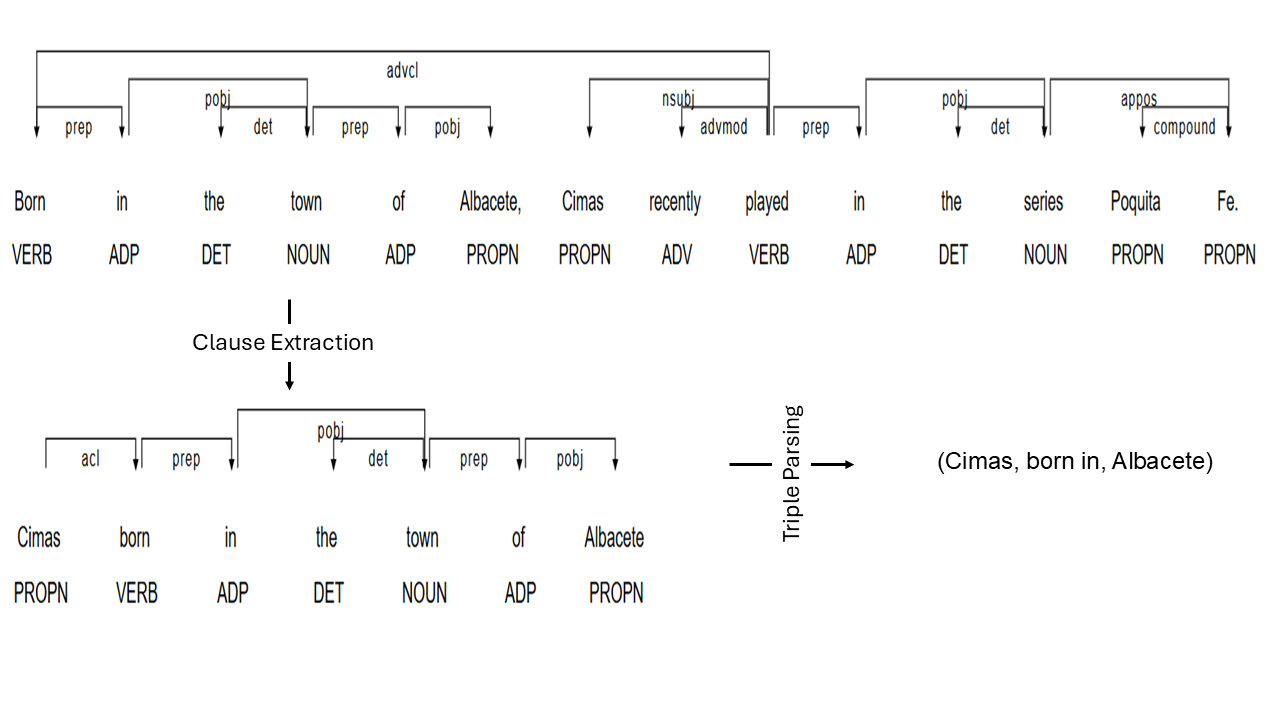}
 \caption{An example of sentence to triple transformation using Open IE.}
 \label{fig:openIEExample}
\end{figure*}

For example,~\cite{angeli-etal-2015-leveraging} uses a two-stage process that: 1.transforms longer sentences into self-contained clauses using a multinomial logistic regression classifier; 2. applies logical inferences to reduce clauses to maximally compact but semantically equivalent sentences, parsing them into simple subject-verb-object triples. A triple extraction example is illustrated in Figure~\ref{fig:openIEExample}.

One of the advantages of Open IE approaches is that they work reasonably well across domains. However, they have the limitation of generating surface-level triples that are not \textit{per se} unified into more abstract predicates that can be later used for triple retrieval. If a reference KB is given, one can try and map relations onto the KB predicates. Otherwise, an unsupervised solution consists of clustering the extracted relation expressions.

\subsection{Relation Clustering}
\label{sec:clustering}
Given a collection of relation instances $R = (r_1,...,r_n)$ the goal of relation clustering is to segment the collection into partitions $C = (C_1,...,C_m)$ , with $m < n$ such that relations within the same partition are more similar than instances across partitions, for some measure of semantic similarity. In cases when the number of resulting partitions cannot be anticipated (like for relation clustering), it is common to resort to \textit{hierarchical clustering} algorithms. One method that is often applied in NLP is HDBSCAN.

\paragraph{HDBSCAN} Hierarchical Density-Based Spatial Clustering of Applications with Noise (HDBSCAN,~\cite{Malzer2020223}) is the hierarchical version of the popular density-based DBSCAN, which is characterized by grouping together points in highly dense regions of the representation space and taking as outliers (therefore leaving unclustered) the data points lying in low-density regions. 

However, HDBSCAN additionally builds a hierarchy of clusters (a dendrogram) over varying density thresholds, by recursively merging smaller clusters of points that are adjacent to each other. The resulting, flat clustering is the most stable one across varying scales.

A standard procedure to make words processable by clustering algorithms is to represent them via pre-trained \textit{word embeddings}, which are distributed representations as dense multidimensional vectors, learned using self-supervised word prediction tasks from large unannotated text corpora, and that capture some features of word meaning. Word embeddings vary as for the language objects they model (from sub-tokens to n-grams), the ANN architecture and the cost function they use for training, and finally the size of the generated vectors representations. Here we will only mention two popular embedding models that we will be using in the following chapters.

\paragraph{GloVe} Global Vectors for Word Representation (GloVe,~\cite{pennington2014glove}) is an unsupervised algorithm that learns 300-dimensional static embeddings aggregating global statistics for word types (that is for predicate lemmas like ``interprets") across all its occurrence contexts in the training corpus (e.g. Wikipedia). Namely, for each lemma it learns a weight matrix that maximizes its co-occurrence probability with lemmas that appear in a text window of given size to the left or right of the lemma, in the reference corpus. The embeddings are called static as, once a lemma is assigned a unique embedding vector at learning time, then at inference time every occurrence of that lemma uses the same embedding, regardless of the context where it appears.

\paragraph{BERT} The Bidirectional Encoder Representations from Transformers (\cite{devlin2019bertpretrainingdeepbidirectional}) is the groundbreaking DL architecture that introduced the \textit{self-attention} mechanism~\cite{vaswani2017attention}, by which each word’s representation is updated by attending to other words in the sequence. For a word $w_i$ in a context (e.g. sentence), BERT embeddings are computed as the sum of embeddings of all other words $w_j$ in the context, weighted by the attention weight matrix controlling how much $w_j$ influences $w_i$. The attention weights are learned by optimizing on self-supervised Masked Language Modeling tasks over large corpora. The resulting embeddings are contextual, as each word’s vector representation is dynamically computed from its surrounding words, rather than being a fixed lookup vector.

\paragraph{Dimensionality Reduction} It is widely recognized as the so called ``curse of dimensionality" problem affects the analysis of high-dimensional data. Specifically, in our case high-dimensional word embedding relation representations require more observed samples to produce a suitable level of density for HDBSCAN to work properly. However, applying UMAP to perform non-linear, manifold-aware dimension reduction~\cite{mcinnes2020umap} has been proven to transform the datasets down to a dimension small enough for HDBSCAN to cluster a large majority of instances.

The UMAP-HDBSCAN combination is controlled by a number of hyper-parameters\footnote{\url{https://umap-learn.readthedocs.io/en/latest/parameters.html}} \footnote{\url{https://hdbscan.readthedocs.io/en/latest/parameter\_selection.html}}, the main ones are described in Table~\ref{tab:UMAPHDBSCANHyperparameters} in  the Appendix~\ref{sec:appendixKGConstruction}, together with some common sample values. In Sections~\ref{relMapping} and ~\ref{sec:topicModeling} we will perform hyperparameter grid search for optimizing the UMAP-HDBSCAN combination for relation clustering and topic modeling, respectively.

\section{Deep Learning Methods}
\label{sec:dlmethods}

Instead of learning functions over an engineered set of features extracted from the input sentences, DL models start with some (learnable or pre-trained and fixed) vector representation of the sentence tokens and incrementally learn the weight parameter matrices of complex ANN architectures that allow mapping token sequences onto output relational triples.

These methods largely vary as regards the input text representations they use and the network architectures deployed to encode the dependencies within sentence components. Moreover, the flexibility of ANN architectures enables the creation of both pipeline frameworks - which sequentially perform entity pair detection followed by relation classification using the detected pairs -  and joint RE frameworks, where the NER and RE tasks are simultaneously solved by the same learning architecture~\cite{zhao2024comprehensivesurveyrelationextraction}.

A survey of the vast variety of DL models for NER/RE is out of the scope of this chapter. We will rather outline here a few classical architectures, some of which will be referenced in the next chapters.

\subsection{BiLSTM for NER}

Named entities are sequences of tokens and the decision whether a token belongs to an entity depends jointly on the features of the token itself and the surrounding ones. Consequently entity detection can be formalized as the task of assigning a label from a tagging schema to each token in an input sentence. In the example below we show an input sentence with on top the expected entity tagging labels, using the IOBES schema\footnote{IOBES stands for Insides, Outsides, Beginnings, Ends and Single-tokens.}~\cite{ratinov-roth-2009-design}, stating for instance that the token ``Raúl" begins a PERSON entity sequence, ``Cimas" is inside a PERSON entity, ``plays" is outside any entity sequence, etc. 
\\
\begin{example}\(\overset{\textbf{B-PER}}{\textit{ Raúl }} \;
\overset{\textbf{I-PER}}{\textit{Cimas}} \;
\overset{\textbf{O}}{\textit{plays}} \;  
\overset{\textbf{B-CHAR}}{\textit{Jose}} \;
\overset{\textbf{I-CHAR}}{\textit{Ramón}} \;
\overset{\textbf{O}}{\textit{in}} \;
\overset{\textbf{O}}{\textit{the}} \;
\overset{\textbf{O}}{\textit{TV}} \;
\overset{\textbf{O}}{\textit{series}} \;
\overset{\textbf{B-MOVIE}}{\textit{Poquita   }} \;
\overset{\textbf{I-MOVIE}}{\textit{ Fe      }} \;
\overset{\textbf{O}}{\textit{   .}}
\)
\end{example}
\\
\\
Intuitively, sequential ANN models like Bidirectional Long Short-Term Memory (BiLSTM) are best suited to carry out this task. BiLSTM networks are extensions of Recurrent Neural Networks featuring: a layer of hidden units $h_t$ that get activated by input element $x_t$ and by input from unit $h_{t-1}$ preceding $h_t$ (recurrently passing left-context to each activation unit); an equal layer running in opposite direction, recurrently passing right context; gating mechanisms that allow to control how far elements in the sequence affect the activation of each hidden unit, enabling to model long-range linguistic dependencies in a sentence.

\begin{figure*}[!ht]
 \centering
 \includegraphics[width=0.70\linewidth, height=1.1\textheight, keepaspectratio]{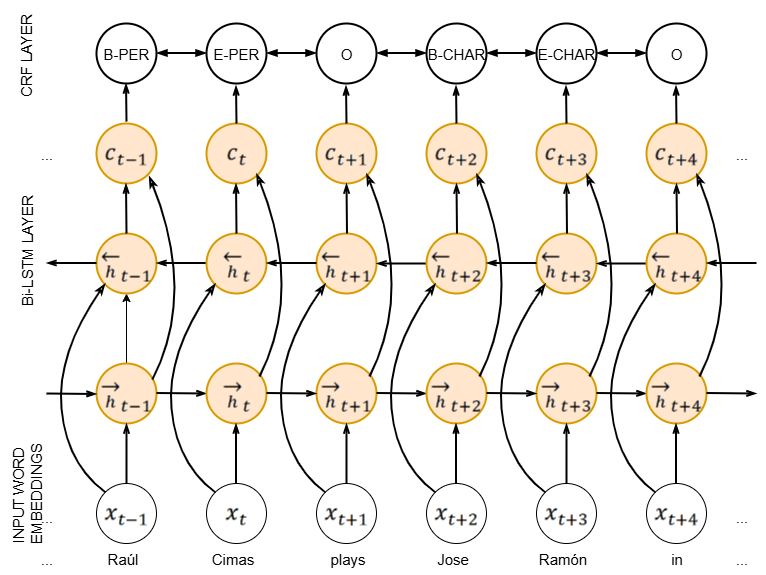}
 \caption{A sample BiLSTM-CRF architecture for NER. Adapted from~\cite{lample-etal-2016-neural}.}
 \label{fig:BLSTMExample}
\end{figure*}

Leaving aside details, Figure~\ref{fig:BLSTMExample} outlines a popular architecture stacking a word embedding vector representation with a BiLSTM layer (in orange) and a sequential conditional random field (CRF,~\cite{10.1561/2200000013})  for generating a sequence of IOBES tags from the input sentence (1) above~\cite{lample-etal-2016-neural}. The BiLSTM is fed a sequence of vectors $x_i$ for each token and builds a representation $c_i$ by concatenating the learned left- and right-context representations $\overrightarrow{h}_i$ and $\overleftarrow{h}_i$. The matrix $P$ of size $n \times k$ (where $k$ is the size of the IOBES tag vocabulary) of scores output by BiLSTM for the sequence of size $n$ is finally modeled jointly with a matrix of tag transition scores via CRF and the log-probability of
the correct tag sequence is maximized during training, generating the tag sequence shown in the top layer in Figure.

\subsection{CNNs for Relation Classification}

In a pipeline approach, once the pair of target entities have been already detected, the RE task boils down to a $n$-class classification problem, with $n$ being the number of possible relation labels.

\begin{figure*}[!ht]
 \centering
 \includegraphics[width=0.75\linewidth, height=1.1\textheight, keepaspectratio]{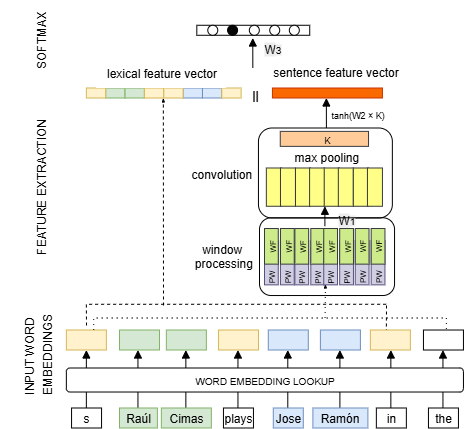}
 \caption{A CNN architecture for relation classification of entity pairs in a sentence. Adapted from~\cite{zeng2014relation}.}
 \label{fig:CNNExample}
\end{figure*}

Instead of relying on a set of lexical and syntactical features extracted by NLP pre-processing modules (typically suffering from error propagation from the deployed tools),~\cite{zeng2014relation} propose to directly feed pre-trained word embedding vectors of the input sentence tokens to a Convolutional Neural Network (CNN) that learns a deep representation of sentence level features of the context of the input entities. As shown in Figure~\ref{fig:CNNExample}, the lexical feature vector is constructed by simply concatenating the embedding vectors of the entity-marked tokens (shown in green and blue at the bottom of the Figure) and of their left- and right-context tokens\footnote{Here a size one context window is illustrated for simplicity.}. Instead, sentence level features are learned by a convolution and max pooling layers, followed by a non-linear layer with $tanh$ activation. The input to this sub-network is computed by a Window Processing module that, for each of the $n$ tokens in the sentence, combines a window of the $w$ embedding vectors around the token with a position feature encoding information on token distance from the two entity tokens. The output of the CNN module is then passed through a non-linear tanh layer and forms a compact representation of the most significant dependencies among the tokens in the sentence.

The lexical and sentence feature vectors are then concatenated into a final feature vector which is finally passed to a softmax layer for predicting the most likely relation label.

\subsection{Joint NER-RE models} 
\label{sec:jointNERRE}
Entity detection and relation classification may benefit from exploiting interrelated signals, for example in the sentence in Figure~\ref{fig:CNNExample} the fact that ``Cimas" is of type \textit{actor} is relevant to the decision whether the predicate ``plays" is an instance of an \textit{acts\_in} relation type, and vice versa. Therefore, recent DL architectures have proven successful at solving the two tasks jointly.

As a notable example, SpERT(Span-based Entity and Relation Transformer)~\cite{eberts2020span}, is a model that reached SOTA performances for some of the RE tasks we will deal with in this thesis. 
It assumes a span-based approach that exhaustively searches every token subsequence of an input text for candidate entities, thus allowing overlapping or nested entities (differently from the IOBES tagging paradigm).

\begin{figure*}[!ht]
 \centering
 \includegraphics[width=0.9\linewidth, height=1.1\textheight, keepaspectratio]{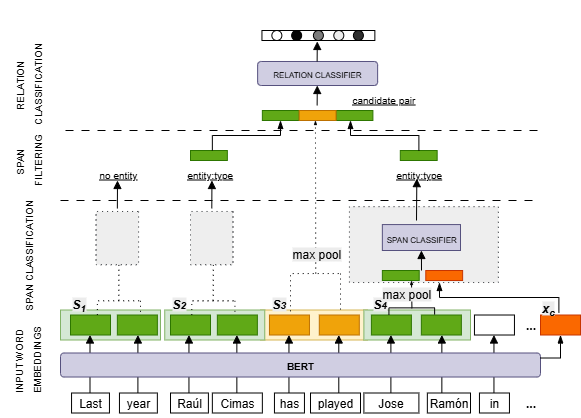}
 \caption{A span-based architecture for joint NER-RE. Adapted from~\cite{eberts2020span}.}
 \label{fig:JointREexample}
\end{figure*}

The relatively simple SpERT's architecture is summarized  Figure~\ref{fig:JointREexample}.

It leverages pre-trained BERT embeddings for representing tokens in the input sentence (plus an additional token $c$ encoding global information about the sentence)\footnote{The BERT embeddings are tuned during the training of the network.}. For each span $s_i$, it fuses its token vectors with max pooling, concatenates them with a context vector $x_c$ capturing global sentence information\footnote{Plus a learned embedding encoding span width, not shown in Figure for simplicity.} and classifies its type with a softmax layer, filtering non-entities (i.e. entities of class \textit{None}). Finally, it classifies the relation holding between all pairs of remaining entities by using a concatenation of the entity-fused BERT embeddings ($s_2$ and $s_4$ in Figure) plus the max-pooled fused embeddings of the token span between the entities (marked in orange in Figure) and applying a single layer sigmoid classifier, such that every relation with an activation score above a confidence score is returned, otherwise no relation is output for the entity pair.  


One major limitation of DL methods is that their performance level depends on the availability of large annotated training sets. Such resources have been built over the years for a number of research competitions, for example SemEval-2010 Task 8~\cite{hendrickx-etal-2010-semeval}, TACRED~\cite{zhang-etal-2017-position} and DOCRed~\cite{yao-etal-2019-docred}. However, they tend to be general or encyclopedic corpora focusing on a limited range of general entity and relation types (e.g. \textit{Person}, \textit{Location}, \textit{Organization}, etc.), languages, and language styles (e.g. English news), which might not fit the specification of one's own application\footnote{Distant supervision has been an explored solution to this issue~\cite{mintz-etal-2009-distant}.}.  

The recent scaling of the Transformer architecture has led to the explosion of Large Language Models (LLMs), which encode extensive general linguistic and world knowledge via pre-training over web scale unannotated text collections, making them robust to unseen domains. Therefore, LLMs represent a powerful solution to the KG construction task, particularly in the low-resource settings we discuss in this thesis.

\section{LLM-based Methods}
\label{llmMethods}

We briefly discussed \textit{self-attention} in  Section~\ref{sec:dlmethods}. Combining a multi-head self-attention with a feed-forward layer makes what is usually referred to as a \textit{transformer block}, which forms the backbone of LLM architectures.
However, it is necessary to distinguish between \textit{encoder} blocks, which can process the entire input sequence in parallel using \textit{bidirectional self-attention} and are used to generate contextual representations of text, and \textit{decoder} blocks,  generating text token by token, using masked self-attention over previously generated tokens only.

Various architectures have been proposed that build upon these processing blocks, all of which strictly speaking fall under the definition of LLMs. For example, the encoder-only model BERT, or the encoder-decoder model T5~\cite{10.5555/3722577.3722647}.

However, in the rest of this thesis, we will use the term LLM to refer to \textit{generative}, decoder-only LLMs (originally named Generative Pre-trained Transformer, or GPT~\cite{Radford2018ImprovingLU}), i.e. models that omit encoder blocks and consist solely of a stack of decoder layers incorporating masked self-attention and feed-forward sub-layers. Rather than processing the input holistically like encoder-decoder models, these models take a token sequence as input and generate a maximum likelihood output token sequence auto-regressively, that is one token at the time and conditioning also on previously predicted tokens\footnote{Maximun probability is in fact only one of various decoding strategies for picking up a token at each generation step.}.

While sacrificing the rich bidirectional understanding of inputs provided by encoder-decoder models, this design choice reduces the number of learnable parameters making generative models extremely scalable for extensive training on sequence generation tasks.


LLMs usually follow a two-step training paradigm:
\begin{itemize}
    \item during initial \textit{language modeling} phase, they undergo a pre-training via self-supervised language masking or word prediction tasks on massive, trillion-token scale text corpora, generating general-purpose \textit{foundational} models, such as Mistral~\cite{jiang2023mistral}, LLaMa 3~\cite{touvron2023llama,huang2024good}, Gemma~\cite{team2024gemma}, and GPT 4.0~\cite{openai2023gpt4};
    \item these base models can be further specialized to downstream tasks either by prompting techniques or via supervised fine-tuning over (typically much smaller) training sets. For example, they can be trained to follow instructions by being presented with pairs of question-response data\footnote{Note that a third standard training phase, called preference tuning, is out of the scope of this thesis.}. In this latter case, the model's weight parameter matrix is (partially) updated.
\end{itemize}

Both foundational and instruction-tuned LLMs have shown strong capacity to carry out standard NLP tasks with near-SOTA performance levels via \textit{in-context learning}, that is by being exposed only to natural language instructions for the target task and optionally a few task solution examples~\cite{NEURIPS2020_1457c0d6}. The instructions used to query an LLM are commonly called \textit{prompts} and prompt design is known to significantly affect LLM performance, depending on its ability to elicit the LLM's vast pre-training knowledge for the new task~\cite{10.1145/3560815}.
In the remaining sections we will briefly introduce some popular prompt engineering techniques, which will be applied in the benchmark evaluation of Chapter~\ref{sec:causalRE}.

Aside from prompt design, LLM output is controlled by a few inference parameters which determine how the model's next token prediction probabilities are applied. They are listed in Table~\ref{tab:hyperparametersInference} in Appendix~\ref{sec:appendixCausalityGraphs}.

\subsection{Instruction Prompting}

Regardless of the target task, a prompt features five basic components:
\begin{enumerate}
    \item \textbf{Role} Prompts of modern LLMs are encoded as sequences of messages, where each message has an assigned role. Role labels are system dependent and reflect the dataset formatting of the specific LLM's instruction tuning, but common role labels are: \textit{system}, for setting background instructions; \textit{user}, encoding the human request and task instruction; \textit{assistant}, for describing the model’s prior or expected output. This is where usually few-shot examples are included.
\item \textbf{Instructions} This contains the task description;
\item \textbf{Data} or input context, that is the actual content on which the task should be applied;
\item \textbf{Output format}, enforcing constraints on how the model's answer should look like;
\item \textbf{Examples} (optional): in few-shot prompts, these are input–output sample pairs that illustrate the task.
\end{enumerate}

Figure~\ref{fig:vanillaPrompt} shows a minimal instruction prompt for RE which directly asks LLMs to extract relation triples from text.

\begin{figure}[!ht]
\begin{tcolorbox}[colback=yellow!5!white,colframe=yellow!50!black,
 colbacktitle=yellow!75!black,fonttitle=\ttfamily\small, title=\textbf{Instruction Prompt}]

\begin{tcolorbox}[
  fontupper=\ttfamily\footnotesize,
  fontlower=\ttfamily\footnotesize,
  sidebyside,
  sidebyside align=top,
  lefthand width=0.7cm,
  frame hidden,
  colback=yellow!5!white,
  boxrule=0mm,
  boxsep=0mm,
  sharp corners,
  colframe=white]

\parbox[t]{0.7cm}{\raggedright\textbf{SYSTEM}:}

\tcblower
You are an Information Extraction assistant.
\end{tcolorbox}

\begin{tcolorbox}[
  fontupper=\ttfamily\small,
  fontlower=\ttfamily\footnotesize,
  sidebyside,
  sidebyside align=top,
  lefthand width=0.7cm,
  frame hidden,
  colback=yellow!5!white,
  boxrule=0mm,
  boxsep=0mm,
  sharp corners,
  colframe=white]

\parbox[t]{0.7cm}{\raggedright\textbf{USER}:}

\tcblower
Given the possible relations:

["acts in", "directs", "directed by", "featuring" ]

What are the relations between the subject entity and the object entity expressed by the sentence?

Provide the output as a triple (head, relation, tail)

\textbf{Sentence}: Raúl Cimas is a Spanish actor, who played Jose Ramón in the comedy series Poquita Fe.

\textbf{Subject}: Raúl Cimas

\textbf{Object}: Poquita Fe

\textbf{Triple}:
\end{tcolorbox}

\end{tcolorbox}
\caption{A baseline instruction prompt for RE.}
\label{fig:vanillaPrompt}
\end{figure}

In Section~\ref{promptDesignSensitivity} we analyze empirically how these five and additional standard prompt design components and features impact LLM performance on a Causal RE task.

\subsection{Few-Shot Learning}

In few-shot learning~\cite{NEURIPS2020_1457c0d6}, the LLM is provided at inference time with a prompt with the following components:
\begin{itemize}
\item \textit{Task Instruction}: A description of the task to be solved.
\item $K$ \textit{Examples of Context-Completion Pairs}, with $K$ typically ranging from 1 to a dozen, depending on the model token size limit.
\item \textit{Input Context}: The specific context for which the model is expected to generate a completion.
\end{itemize}

The context-completion pairs act as a form of conditioning, enabling the pre-trained model to leverage its knowledge for a new task without updating any model parameters~\cite{Radford2019LanguageMA}.

\subsection{Prompt Chaining}
\label{sec:promptChain}
Prompt chaining is a technique that breaks down the instruction prompt of a complex task into a recursive sequence of simpler sub-prompts. Each sub-prompt in the sequence takes as input the outputs of previous sub-prompts in the chain~\cite{sun-etal-2024-prompt}. This approach is motivated by the well-documented observation that single instruction prompts often perform poorly for RE tasks. This is because they require LLMs to handle three non-trivial reasoning processes in a single step: i) Extracting the semantic relationship between the subject and object entities in the text, ii) understanding the semantics of the relation labels, and iii) matching the extracted relationship semantics to the appropriate relation labels~\cite{jimenez-gutierrez-etal-2022-thinking,li-etal-2023-revisiting-large}.

\subsection{Chain-of-Thought}
\label{sec:CoT}

Similar to prompt chaining, \textcolor{black}{Chain-of-Thought (CoT)~\cite{kojima2023largelanguagemodelszeroshot}} is a prompt engineering technique that has proven effective in eliciting structured reasoning from LLMs for solving complex problems. Unlike prompt chaining, which involves multiple generations and passing intermediate results, CoT guides the model through the reasoning steps (or ``thoughts'') in a single prompt. This can be achieved either by providing reasoning examples (few-shot CoT) or by explicitly instructing the model to reason step-by-step (zero-shot CoT).

\subsection{Instruction Fine-Tuning}

\begin{figure}[!ht]
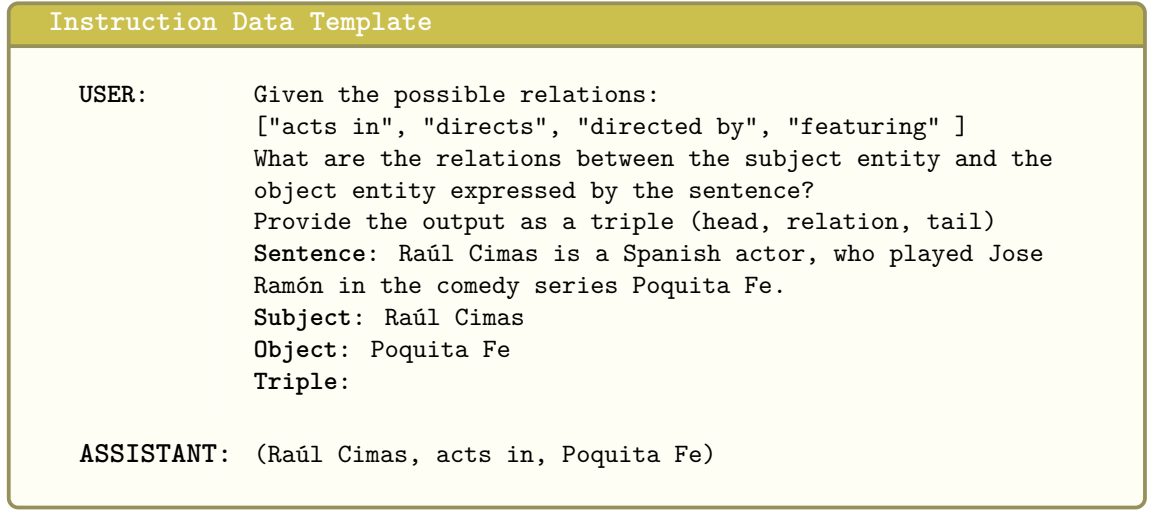

\begin{tcolorbox}[colback=yellow!5!white,colframe=yellow!50!black,
 colbacktitle=yellow!75!black,fonttitle=\ttfamily\small, title=\textbf{Instruction Data Template}]

\begin{tcolorbox}[
  fontupper=\ttfamily\footnotesize,
  fontlower=\ttfamily\footnotesize,
  sidebyside,
  sidebyside align=top,
  lefthand width=1.3cm,
  frame hidden,
  colback=yellow!5!white,
  boxrule=0mm,
  boxsep=0mm,
  sharp corners,
  colframe=white]

\parbox[t]{1.3cm}{\raggedright\textbf{USER}:}

\tcblower
Given the possible relations:

["acts in", "directs", "directed by", "featuring" ]

What are the relations between the subject entity and the object entity expressed by the sentence?

Provide the output as a triple (head, relation, tail)

\textbf{Sentence}: Raúl Cimas is a Spanish actor, who played Jose Ramón in the comedy series Poquita Fe.

\textbf{Subject}: Raúl Cimas

\textbf{Object}: Poquita Fe

\textbf{Triple}:\end{tcolorbox}

\begin{tcolorbox}[
  fontupper=\ttfamily\small,
  fontlower=\ttfamily\footnotesize,
  sidebyside,
  sidebyside align=top,
  lefthand width=1.3cm,
  frame hidden,
  colback=yellow!5!white,
  boxrule=0mm,
  boxsep=0mm,
  sharp corners,
  colframe=white]

\parbox[t]{1.3cm}{\raggedright\textbf{ASSISTANT}:}

\tcblower
(Raúl Cimas, acts in, Poquita Fe)

\end{tcolorbox}

\end{tcolorbox}
\caption{Schema of a training data template for instruction tuning on a RE task.}
\label{fig:vanillaDataTemplate}
\end{figure}

Pre-trained, foundational models typically excel at text completion tasks, but are not necessarily able to follow instructions for an user-defined task. 
A potential solution would be to apply standard backpropagation to optimize the entire weight matrix of the model to the desired task, training it on a (typically much smaller) labeled dataset of instruction-response pairs. For example, for the base RE instruction prompt of Figure~\ref{fig:vanillaPrompt}, one such pair might be encoded in a template like the one shown in Figure~\ref{fig:vanillaDataTemplate}.

However, this full fine-tuning approach is rarely applied in low-resource settings for its heavy computational cost and because it has a ``catastrofic forgetting" side-effect, such that the model loses much of its previously acquired general knowledge from pretraining after being fine-tuned on a specific downstream task.

Instead, PEFT (Parameter-Efficient Fine-Tuning) methods adapt LLMs to downstream tasks or domains by modifying only a small subset of parameters (typically less than $1\%$ of the total parameter matrix), called \textit{adapters}, while keeping the majority of the model weights frozen. In particular, in Chapter~\ref{sec:causalRE} we will apply the Low-Rank Adaptation (LoRA) PEFT technique~\cite{Yu_2023}. For each transformer block weight matrix\footnote{Notice that each of these weight matrices is roughly on the order of $1000 \times 1000$ in size, depending on the model.} $W \in \mathbb{R}^{d \times k}$, LORA approximates it with a linear combination $W + \Delta W$ and then decomposes the update matrix $\Delta W$ into two low-rank matrices:

\begin{equation}
    \Delta W = BA
\end{equation}
with $A \in \mathbb{R}^{r \times k}$, $B \in \mathbb{R}^{d \times r}$ and $r \ll min(d,k) \ $.

Now the forward pass computation within the transformer block is:
\begin{equation}
    h=Wx +\alpha BAx
\end{equation}

but $W$ is kept frozen during training, while the factorized weight matrices $A$ and $B$ that are updated have a total size of $r(d+k)$, which is much lower than $d \times k$.

By training only lightweight low-rank updates to specific weight matrices (e.g., attention projections), LORA dramatically reduces training cost while keeping inference efficient, making it the most popular approach for domain adaptation of LLMs.

\chapter{Digital Transformation Monitoring}
\label{sec:digitalTransformation}
\section{The Process of Digital Transformation}

Digital Transformation (DT) is recognized as ``an holistic reconfiguration of organizational strategies, processes, and culture enabled by digital technologies, with the aim of creating new forms of value and ensuring long-term adaptability"~\cite{VIAL2019118,10.25300/MISQ/2013/37:2.3}. DT is made possible by ``core enabling technologies”,  an evolving set of digital tools and infrastructures that act by reshaping processes, customer experiences, and business models. These include; Cloud Computing, Big Data and Advanced Analytics, allowing the collection and processing of massive and heterogeneous datasets as well as predictive modeling and data data-driven decision making; Artificial Intelligence (AI) and Machine Learning (ML); Internet of Things (IoT), networked sensors and devices generating real-time data; Robotic Process Automation (RPA); Blockchain and Distributed ledger Technologies, ensuring trust, traceability, and transparency in digital transactions. However, unlike mere digitization and digitalization, which focus on technological conversion and enhancement of existing processes via automation, DT implies fundamental organizational change and innovation in business models(~\cite{westerman2011digital}.

Consequently, monitoring the process of DT involves tracking an  heterogeneous range of domain entities from both
scholarly and industrial publications (scientific papers, patents) as well as in the fast-reactive news and social media, tracking concepts like computational methods, algorithms, infrastructures and platforms as well as key players as varied as researchers, innovators, academic institutions, industry and financial corporations. This vast set of domain entities is interconnected by an heterogeneous network of semantic relations, including software processes like method implementation, customization, model training and deployment, as well as managerial and financial activities like technology adoption, company acquisition and merging.    
 
\subsection{Monitoring Unconventional Sources}

The European Commission's Competence Center on Composite Indicators and Scoreboards\footnote{\url{https://composite-indicators.jrc.ec.europa.eu/}.} at the Joint Research Centre (JRC)\footnote{The Joint Research Centre (JRC) of the European Commission (EC): \url{https://ec.europa.eu/info/departments/joint-research-centre_en}.} is carrying out research activities aimed to track societal and economic activities in European countries using unconventional data~\cite{Colagrossi2022323}. In particular, they explore the application of data-driven and AI modeling to the creation of tools assisting investors in decision-making and policymakers in creating policy interventions, assessing their potential to boost economic growth and enhance societal well-being.
Applying such technologies to social media and news has a great potential for forecasting and nowcasting methods, since they provide a larger set of information than standard, lower-frequency socio-economic indicators~\cite{barbaglia2022,consoli2022}.

DT is an ideal target for this endeavor, both for its disruptive change potential in the EU socio-economic ecosystem and because the discourse on DT is pervasive through the news and crowdsourced content platforms such as social media. Therefore, we have designed, implemented and evaluated two prototype pipelines contributing to a under-development DT monitoring system from alternative sources, namely:
\begin{itemize}
\item \textit{Tripl\'{e}toile}, a pipeline for the extraction of a knowledge graph of open-domain entities from micro-blogging posts on the social media platform \textit{X}\footnote{\url{https://x.com/}} (formerly Twitter);
\item an enhanced architecture for the extraction of a knowledge graph of open-domain entities from news articles about digital health technology
\end{itemize}

\subsection{Challenges}

Examining, connecting, and understanding content sourced from microblogging platforms presents several challenges, particularly demanding due to the Internet's diverse array of social platforms, featuring natural language text in varying formats, structures, and lengths.

Social media analysts and various stakeholders commonly navigate them via aggregation tools such as Hootsuite\footnote{\url{https://www.hootsuite.com/}}, Brandwatch\footnote{\url{https://www.brandwatch.com/}}, Talkwalker\footnote{\url{https://www.talkwalker.com/}}, Sprout Social\footnote{\url{https://sproutsocial.com/}}. However, these platforms are constrained to basic queries and merely provide a list of pertinent documents that require manual analysis, while not supporting advanced queries regarding the entities mentioned in the posts.

In order to enable the detection and tracking of potential trends, gauge the influence of events or individuals, and understand their relationships, the research community has put forth numerous proposals aimed at generating organized, interconnected, and machine-readable data frameworks of social analysis knowledge found within text from microblogging platforms, typically using KG technologies~\cite{7764416,DORPINGHAUS2022100337,10.1145/3366424.3383112}. Nonetheless, creating extensive and high-quality KGs from social media is a current open problem. Support tools that aid social media experts in structuring their knowledge (~\cite{webmedia_estendido}) represent poorly scalable solutions, while information extraction (IE) approaches~\cite{DORPINGHAUS2022100337,10.3233/SW-160240,MARTINEZRODRIGUEZ2018339} have the potential for scalability but often struggle to generate outputs of sufficient quality for practical applications~\cite{DESSI2021253}.

Instead, crafting a large-scale, coherent, and semantically sound representation of social media texts drawn from millions of posts, involves addressing at least the following challenges:
\begin{itemize}
\item integrating the extracted information from various posts into a cohesive representation, merging operations of entities and relations via linking to external knowledge bases;
\item defining a flexible ontological framework to formalize a range of statements originating from social media posts
\item estimating the validity of the resultant triples and its correlation with triple support from text sources;
\end{itemize}

In order to address these issues, we designed two scalable and flexible architectures for triple extraction from social media and news text. The proposed pipelines, based on open IE paradigm, support the detection and merging of entity instances matched in text as well as the generalization of various relationships among these entities by using hierarchical clustering, word embeddings, and dimensionality reduction techniques. The manual evaluation we conduct on the triple sets generated by the pipelines reveal that they outperform alternative methods in terms of accuracy, while at the same time generating a relatively higher number of triples.

\section{Data Collections}

\subsection{Social Media}

\begin{figure}
\centering
\includegraphics[width=0.9\textwidth]{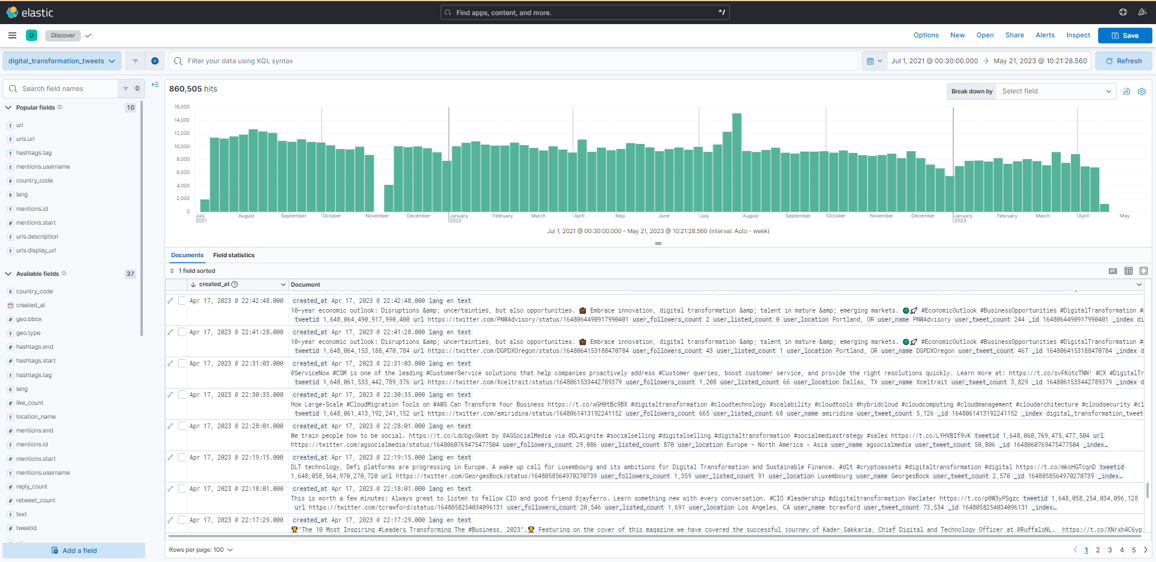} 
\caption{Snapshot of a Kibana dashboard visualization of the tweet collection Elastic Search index.}
\label{fig:esDataLake}
\end{figure}

For our experiments on DT monitoring from social media, we collected a topic-specific dataset of tweets by using the (now discontinued) Twitter public API v2 full-archive search endpoint\footnote{\url{https://developer.x.com/en/docs/x-api/early-access}}. Namely, we retrieved English language tweets from 2022 containing the hashtag \#DigitalTransformation, removing all retweets. We store the resulting corpus of approximately 4M tweets in an Elastic Search index (shown in Figure~\ref{fig:esDataLake}), keeping tweet metadata and tweet ids, for linking back from the extracted triples.

From the stored collection, we sampled a dataset of around 100k, after removal of duplicates and near-duplicates\footnote{Namely  tweets over a 0.85 Levenshtein string similarity threshold, computed after applying the preprocessing described in Section~\ref{preprocess}.}. This is the input data to our social media graph generation pipeline.

\subsection{News}
The news analysis pipeline is applied to a topic-specific news dataset reporting updates on different aspects of the Digital Health domain.

The initial dataset comprises around 7.8 million English-language news articles gathered from the Dow Jones Data, News, and Analytics (DNA) platform\footnote{\url{https://professional.dowjones.com/developer-platform/}}, covering the time frame September 1987 through December 2023 and originating from diverse global English-language outlets, such as The Wall Street Journal, the New York Times, and The Guardian.

In addition to the basic article data such as title, full text,  publication date, etc., DNA provides a range of curated content-based descriptors that are useful for filtering along specific dimensions. These include an 8-level taxonomy comprising approximately nine hundred Subject codes; a 7-level industry code taxonomy, and a set of Region codes encompassing all countries and regions mentioned in the news items.

We started by discarding spurious news items\footnote{Articles with missing titles or with text body character length lower than 300.} and by filtering and merging region codes, ending up with a two-valued (\textit{Europe/US}) macro-area attribute.

We then tested for various combinations of DNA metadata tags as a means to collect a representative sample of news articles about Digital Health technologies. However, health-related Subject tags fall short of retrieving financial/market news updates involving health tech key players, while DNA's Industries classification schema is too coarse-grained to capture emerging technologies and products in this domain. Therefore, we opted for using a trained Deep Learning binary classifier to this purpose.

\paragraph{Topic Classifier}
We fine-tuned the BERT (Bidirectional Encoder Representations from Transformers)\footnote{\url{https://tfhub.dev/tensorflow/small\_bert/bert\_en\_uncased\_L-2\_H-128\_A-2/1} } language model using a near-balanced small set of 9097 news items sampled from DNA and several RSS feeds from specialized news outlets in health tech\footnote{For example, 
 \url{https://www.healthtechdigital.com/}, \url{https://techcrunch.com/tag/healthtech/feed/} \url{https://www.digitalhealth.net/news/}.}. We will refer to positive instances as digital health-related documents, while negative instances will denote non-digital health-related documents.

Out of the 4602 negative instances, 3000 were concatenated title and full text of a sample of 500 items for each set of `negative' topic codes\footnote{Namely, \textit{gcat} (Political/General News), \textit{mcat} (Commodity/Financial Market News), \textit{ccat}(Corporate/Industrial News), \textit{ecat}(Economic News), \textit{gent}(Arts/Entertainment), \textit{gcrim}(Crime/Legal Action).} and 1602 were title and full text of articles scraped from `negative' topic feeds of technology news outlets\footnote{For example, \url{https://techcrunch.com/tag/security/}.}. As for the 4495 positive instances, 4187 consisted of articles from the health tech news outlets mentioned above, while positive instances from DNA were sampled by filtering for health-related Subject codes and manually checking the results, ending up with a subset of 308 health tech items.

The textual data underwent preprocessing, which involved the removal of URLs, all-numeric tokens, and DNA and news outlet-specific tokens (e.g., ``Reuters", ``Reuters Limited", ``techcrunch"). Additionally, all texts were truncated to 1000 characters to eliminate any correlation between the topic and text length features of the article sources.

We then performed fine-tuning using 10-fold stratified cross-validation with 80-20\% data splits and Binary Cross Entropy as Loss function, training for 10 epochs with Early Stopping on 1 epoch of non-increasing Accuracy score.
To mitigate over-fitting on the relatively small training set we kept the model size small (4.3M trainable parameters) and added a dropout regularization layer in the training phase (0.2 dropout rate).

The model reaches average cross-validation F1 score of 98.6\%. Moreover, on an additional, hold-out test set comprising 100 negative and 100 positive instances, sourced from DNA using Subject codes filtering, the classifier achieves 98.8\% F1 score, with the Recall falling short of the Precision (93\% and 98.9\%, respectively). The results indicate that, while missing a high number of positive instances, the model is able to sample a consistent subset of relevant health tech articles from the DNA multi-domain corpus. Therefore, we deploy it on the entire DNA dataset to achieve an overall set of 
97k health tech articles (1.2\% of the entire DNA input data) for our further analysis. 
The model, after training on the entire train set, has been made publicly available at the project repository\footnote{
\url{https://github.com/zavavan/dtm_kg/tree/master/data-collection/dna/bert_fine_tuned_healthTech}
}.

\section{Architectures}
\label{architectures}

\begin{figure}
\centering
\includegraphics[width=1.0\textwidth]{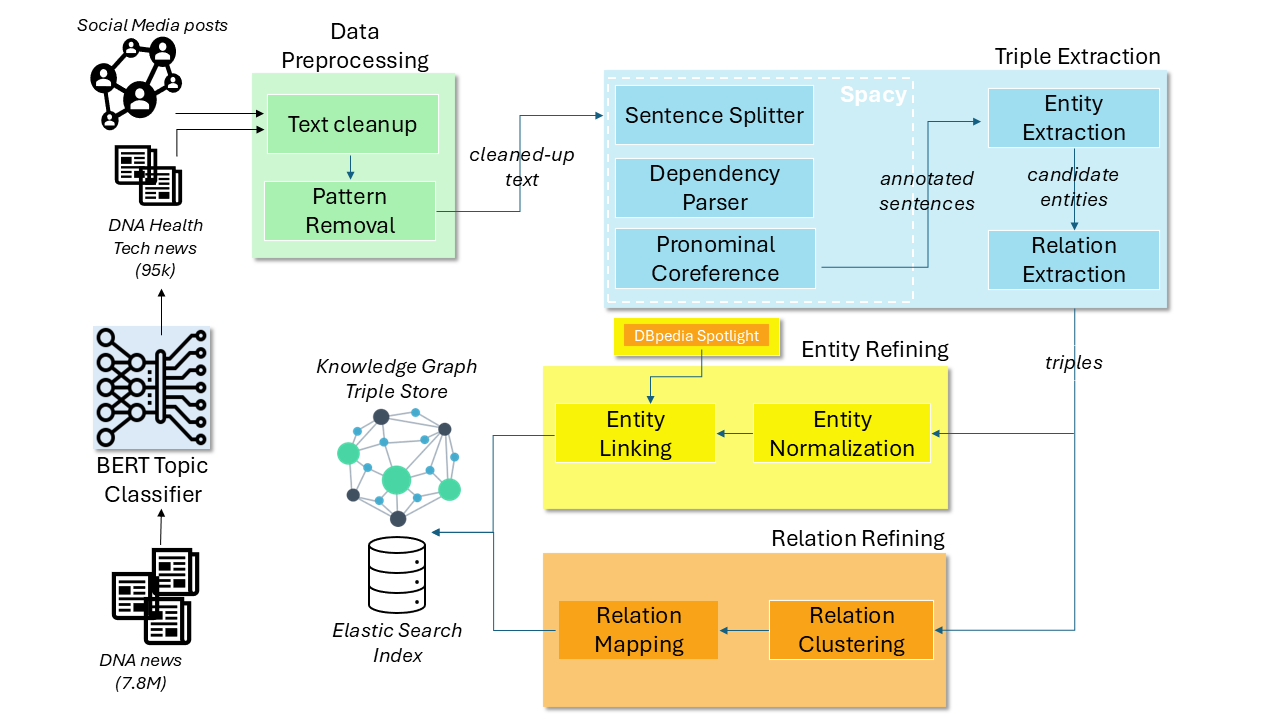} 
\caption{Merged flowchart of the pipelines for knowledge graph generation from micro-blogging and news wire text sources.}
\label{fig:mergedArchitectures}
\end{figure}

Figure~\ref{fig:mergedArchitectures} shows a merged workflow of the two proposed architectures for KG generation. The information extraction architectures consist of customized NLP pipelines built using the spaCy libraries~\cite{spacy2}\footnote{\url{https://github.com/explosion/spacy-models/releases/tag/en\_core\_web\_lg-3.6.0}.} coupled with a series of novel Entity and Relation processing modules.

The two pipelines share the same core modules, with the news-based pipeline adding a pre-filtering step based on the deep learning-based topic classifier described above.

The main blocks of the KG generation architecture include:
\begin{itemize}
 \item \emph{Data Preprocessing}, a step responsible for the normalization of the micro-blogging text in order to make it processable by the downstream text analysis modules;
 
 \item \emph{Triple Extraction}, a block comprising core modules applying text processing libraries and models for the extraction of candidate entity-relation triples. It generates a set $E = {e_0,...,e_n}$ of non-unified, candidate entity phrases, a set of verbal relations $V = {v_0,...,v_k}$ and a set of triples $S = {s_0,...,s_k}$ in the form $<e_{m},v_i,e_{n}>$ where $v_i\in V$ and $e_{m},e_{n}\in E$.
 
 \item \emph{Entity Refining}, a block responsible for the cleaning and generalization of entity mentions to canonical forms, in view of subsequent entity merging;
 \item \emph{Relation Clustering}, in which relation instance verbal forms are mapped to canonical forms, computed as a representative element of the relation cluster they belong to;
\end{itemize}
The final objective of the pipeline is to enable the generalization from the surface form triples in set $S$ to a smaller set $T = {t_0,...,t_h}$ of triples in the form of $<\epsilon_{m},r,\epsilon_{n}>$, where each $\epsilon_{i} \in E$ represents a unified entity and $r$ is a label drawn from a generalized relation vocabulary $R$. In other terms, the final output is a knowledge graph of generalized triples annotated with references to the micro-blogging/news text items they were matched in.

The following subsections describe in more detail the individual components of the pipeline across the four main blocks and how they are applied. The code repositories for the two pipelines are referenced in the Appendix~\ref{sec:appendixDigitalTransforamation}.

\section{Text Pre-processing}
\label{preprocess}

For news data we are able to perform only minimal normalization\footnote{Basically removing URLs and a list of other news platform-specific token patterns.} since the models we subsequently apply for triple extraction are known to perform with high accuracy on benchmark corpora with comparable characteristics\footnote{For example, the OntoNotes corpus (\url{https://catalog.ldc.upenn.edu/docs/LDC2013T19/OntoNotes-Release-5.0.pdf})}.

Twitter status updates (tweets) instead, feature an informal (often plainly ungrammatical) style and an abundance of platform-specific conventions that are known to be hard to process for standard NLP tools, thus requiring ad-hoc preprocessing. 

We follow a two-fold approach to tweet normalization~\cite{Siddharth_Blessing_Luo_2022} which can be readily extended to normalize social content from other platforms~\cite{10.1115/1.4029562,CHIARELLO2020103299}.

On one hand, we remove tokens and token sequences encoding platform-specific metadata or denoting communicative conventions that (typically) do not carry any syntactic function in the tweet sentence, such as sentiment emoticons and smileys, reserved tokens (e.g., RT for `retweet') and URLs.

\begin{figure}
\centering
\begin{subfigure}{1.0\textwidth}
\centering
\includegraphics[width=1.0\textwidth]{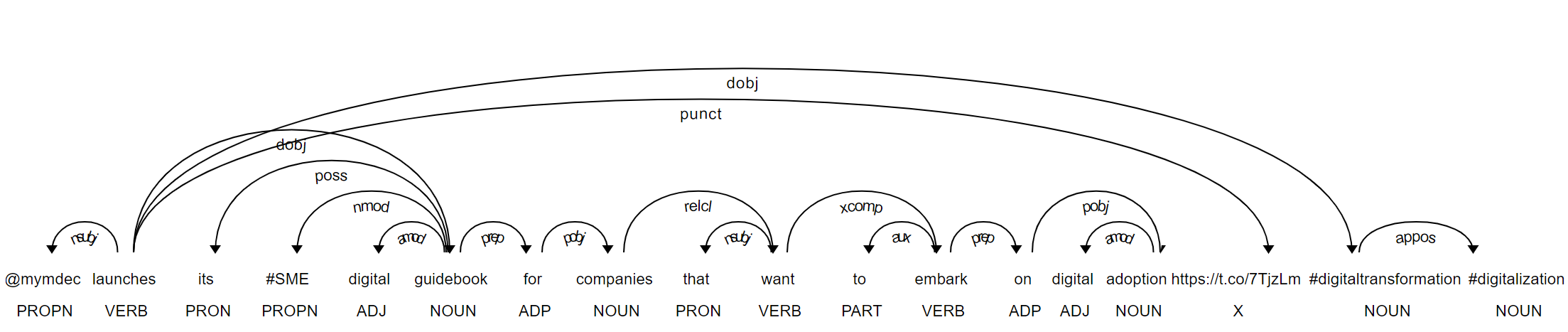}
 \caption{Dependency parse of a tweet's original text.}
 \label{fig:first}
\end{subfigure}
\hfill
\begin{subfigure}{1.0\textwidth}
\centering
\includegraphics[width=1.0\textwidth]{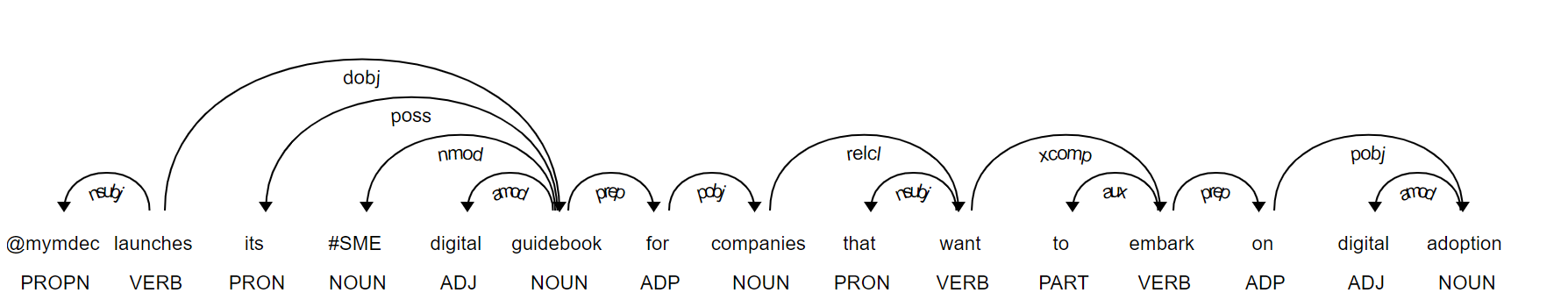}
 \caption{Dependency parse of a tweet after preprocessing}
 \label{fig:second}
\end{subfigure}
\hfill 
\caption{Example of tweet preprocessing.}
\label{fig:tweet preprocessing}
\end{figure}

On the other hand, we keep by default other platform-specific tokens that can carry syntactic functions depending on the context, like hashtags and @ entity mentions (e.g. \textit{\#digitaltransformation, @NASA}). Then, we identify token patterns that typically disrupt the syntactic parsing of the sentence, and remove them from the original tweet. For example, among the preprocessing heuristics:
\begin{enumerate}
 \item we remove sequences of $n$ entity mentions and retweet markers at the beginning of a sentence, for $n > 1$ or when the sequence is not followed by a verb. For example, we remove the leading sequence in \textit{``@bansijpatel @RTatsat @kiranpatel1977 Thanks for updating the information with us.''} but not in \textit{``@AMDRyzen enabling \#DataAnalytics in [...]''}.
 \item for any sequence of size $n > 1$ hashtags/mentions/URL, we drop the sub-sequence with indexes $[1:n]$ or drop the entire sequence if preceded by a sentence closing marker like ('!',':','?','.'). For example, in the text \textit{``According to the @PayNews survey, 84 percent of \#employees in the U.S. have instant access to \#information about their pay and \#benefits \#Sapper \#AI \#hr \#support \#goals[...]''} we keep only the first element of the trailing hash tag sequence. 
\end{enumerate}
Entity mentions and hashtags, that are typically removed from tweet preprocessing pipelines for NLP tasks such as sentiment analysis, are highly relevant for knowledge graph generation as they can be nominal subjects, objects, or modifiers within dependency parse trees and therefore can be extracted as elements of candidate triples, like the tokens @mymdec and \#SME in Figure~\ref{fig:first}. Notice, though, that the trailing sequence of purely referential elements can often lead to noisy edges, for example in the figure the parser wrongly draws a $dobj$ dependency edge from the main verb ``launches'' onto the hashtag \#digitaltransformation.

Figure~\ref{fig:second} shows that the application of the second preprocessing heuristics above can enhance the parsing of the tweet, without losing too much information.
The preprocessing step is carried out using the output of spaCy's English transformer pipeline \textit{en\_core\_web\_trf-3.6.1}\footnote{\url{https://github.com/explosion/spacy-models/releases/tag/en\_core\_web\_trf-3.6.1}}.

\section{Triple Extraction}
\label{tripleExtr}
In the triple extraction block, preprocessed tweets are split into sentences and each sentence is fed to a spaCy pipeline.
Building upon the works in ~\cite{DBLP:conf/semweb/DessiORBM22} and ~\cite{DBLP:journals/kbs/DessiORBM22}, we define a set of procedures to extract candidate nominal entities and predicative triples connecting them, out of dependency parse trees computed by spaCy models.

\subsection{Entity Extraction}
\label{entityExtr}

\begin{figure}
\centering
\includegraphics[width=1.0\textwidth]{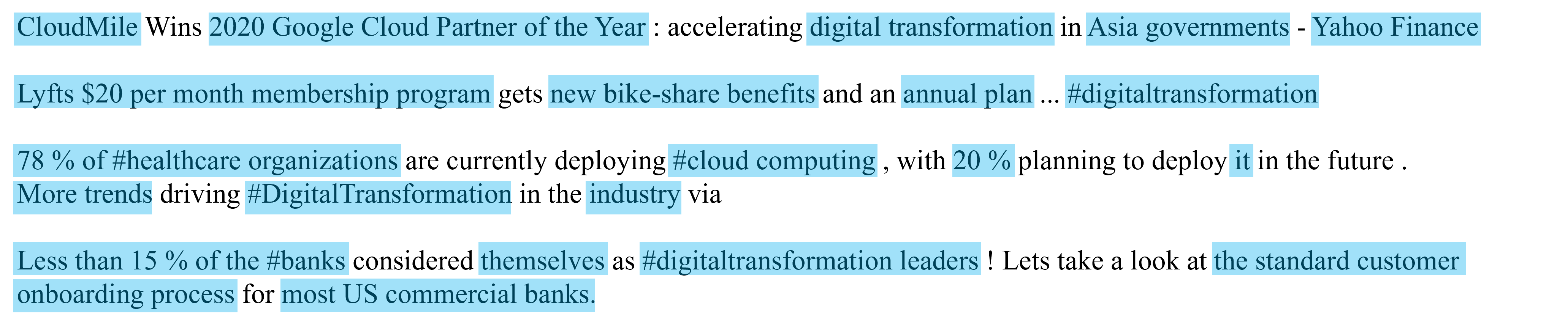} 
\caption{Visualization of candidate entities (highlighted in blue) extracted from a few sample tweets.}
\label{fig:SampleEntities}
\end{figure}

The entity extraction module detects local nominal phrases with a restricted range of syntactic modifications (e.g., compound nouns and adjectives). Then it connects and expands them with: a) a non-recursive set of attached prepositional phrases; b) spaCy quantity-type entities, such as \textit{money}, \textit{percent}, \textit{quantity} and \textit{cardinal}.
We also use pronominal anaphora links, output by the spaCy module coreferee\footnote{\url{https://github.com/richardpaulhudson/coreferee}} replacing them with the expanded entity spans of the tokens they reference. In Figure~\ref{fig:SampleEntities} we show a sample of extracted candidate entities.

For multi-token entity spans including quantifying modifiers (e.g. \textit{`Less than 15\% of the \#banks'}) we maintain a structured representation separating the lexical head (\textit{`\#banks'}) from the quantifying modification of the noun phrase (\textit{`Less than 15\%'}), which then allows a more accurate entity normalization (see Section~\ref{sec:EntityRefining} below).

For the tweet collection, around $33.9\%$ and $6.44\%$ included hashtags and @ entity mentions, respectively; $3.34\%$ were complex noun phrases with prepositional attachments while around $16.6\%$ contained quantitative modifiers of any type (currency, percent, etc.). Out of all the generated triples, a $5.98\%$ had either the subject or object entity consisting of a resolved pronominal anaphora.

Notice that at this stage the hashtag ``\#digitaltransformation" in the second sentence and the noun phrase ``digital transformation" in the first are not mapped to the same general concept \textit{digital transformation} yet, so that the triples in which they occur would still be considered as unrelated.

\subsection{Relation Extraction}
\label{sec:relExtraction}
In the relation extraction module, for each sentence $s_i$ all the shortest dependency paths (see Section~\ref{subsec:nlpMethods}) of the dependency tree between each pair of entities $(e_{m},e_{n})$ containing a verb and matching any of the patterns listed in Table~\ref{TargetPaths} below are selected. 

\begin{table}
\centering
 \begin{center}
 \begin{small}
 \begin{tabular}{|lc|} 
 \hline
 \multicolumn{2}{|c|}{Target dependency paths}\\\hline
 $[nsubj,dobj]$ & \\
$[acl,relcl,dobj]$ &\\
$[acl,dobj]$ &\\
$[nsubjpass,agent,pobj]$ &\\
$[nsubj,dobj,conj]$ &\\
$[nsubj,conj]$ &\\\hline
 \multicolumn{2}{|c|}{Sample discarded paths}\\\hline
 $[obj,pobj]$ &\\
 $[obl,pobj]$ &\\
 $[nsubj;pobj;nmod]$ &\\ 
 \hline
 \end{tabular}
 \end{small}
 \end{center}
 \vspace{-0.4cm}
 \caption{List of target and some of the discarded relation dependency paths.} 
 \label{TargetPaths}
\end{table}

The deployed path set has been selected through an expert validation process, carried out on an open-domain text corpus. 
This process consisted of the collection of the twenty most frequent paths, among all shortest paths connecting any pair of (automatically detected) entities in the corpus. Then, three independent evaluators assessed the correctness of a random sample of 20 triples generated by each of these top frequency paths, majority vote was used to label each triple as correct/incorrect and only the subset of paths with a prevalence of correct triples (i.e., more than 10) were considered reliable and added to the list\footnote{The resource can be found at the URL 
\url{https://github.com/zavavan/dtm_kg/blob/master/resources/paths.txt}}.    

However, we found that in our tweet collection the extraction of triples via the dependency path $[acl,dobj]$ was a potential source of noise, in cases where the noun's clausal modifier was an infinitive verb, as exemplified in the following sentence:

\textit{``Salesforce really has the power to transform your business."}

that incorrectly generates a triple such as $<power,transform,business>$. Consequently, we added a constraint to the dependency path $[acl,dobj]$ in order to filter out those paths where verb nodes had a relation \textit{aux} with an infinitive particle node\footnote{In the example above, \textit{transform} has an \textit{aux} relation to the particle \textit{to} and, therefore, it is discarded. More precisely, the following expressions hold:

\textbf{\url{SUBJ}}=\textit{power} $\xrightarrow{\text{acl}}$ \textbf{\url{PRED}}=\textit{transform} $\xrightarrow{\text{dobj}}$ \textbf{\url{OBJ}}=\textit{your business}

\textbf{\url{PRED}}=\textit{transform} $\xrightarrow{\text{aux}}$ \textbf{\url{to}}.
}.

Analogously to what pointed out for the entities, note that the $v$ in $V$ are surface forms, that is individual inflected verbal phrases that do not enable as such to generalize triples over morphological or lexical variations. For example, the following triples:

$<BLEND360,acquires,Engagement Factory>$

$<BLEND360,acquired,Engagement Factory>$

$<BLEND360,bought,Engagement Factory>$ 

are considered distinct facts at this stage.


\section{Entity Refining}
\label{sec:EntityRefining}

The function of this module is to clean up and normalize the candidate entities into a normalized form that allows the merging across entity name variants\footnote{Splitting is another typical subtask of Entity Refining functions, for example by separating the individual entities in parsed coordinated noun phrases like in \textit{`\#testautomation and \#datamanagement can accelerate your \#digitaltransformation'}. However, we deal with these cases earlier on at the triple extraction phase by generating a triple for each coordinated entity.}.

Entities are first cleaned up by removing leading/trailing punctuation marks as well as stop-words.
Afterwords, we distinguish the following cases for normalization:
\begin{itemize}
 \item \textcolor{black}{For hashtags and @ mentions, we remove hashtags and @ symbols, split the ``camel case'' forms (e.g., \textit{\#SmartCities}) and lowercase the resulting string.}
 \item For all other entities, we lemmatize and lowercase all component tokens whose POS tag is neither Verb nor Proper Noun, otherwise we simply lowercase.
 \item For nouns that have variants in American English, we finally map to the British English variant. 
\end{itemize}

We make use of such normalized versions of the candidate entities for merging them, by using the spaCy-integrated DBpedia Spotlight Entity Linking library\footnote{\url{https://spacy.io/universe/project/spacy-dbpedia-spotlight}}. 


The DBpedia Spotlight model is trained to perform both entity detection and linking at once. In order to power this module with the entity normalization, performed by our pipeline, we run the module over modified article sentences where the original subjects and object entity spans are replaced with their normalized forms. Entities that are linked to the same DBpedia entries are merged. More precisely, we link the normalized versions of the entities to the corresponding DBpedia entries of the spaCy native entities whose text spans are both:
\begin{itemize}
 \item included within the subject or object text spans of the corresponding normalized entities;
 \item overlapping with the syntactic heads of the corresponding normalized entities. 
\end{itemize}
In other terms, we let spaCy's DBpedia Spotlight perform the merging of entities that were normalized to the same or similar forms, by linking to the same DBpedia entries.
For example, the two candidate entities \textit{``Gartner"} and \textit{``@Gartner\_inc"} are merged 
together by linking them to the DBpedia entry of the Gartner consulting firm {\url{http://dbpedia.org/resource/Gartner}). 

In case only the first condition is met, we assign a semantic ``relatedness" link between the candidate entity and the DBpedia entry, indicating that the former is not an instance of, but rather is related to the latter\footnote{We keep out the cases when only the second condition is met, as they typically arise from inaccuracies of the entity span detection.}. 
For example, the span \textit{`@gartner\_survey'} is considered only ``related' to the DBpedia entry for Gartner.

In Section~\ref{sec:resultsDTMonitoring} we describe how these relations are encoded in the resulting knowledge graph by inheriting from existing ontology relations.

\section{Relation Refining}
\label{relMapping}

The extracted triples often contain numerous distinct relations that convey similar meanings. To minimize redundancy and support semantic retrieval of the triples in the generated graph, these extracted relations need to be consolidated into a smaller set of predefined relations. This block aims to find the best predicate label $r$ for each relation verb $v$ in a triple $<e_{m},v,e_{n}>$ and to map $v$ to $r$ in the resulting triples.

The approach we followed consisted of deriving a word embedding representation of the verb predicates from a pre-trained model, computing an optimized clustering of the relation vectors, and finally using a representative instance of each cluster to map verb predicates. A similar method, using however a ``bag of words" vector representation, was proposed by~\cite{10.3115/1218955.1219008}.

After experimenting with several (contextual and non-contextual) word embedding models and clustering algorithms, we converged to a setup using static word embeddings learned with GloVe (see Section~\ref{sec:clustering}) and applying HDBSCAN clustering to the vectors. We tested using verb phrase contextual embeddings from Huggingface's \textit{bert-large-uncased}\footnote{https://huggingface.co/bert-large-uncased} and Sentence-BERT\footnote{\url{https://sbert.net/}}. However, it turned out that the optimal cluster scores, in this case, were achieved for a number of clusters too close to the number of items in the dataset\footnote{In other terms, these representations were not suitable for generalizing enough over relations, probably due to the context-specific information they are encoding.}. Table~\ref{alternativeEmbeddings} in Appendix~\ref{sec:appendixDigitalTransforamation} reports the clustering scores and number of resulting clusters for some best performing configurations using the different embedding models, for the tweet collection.

\paragraph{Relation Embeddings}
For each single or multi-token relation predicate, we get the static, 300-dimensional word embedding vector made available for text span objects in the spaCy \textit{en\_core\_web\_lg-3.6.0} pipeline\footnote{\url{https://github.com/explosion/spacy-models/releases/tag/en\_core\_web\_lg-3.6.0}}.
 
\paragraph{Dimensionality Reduction and Clustering}
We used the HDBSCAN clustering algorithm enhanced by previously applying the UMAP dimension reduction technique on the word embeddings vectors (see Section~\ref{sec:clustering}). 

In order to optimize the combination of UMAP and HDBSCAN, we perform a grid search over the hyperparameters of both algorithms and evaluate the clustering using the score indicated in Equation~\ref{eq1}:
\textcolor{black}{
\begin{equation}
S = {silhouette_X}\cdot{clustered_X},
\label{eq1}
\end{equation}
}

where the silhouette coefficient $silhouette_x$ of an instance $x \in X$ is defined in Equation~\ref{eq2}:

\begin{equation}
(b – a) / max(a, b),
\label{eq2}
\end{equation}

with $a$ being the mean distance to the other instances in the same cluster and $b$ being the mean distance to the instances of the next closest cluster. 

In the $S$ score formula, the $silhouette_X$ is the mean silhouette coefficient over all the instances of the dataset $X$ that were  clustered by HDBSCAN~\cite{Batool2021} 
while $clustered_X$ is the fraction of instances of $X$ that were actually clustered by HDBSCAN. 

In practice, we optimize for the classical measure of cluster cohesion and separation while penalizing the configurations with low coverage of the dataset.
We finally chose a subset of best-scoring hyperparameter configurations and plotted their $S$ score over the number of output clusters they generate, so that we are able to pick a sub-optimal configuration that strikes a balance between generalization (fewer clusters) and accuracy (cluster number closer to the dataset size).

In our experiments with the \textit{X/Twitter} data, we started with a set of 29,335 raw triples\footnote{These are surface-level candidate triples from the Triple Extractor, counted prior to entity and relation merging.} from which we computed and standardized 2,539 unique 300-dimensional word embeddings from GloVe. Figure~\ref{fig:plotClusterNumScore} reports the $S$ score over the number of resulting clusters for a subset of best-scoring UMAP-HDBSCAN configurations in the grid search. Highest score values are initially reached for configurations sensitive to the global structure of the dataset (very few clusters, very coarse-grained generalization), while they subsequently tend to grow with the increasing number of clusters until they flatten again. We picked up a sub-optimal configuration with an overall score of around 0.65, silhouette score on clustered points 0.73 and data clustering percentage 0.89, returning 327 clusters, with an average cluster size of 7.6 elements\footnote{This corresponds to an UMAP 2 dimension-reduced representation of the vector dataset, obtained using a $n\_neighbors = 5$ hyper-parameter and to a $min\_cluster\_size=3$ and $min\_samples=1$, for HDBSCAN.}.


\begin{figure}
 \centering
 \includegraphics[width=0.9\textwidth]{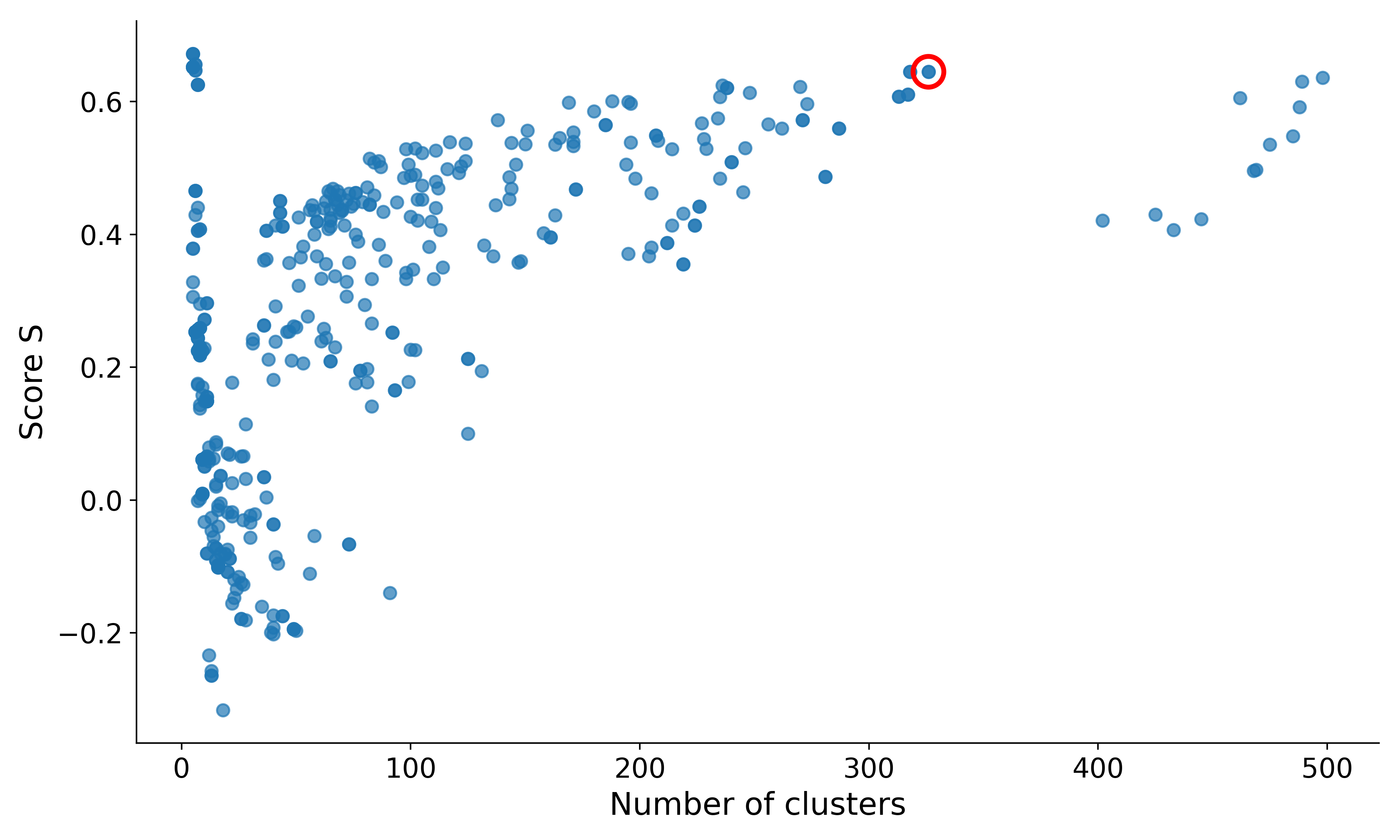}
 \caption{$S$ score over number of generated clusters for a subset of best-scoring UMAP-HDSCAN hyperparameter configurations on GLOVE embeddings of relations, from the tweet collection, with the picked up sub-optimal value circled in red.}
\label{fig:plotClusterNumScore} 
\hfill
\end{figure}

The same procedure was followed for the news dataset, where the chosen configuration reached a $S$ score of 0.62, with $silhouette_X = 0.65$, $clustered_X = 0.92$ and 2 UMAP components.

\paragraph{Relation Mapping}
Finally, for each relation verb $v$ in the dataset, we replace it with the predicate label $r$ consisting of the lemma of the most frequent relation in the cluster of $v$\footnote{\url{https://hdbscan.readthedocs.io/en/latest/api.html}. Notice that as HDBSCAN can generate clusters of arbitrary forms, it does not hold a notion of cluster centroid and there are typically multiple `most representative' data points in a cluster, based on soft clustering.}. Otherwise, we map it to itself if $v$ was an outlier and not clustered. For the news dataset, we slightly modify the heuristics and map each $v$ to a predicate label $r$ consisting of the most frequent lemma among the ``exemplars" relations returned by HDBSCAN for the cluster of $v$.

Thus, the three distinct triples shown in the last example of Section~\ref{sec:relExtraction} would be merged and the resulting triple would be: 

$<BLEND360,BUY,Engagement Factory>$\footnote{A CSV file with a sample of the most frequent normalized triples for the tweet dataset, together with the originally matched relations can be found in the project repository at \url{https://github.com/zavavan/dtm_kg/blob/master/data-collection/twitter/sampleNormalizedTriples.tsv}}.

Table~\ref{RelationMappingSample} shows some sample mappings from relations extracted from the tweet collection to their associated predicate labels, consisting of the lemma of the most frequent relation in their clusters.

\begin{table}
\centering
 \begin{center}
 \begin{small}
 \begin{tabular}{|c|c|p{6cm}|} 
 \hline
 Relation Verb & Relation Predicate & Example \\ \hline
 
 fuel & FUEL & \textit{`How the UR+ Ecosystem is Fueling Cobot Market Growth'}\\
 driven by & FUEL & \textit{`Digital transformation in Ho Chi Minh is being driven by remote working'}\\
 accelerated by & FUEL & \textit{`huge social trends being accelerated by the pandemic.'}\\
 identify & IDENTIFY & \textit{`Machine learning can identify signs of Alzheimers in patients '}\\
quantify & IDENTIFY & \textit{`Research quantifies G's potential in roaming and manufacturing '}\\
predict & IDENTIFY & \textit{`AI-supported test can predict eye disease that leads to blindness'}\\\hline
 \end{tabular}
 \end{small}
 \end{center}
 \vspace{-0.4cm}
 \caption{Sample relation verb-predicate mapping.} 
 \label{RelationMappingSample}
\end{table}


\section{Evaluation}
\label{eval}

We perform a twofold evaluation of our Tripl\'{e}toile pipeline, based on data from the tweet collection: a. we manually assess the truthfulness of a sample set of statements, estimating Precision, Recall and F-measure; b. we evaluate our pipeline Precision against a number of alternative methods.

\paragraph{Human Expert Assessment}
We randomly select 483 statements, equally distributed among high-support (support greater or equal to 5) and low-support triple groups and we have each triple assessed by three evaluators as True or False. The True label was assigned when all the following criteria were satisfied:
\begin{itemize}
\item{the subj and obj entities are linked by a relation in the tweet text;}
\item{the assigned relation label entails the relation verb in the tweet text;}
\item{the spans of the subject/object of extracted triples include the syntactic head of the relation's subject/object\footnote{For example, a triple $<78\%\_of\_\#healthcare,USE,Digital\_Transformation>$ would be marked as False if extracted from the text \textit{``78\% of \#healthcare organisations deploy \#DigitalTransformation"} as the syntactic head is ``organisations".}}.
\end{itemize}

We calculated the average pair-wise Cohen $\kappa$ inter-rater agreement and the Fleiss $\kappa_F$ agreement of all the 3 raters~\cite{Falotico2015463}, resulting in a score of 0.61 and 0.558, respectively. These values generally indicate a substantial level of agreement, although one annotator featured an outlier rating on a specific category of cases.

The majority vote-based assessment of the three annotators yielded a Precision of 0.96. In order to compute Recall, the three annotators were tasked to extract triples that they deemed correct from the same tweets containing the 483 selected statements. They extracted a total of 484 triples from a resulting set of 491 tweets (we considered the union of all the triples extracted by each annotator). Consequently, we were able to calculate the number of true positives ($TP=464$), false positives ($FP=19$), and true negatives ($TN=20$) and compute a Recall of 0.95 and F1 score of 0.95. Individual rater estimates ranged from 0.90 to 0.96. These results indicate that the pipeline can extract triples with good precision from noisy text like tweets, while at the same time missing only a few triples.


\paragraph{Comparative Evaluation}
For a comparative evaluation vis-à-vis alternative triple extraction methods, we randomly sampled 500 tweets from the 100k-sized original dataset and used our pipeline to extract candidate entities. We then merged this set of entities with the one generated by the DyGIEpp Extractor~\cite{wadden-etal-2019-entity}.

DyGIEpp is an joint NER-RE framework, similar to the one outlined in Section~\ref{sec:jointNERRE}
. In order to detect entities, DyGIEpp uses a feed-forward neural network on textual span representations and computes a score for each entity type; an entity is detected considering the highest score for an entity type if a minimum threshold is met.

We then employed four alternative methods to identify relationships between these input entities and to extract triples from the 500 tweets. Specifically, we compared:

\begin{itemize}

 \item \textit{OpenIE Extractor}, the IE tool of the Stanford Core NLP suite, described in Section~\ref{openIE};

 \item \textit{PoST Extractor}, a module built on top of the Stanford Core NLP suite that uses PoS tags to find all verbs that exist between two candidate entities in a sentence to extract verb relations, using a window of max token distance 15 between the entities;

 \item \textit{Dependency-based Extractor}, a module that extracts dependency trees using the dependency parser of the Stanford Core NLP suite, maps entities previously extracted using DyGIEpp into the sentence tokens, and exploits $12$ hand-crafted paths\footnote{\url{https://github.com/danilo-dessi/SKG-pipeline/blob/main/resources/path.txt}} to find verb connecting entities.

 \item \textit{Entity and Relationship Refiner}, a module that applies \textit{Entity and Relationship Handlers} as described in~\cite{DBLP:journals/kbs/DessiORBM22} to the set that includes \textit{OpenIE Extractor}, \textit{PoST Extractor}, and \textit{Dependency-based Extractor} triples. 
 
\end{itemize}

The number of extracted triples from the dataset ranged from 339 for the Dependency-based Extractor to a maximum of 1,015 for the PoST Extractor, which does not impose any filter on the dependency relations connecting verbs that exist between pairs candidate entities. After PoST Extractor, Tripl\'{e}toile is the one generating the largest number of triples (663) among the methods using the extended range of candidate entities, with OpenIE Extractor producing 588 triples Entity and Relationship Refiner reaching 348 triples.
In order to use these numbers as an indirect assessment of the relative Recall levels of the different pipelines, we manually assessed also the Precision on a limited random sample of 150 triples generated by each method, using the same human assessment setup described earlier and reaching a strong $\kappa_F$ agreement coefficient of 0.86\footnote{Notice that these test sets are not generated from the same tweet subset for each pipeline. Notice also that the random sampling was done without using any information on the triple support, which was not available for the alternative pipelines.}. We report the results in Table ~\ref{ComparativeEvaluation}, with the precision of our pipeline largely outperforming all the alternative methods. 

\begin{table}
\centering
 \begin{center}
 \begin{small}
 \begin{tabular}{|c|c|} 
 \hline
 Extraction Method & Precision \\ \hline
 OpenIE Extractor & 0.52\\
 PoST Extractor & 0.17\\
 Dependency-based Extractor & 0.77\\
 Entity and Relationship Refiner & 0.31\\
 Tripl\'{e}toile & \textbf{0.82}\\
\hline
 \end{tabular}
 \end{small}
 \end{center}
 \vspace{-0.4cm}
 \caption{Precision of the triples extracted from a set of alternative methods from a collection of 500 tweets, using a combination of Tripl\'{e}toile and DyGIEpp input entities.} 
 \label{ComparativeEvaluation}
\end{table}

 Overall, we conclude that the proposed method is able to extract semantically accurate triples from noisy text data such as micro-blogging posts, while featuring a competitive recall against other NLP methods. We do not arguably expect a decrease in accuracy decay on the more standard language style of news data, although we lack a dedicated evaluation for it. Therefore, we deploy our method for the generation of large scale Knowledge Graphs from both use case text collections. 

\section{Digital Transformation Social Media Monitor  Knowledge Graph}
\label{sec:resultsDTMonitoring}

The \textit{\textit{DTSMM\_KG}} (Digital Transformation Social Media Monitor Knowledge Graph) comprises approximately 22,270 statements (triples). We represented all statements extracted from the tweets using an \textit{DTSMM\_KG} ontology (with  namespace prefix \textit{dtsmm-ont}) we created for this purpose. Table~\ref{SampleTriplesDT} shows a statement sample.

\begin{table}
\centering
 \begin{center}
 \begin{small}
 \begin{tabular}{c|c|c} 
 \hline
 Subject Entity & Relation & Object Entity \\ \hline
 pandemic & accelerate & digital\_transformation \\
 artificial\_intelligence & impact & insurance\_sector \\
 microsoft & buy & riskiq \\
 data-driven\_insight & drive & decision-making \\ 
agile\_business & demand & effective\_marketing\_capability\\
 hootsuite & buy & ai\_chatbot\_firm \\
 automl & generate & data-driven\_insight \\
 image\_classification & use & transfer\_learning \\
 new\_belgium\_brewing & implement & digital workflow\_place\_solution \\
 e-rupi & back & existing indian rupee \\
82\%\_of\_cio & implement & new\_technology \\
image\_recognition\_framework & use & artificial\_intelligence \\
microinsurance & close & africa\_insurance\_gap\\
hsbc\_qatar & introduce & mobile\_payment \\
ford\_motor\_company & explore & blockchain\_technology \\
\hline
 \end{tabular}
 \end{small}
 \end{center}
 \vspace{-0.4cm}
 \caption{A sample of statements extracted from the tweet collection by the Tripl\'{e}toile pipeline.} 
 \label{SampleTriplesDT}
\end{table}

Each statement is reified into \textit{dtsmm-ont:Statement} class instances, which defines a specific claim extracted from a given number of tweets. Namely, each  \textit{dtsmm-ont:Statement} object includes:
\begin{itemize}
\item the reification of the triple itself via \textit{rdf:subject}, \textit{rdf:predicate} and \textit{rdf:object} predicates;
\item a data property \textit{dtsmm-ont:hasSupport} reporting the number of tweets supporting the claim;
\item a number of object property instances \textit{dtsmm-ont:comesfromTweet} ranging over ontology instances of type \textit{dtsmm-ont:Tweet} (which was inherited from \textit{schema:SocialMediaPosting}) supporting the claim;
\item A boolean data property \textit{dtsmm-ont:negation} flagging whether a negation of the claim's predicate was parsed from the source text. 
\end{itemize}

Figure~\ref{fig:StatementReification} shows a shortened example of a claim reification having the \textit{DTSMM\_KG} ontology's instance \textit{machine\_learning} as \textit{rdf:object} and support equal to six.

\begin{figure}
\begin{tcolorbox}
\begin{footnotesize}
\begin{flushleft}
\begin{verbatim}
dtsmm-ont:statement_10100 a dtsmm-ont:Statement,
 rdf:Statement ;
dtsmm-ont:negation false ;
dtsmm-ont:comesfromTweet dtsmm:tweet_1424266328882429952 ;
...
dtsmm-ont:hasSupport 6 ;
rdf:subject dtsmm:multi_page_document_classification ;
rdf:predicate dtsmm-ont:use ;
rdf:object dtsmm:machine_learning .
\end{verbatim} 
\end{flushleft}
\end{footnotesize}
\end{tcolorbox}
 \caption{A shortened example of reification for a Statement concerning the instance \textit{machine\_learning}, grounded by 6 tweets, with the three dots referring to the hidden \textit{dtsmm-ont:comesfromTweet} predicates.}
 \label{fig:StatementReification}
\end{figure}

Figure~\ref{fig:ML_Subgraph} instead illustrates a sub-graph of \textit{DTSMM\_KG} containing a few sample triples having the instance \textit{machine\_learning} as  subject. Here, we report just the statements, hiding claim reification for the sake of readability.

\begin{figure}
 \centering
 \includegraphics[width=0.9\textwidth]{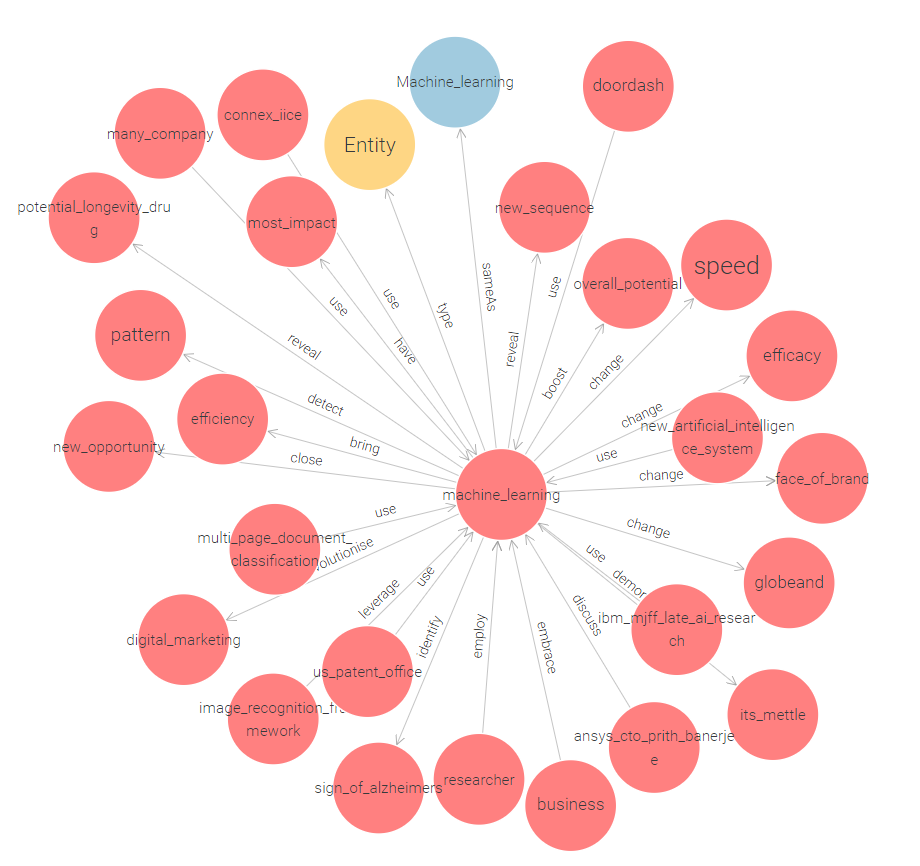}
 \caption{A sample \textit{DTSMM\_KG} subgraph showing a few claims for the instance \textit{machine\_learning}.}
 \label{fig:ML_Subgraph} 
 \hfill
\end{figure}

The linking of the \textit{DTSMM\_KG} instances to the DBpedia ontology (see Section~\ref{sec:EntityRefining}) is implemented by using the \textit{owl:sameAs} predicate, while the ``relatedness" link between candidate entities and the DBpedia entries are encoded using SKOS predicate \textit{skos:related}\footnote{\url{https://www.w3.org/2004/02/skos/}}. Overall, \textit{DTSMM\_KG} provides 2,857 \textit{owl:sameAs} links and 3,309 \textit{skos:related} links to to DBPedia entries. 

The resulting data have been made publicly accessible under Creative Commons Attribution 4.0 International (CC BY 4.0) license\footnote{Creative Commons Attribution 4.0 International license: \url{https://creativecommons.org/licenses/by/4.0/}} within the Joint Research Centre Data Catalogue\footnote{\url{https://data.jrc.ec.europa.eu/dataset/f7be47f7-49a2-44e8-9dc8-043735af4139}}, as well as within the European Data portal\footnote{\url{https://data.europa.eu/88u/dataset/f7be47f7-49a2-44e8-9dc8-043735af4139}}, the official data repository of the European Commission. The direct link to the 
Digital Transformation knowledge graph, available in Terse RDF Triple Language (Turtle), is \url{https://jeodpp.jrc.ec.europa.eu/ftp/jrc-opendata/CC-COIN/se-tracker/DTSMM_KG.ttl}.

\subsection{Knowledge Access Example: Graph RAG application}

Access to \textit{DTSMM\_KG} can be useful in application scenarios where one needs to bridge the gap between the wealth of information available in real-time data streams, like social media, and more static, conventional sources.
For example, it might serve as a critical resource for enriching Graph-based Retrieval-Augmented Generation architectures~\cite{RAGseminal,edge2025localglobalgraphrag}, in order to dynamically pulling in contextual information via KG querying during the generation process, thus enhancing the quality and relevance of the results.

As a simplified Question Answering example, assume one is supplying the following question to a RAG system:

\begin{equation}
\textit{Is Microsoft dedicating resources to computer security technologies?}
\label{eq:ragquestion}
\end{equation}

and assume a NER model is able to recognize the entities \textit{Microsoft} and \textit{Computer Security} from the text. \textit{DTSMM\_KG} can be queried to detect whether \textit{Microsoft} entities are declared into its ontology:

\begin{figure}[hb!]
\begin{flushleft}
\begin{tcolorbox}
\begin{footnotesize}
\begin{verbatim}
SELECT DISTINCT *
FROM <DTSMM_KG> 
WHERE { <http://dtsmmkg.org/dtsmmkg/resource/microsoft> ?p ?o . }
\end{verbatim}
\end{footnotesize}
\end{tcolorbox}
\end{flushleft}
 \label{fig:Query1}
\end{figure}

which would produce the following resulting triple (in RDF Turtle format): 

\begin{figure}[hb!]
\begin{flushleft}
\begin{tcolorbox}
\begin{footnotesize}
\begin{verbatim}
@prefix dtsmm: <http://dtsmmkg.org/dtsmmkg/resource/> .
@prefix dtsmm-ont: <http://dtsmmkg.org/dtsmmkg/ontology#> .
@prefix owl: <http://www.w3.org/2002/07/owl#> .

dtsmm:microsoft a dtsmm-ont:Entity ;
 owl:sameAs <http://dbpedia.org/resource/Microsoft> .
\end{verbatim}
\end{footnotesize}
\end{tcolorbox}
\end{flushleft}
 \label{fig:Query2}
\end{figure}

informing us that the resource \textit{Microsoft} is defined and equal to the well-known DBpedia entity \url{http://dbpedia.org/resource/Microsoft}. 

This would allow to infer additional information, external to our knowledge-base, via the DBpedia SPARQL endpoint\footnote{Available at \url{https://dbpedia.org/sparql}} using the query:

\begin{figure}[hb!]
\begin{flushleft}
\begin{tcolorbox}
\begin{footnotesize}
\begin{verbatim}
SELECT DISTINCT *
FROM <DTSMM_KG> 
WHERE { <http://dbpedia.org/resource/Microsoft> ?p ?o . }
\end{verbatim}
\end{footnotesize}
\end{tcolorbox}
\end{flushleft}
 \label{fig:Query3}
\end{figure}

which returns 960 knowledge triples about Microsoft\footnote{One can see see all the triples by browsing directly DBpedia to the URL \url{http://dbpedia.org/resource/Microsoft}.}.

This existing knowledge can be enriched with new relations extracted from \textit{DTSMM\_KG}. For example, looking for triples involving the subject Microsoft with an ``acquire" predicate type (i.e. \url{http://dtsmmkg.org/dtsmmkg/ontology#acquire}) via the SPARQL query:

\begin{figure}[htp!]
\begin{flushleft}
\begin{tcolorbox}
\begin{footnotesize}
\begin{verbatim}
prefix dtsmm: <http://dtsmmkg.org/dtsmmkg/resource/> 
prefix rdf: <http://www.w3.org/1999/02/22-rdf-syntax-ns#> 

SELECT DISTINCT *
FROM <DTSMM_KG> 
WHERE { 
 ?statement rdf:subject dtsmm:microsoft . 
 ?statement rdf:predicate dtsmm-ont:acquire . 
 ?statement rdf:object ?object . 
}
\end{verbatim}
\end{footnotesize}
\end{tcolorbox}
\end{flushleft}
 \label{fig:Query4}
\end{figure}

we would get to know that Microsoft has acquired companies like \textit{Cloudknox\_Security}, \textit{CyberX} and \textit{RiskIQ}. In SPARQL we might then ask for information about this last: 


\begin{figure}[htp!]
\begin{flushleft}
\begin{tcolorbox}
\begin{footnotesize}
\begin{verbatim}
SELECT DISTINCT *
FROM <DTSMM_KG> 
WHERE { <http://dtsmmkg.org/dtsmmkg/resource/riskiq> ?p ?o . } 
\end{verbatim}
\end{footnotesize}
\end{tcolorbox}
\end{flushleft}
 \label{fig:Query5}
\end{figure}

with Turtle result as follows: 

\begin{figure}[htp!]
\begin{flushleft}
\begin{tcolorbox}
\begin{footnotesize}
\begin{verbatim}
@prefix dtsmm: <http://dtsmmkg.org/dtsmmkg/resource/> .
@prefix dtsmm-ont: <http://dtsmmkg.org/dtsmmkg/ontology#> .
@prefix owl: <http://www.w3.org/2002/07/owl#> .
@prefix skos: <http://www.w3.org/2004/02/skos/core#> .

dtsmm:cybersecurity_firm_riskiq a dtsmm-ont:Entity ;
 owl:sameAs <http://dbpedia.org/resource/RiskIQ> ;
 skos:related <http://dbpedia.org/resource/Computer_security> .
\end{verbatim}
\end{footnotesize}
\end{tcolorbox}
\end{flushleft}
 \label{fig:Query6}
\end{figure}


The results would tell us that this is a Computer Security company.
If we now would supply via RAG the existing 960 DBpedia knowledge triples on Microsoft plus the extracted relation triples deriving from our KG in-context to a LLM (in this example we used OpenAI GPT-4 Turbo\footnote{\url{https://platform.openai.com/docs/models/gpt-4-turbo-and-gpt-4}}), and then ask our initial question ~\ref{eq:ragquestion}, specifying to be brief, we would get the following answer from the system:\\

\textit{Yes, Microsoft is dedicating substantial resources to computer security technologies, as evidenced by its acquisitions of companies like RiskIQ, a leader in global threat intelligence and attack surface management, and CyberX, which specializes in securing IoT devices.}\\

where the latter information comes exactly from \textit{DTSMM\_KG}.

In summary, when generating textual content, the RAG model can then reference the most recent updates and developments in Digital Transformation encapsulated within our knowledge graph, which acts as a pool of novel knowledge taken from social media that RAG models can tap into.

\section{Digital Health News Knowledge Graph}

From the DNA news dataset we generated the Digital Health News Knowledge Graph (referred to as \textit{DHNEWS\_KG}), a KG comprising roughly 431k distinct (non-reified) triples, connecting 186k unique entities via a total of 1866 generalized relations. 
In the corresponding ontology, designed to describe \textit{DHNEWS\_KG} (\textit{dhnewskg-ont} namespace prefix), each extracted claim is successively reified into instances of the \texttt{dhnewskg-ont:Statement} class. Figure ~\ref{fig:StatementReificationDigitalHealth} provides an example of a claim reification, with the ontology instance \texttt{dhnewskg:drug\_tamoxifen} serving as \texttt{rdf:subject} and the \texttt{dhnewskg-ont:hasSupport} data property reporting the number of news articles
supporting the claim.

\begin{figure}
\begin{flushleft}
\begin{tcolorbox}
\begin{footnotesize}
\begin{verbatim}
dhnewskg-ont:statement_90694 a rdf:Statement ;
dhnewskg-ont:comesfromNewsArticle dhnewskg:lba0000020030305dz34000cl ;
dhnewskg-ont:hasSupport 1 ;
dhnewskg-ont:negation false ;
rdf:object dhnewskg:receptor ;
rdf:predicate dhnewskg-ont:block ;
rdf:subject dhnewskg:drug_tamoxifen .
\end{verbatim} 
\end{footnotesize}
\end{tcolorbox}
\end{flushleft}
 \caption{A sample reification for a statement concerning the \textit{DTSMM\_KG} resources \texttt{dhnewskg:drug\_tamoxifen} and \texttt{dhnewskg:receptor}. 
 }
 \label{fig:StatementReificationDigitalHealth}
\end{figure}

%
%
%
%

A sample of generated (un-reified) statements is illustrated in Table~\ref{SampleTriplesNews}, together with their support. 
The support distribution of the triples has a marked long-tail pattern, with a few key statements occurring frequently and a vast majority matched only a few times.

\begin{table}[t]
\caption{Sample statements from \textit{DTSMM\_KG}, with their support.}
\centering
 \begin{center}
  \begin{small}
 \begin{tabular}{l|l|l|l} 
 \hline
 \textbf{Subject Entity} & \textbf{Relation} & \textbf{Object Entity} & \textbf{Support}\\ \hline
Italy &report & coronavirus death & 374 \\
clinical trial& involve & patient & 128 \\
interactive graphic	&track & global spread & 90 \\
fitch & undertake & sensitivity analysis &75\\
administration & approve & drug & 47 \\
meningitis immunity&fight& endometrial cancer &35\\
dow chemical	&develop & drug & 28\\
drug tamoxifen &reduce & risk & 17\\
fibrocell &announce & fda acceptance & 2\\
zealand pharma &announce & fda acceptance & 1\\

\hline
 \end{tabular}
 \end{small}
 \end{center}
 \label{SampleTriplesNews}
\end{table}

\textit{DHNEWS\_KG} inherits entity typization of entities linked to  DBpedia via \texttt{owl:sameAs} predicate. Table~\ref{FrequentEntityTypes} lists the 20 predominant DBpedia-inherited types within the graph. All not-linked entities are classified into the generic \texttt{dhnewskg:Entity} type. 
Of the total set of unique \textit{DHNEWS\_KG} entities, around 8\% have been linked to DBpedia entries using 14975 \texttt{owl:sameAs} and 33345 \texttt{skos:related} predicates. Overall, 23.8\% of all triples had either subject or object entities linked to DBpedia.


\begin{table}[t]
\caption{Number of matches and unique matches of the 20 most represented DBpedia entity types in \textit{DHNEWS\_KG}.}
\centering
 \begin{center}
 \begin{small}     
 \begin{tabular}{c|c|c} 
 \hline
 \textbf{DBpedia Entity Type} & \textbf{\#Matched Entities} & \textbf{\#Unique Entities} \\ \hline
DBpedia:Organisation & 20050 & 2640\\
DBpedia:Company & 15667 & 1736\\
DBpedia:Country & 12124 & 324\\
DBpedia:Disease & 7881 & 918\\
DBpedia:Person & 6594 & 2611\\ 
DBpedia:ChemicalSubstance & 6583 & 1378\\ 
DBpedia:Drug& 5927 & 1180\\
DBpedia:Politician & 4069 & 1187\\ 
DBpedia:Work &1872 & 567\\
DBpedia:MonoclonalAntibody & 1258 & 128\\ 
DBpedia:GovernmentAgency & 1123 & 107\\ 
DBpedia:AdministrativeRegion & 1088 & 185\\ 
DBpedia:City & 971 & 205\\ 
DBpedia:Bank & 789 & 128\\
DBpedia:Biomolecule & 754 & 165\\ 
DBpedia:Group & 729 & 109\\
DBpedia:AnatomicalStructure & 656 & 151\\
DBpedia:ChemicalCompound & 631 & 191\\ 
DBpedia:ArchitecturalStructure &593 & 183\\
DBpedia:Gene & 586 & 124\\
dhnewskg-ont:Entity & 800527 & 185653\\
\hline
 \end{tabular}
 \end{small}
\end{center}
 \label{FrequentEntityTypes}
\end{table}

We made \textit{DHNEWS\_KG} publicly available via data access endpoints. 
Namely, we set up a Virtuoso SPARQL endpoint for this purpose where \textit{DHNEWS\_KG} can be queried, and analytical information on target entities, attributes, and relations can be retrieved in user-specified data formats\footnote{
\url{https://api-vast.jrc.service.ec.europa.eu/sparql/}. Currently, the access is password protected, with credentials available upon request to authors. Provisional credentials for the reviewing process: (dtsmm\_user, dtsmm\_user\_2024).
}. 
As an example, a SPARQL query like the one in Figure~\ref{fig:biogenQuery2} can be run to return all the 480 statements from the `\textit{DHNEWS\_KG}' graph 
having the target entity \texttt{dhnewskg:biogen} as subject, where \texttt{dhnewskg:biogen} is a graph resource linked to the DBpedia entry for the American multinational biotechnology company Biogen Inc.

\begin{figure}
\begin{flushleft}
\begin{tcolorbox}
\begin{footnotesize}
\begin{verbatim}
PREFIX dhnewskg: <http://dhnewskg.org/dhnewskg/resource/>
PREFIX dhnewskg-ont: <http://dhnewskg.org/dhnewskg/ontology#>
SELECT ?statement
FROM <DHNEWS_KG> 
WHERE { ?statement a rdf:Statement .
 ?statement rdf:subject dhnewskg:biogen . }
\end{verbatim}
\end{footnotesize}
\end{tcolorbox}
\end{flushleft}
 \caption{Query returning all \textit{DHNEWS\_KG} statements with the graph entity \texttt{dhnewskg:biogen} as \texttt{rdf:subject}.}
 \label{fig:biogenQuery2}
\end{figure}


Finally, we built aggregated analyses on top of  \textit{DHNEWS\_KG} and made it available through a collection of interactive visualizations, accessible in the Hugging Face Space \url{https://huggingface.co/spaces/zavavan/Digital_Health_News_Analytics_Dashboard}.

The visualizations across all the dashboard panel allow filtering per geographical macro-area (values  \textit{Europe/US/EU-US} values), enabling the detection of significant trends in digital health technologies within the European and US contexts.




The Top Entity Types bar plots in the dashboard show the predominant DBpedia-inherited entity types within the graph for triples tagged with Europe, US, and EU-US region codes via their article support.


For a subset of predominant types, the Top Key Entities plots track the occurrence of several key entities per year, with occurrence indicating that the entity is either the Subject or Object of an extracted triple in the KG. One can point out here how some major pharmaceutical industry corporations 
seem to have an impact on a global scale (both for Europe and the US), while major information technology giant corporations show different impacts in the Digital Health industry in the two contexts.


Lastly, within the Entity Chord Diagrams panel 
we present the most frequently connected entity pairs within the KG through chord illustrations, serving as both Subjects and Objects of predicative triples. The size of the chords corresponds to the support of the depicted relation\footnote{For the sake of visualization, we pre-filtered for relations with a minimum number of 20 occurrences in the dataset.}.

\chapter{Mapping the AECO Research Landscape}
\label{sec:aeco}
\section{Building Scientific Knowledge Graphs}
\label{sec:scientificKGs}
KGs have proved effective for representing research knowledge discussed in scientific papers and patents across several different domains~\cite{DESSI2021253, 10.1115/1.4052293,XiaoLi}. New generation ``scientific KGs" have shifted from representing purely bibliographic information of research publications to support the construction of extensive networks of machine-readable information about entities and relationships pertaining to a certain domain, enabling fine-grained semantic queries over large scientific text collections of the form: ``retrieve all methods that are used for Indoor Air Remediation in the time range $t$''.

Therefore, they can support downstream analytical services like technology trend analysis. For example,~\cite{luan2018multitask} uses the statistics of relation triples of type $<Method;Used-for;Task>$ automatically extracted from paper abstracts to reconstruct historical trends of the top applications of target methods such as \textit{artificial neural networks} in different areas like speech recognition and computer vision.

The generation of large-scale and accurate knowledge graphs of
scientific knowledge presents various challenges, specific to the content and style of scholarly communication. Keeping aside the architectures proposing assistant tools for human expert KG editing (\cite{10.1145/3360901.3364435}) which are typically not scalable, systems based on automatic IE pipelines have to solve generally more demanding tasks than in the general domain, due to the following reasons:
\begin{enumerate}    \item the domain entities to be detected are generally \textit{terms}, rather than named entities  (e.g. \textit{Geothermal Heat Pump System} as opposed to a named Person entity like e.g. \textit{Bill Gates}). Terms are more abstract concepts that are linked to each other by (possibly multiple) subclass relations. For example, a \textit{Geothermal Heat Pump System} is an instance of the more general class \textit{Heating, Ventilation, and Air Conditioning System}. Consequently, there are multiple ways to refer to the same terms at various levels of granularity, making mention detection more difficult.     
\item terms are often referred to via domain-specific and often ambiguous acronyms (e.g. \textit{HVAC system} for \textit{Heating, Ventilation, and Air Conditioning system}); 

\item extracting information from scientific articles requires
extracting relations across sentences using co-reference resolution, while the application of standard sentence-level IE systems would result in lower relation coverage and would generate sparse KGs.
\end{enumerate}

Methods for automatic generation of scientific KGs typically build upon supervised Relation Extraction (RE) models for capturing fine-grained relations between scientific entities in text~\cite{zhao2023comprehensive}. In the scholarly domain, the most influential entity and relation schema specification for this task is defined by the SciERC initiative ~\cite{luan2018multitask}\footnote{\url{https://nlp.cs.washington.edu/sciIE/}}.

The SciERC annotation schema defines six types of scientific entities, namely \textit{Task}, \textit{Method}, \textit{Metric}, \textit{Material}, \textit{OtherScientificTerm} and \textit{Generic} and five directed relation labels (\textit{Used-for}, \textit{Evaluate-for},  \textit{Part-of},   \textit{Feature-of}, \textit{Hyponym-Of}) plus two symmetric relations (\textit{Compare} and \textit{Conjunction})\footnote{The meanings of the entity and relation types are largely self-explanatory; however, for a detailed description, please refer to the full SciERC specifications: \url{https://nlp.cs.washington.edu/sciIE/annotation\_guideline.pdf}.}. Moreover, co-reference links are annotated
between identical scientific entities within and across sentences. 

As an illustration of this schema, Figure~\ref{fig:bratRelationVisualizations}.a shows the annotation of an NLP paper fragment, where the term ``probabilistic context-free grammar (PCFG)" is detected as \textit{Method} and linked to the \textit{Method} ``MORphological PArser MORPA" via \textit{Used-for} relation, encoding the fact that the former is applied within the latter. Notice also that an entity of type \textit{Generic} (the pronoun ``it") is annotated as it is involved in a relation and is co-referred with another entity (that is the MORPA parser, edge not shown here).

The SciERC initiative published also a training dataset, comprising annotations of 500 scientific abstracts, taken from conference/workshop
proceedings in various AI sub-communities. This work has laid the ground for several recent efforts towards the creation of large-scale scholarly KG infrastructures, such as CS-KG 2.0~\cite{dessi2025cs} and AI-KG~\cite{dessi2020ai}. However, the availability of trained IE tools and training dataset has limited until now these initiatives to Computer Science, AI and related domains.

\begin{figure*}[!ht]
\centering
\begin{minipage}[b]{0.8\textwidth}
    \centering
    \includegraphics[width=\linewidth]{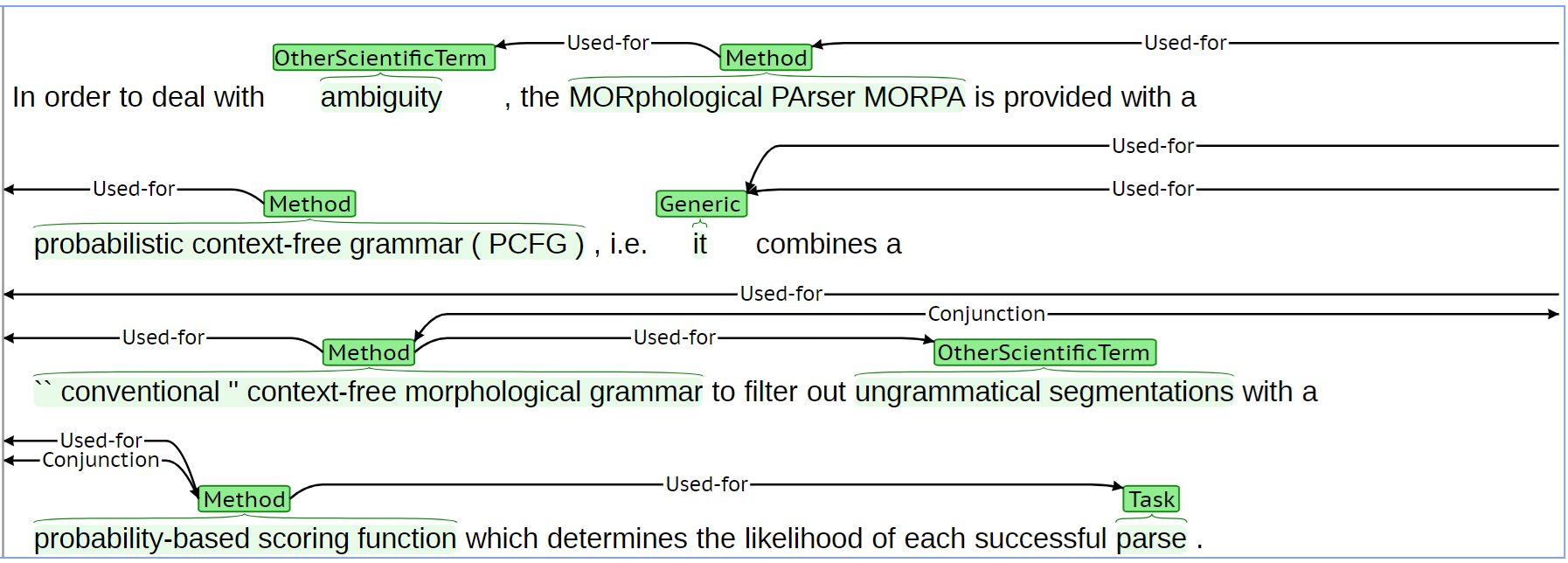}
    \\(a)
\end{minipage}
\hfill
\begin{minipage}[b]{0.8\textwidth}
    \centering
    \includegraphics[width=\linewidth]{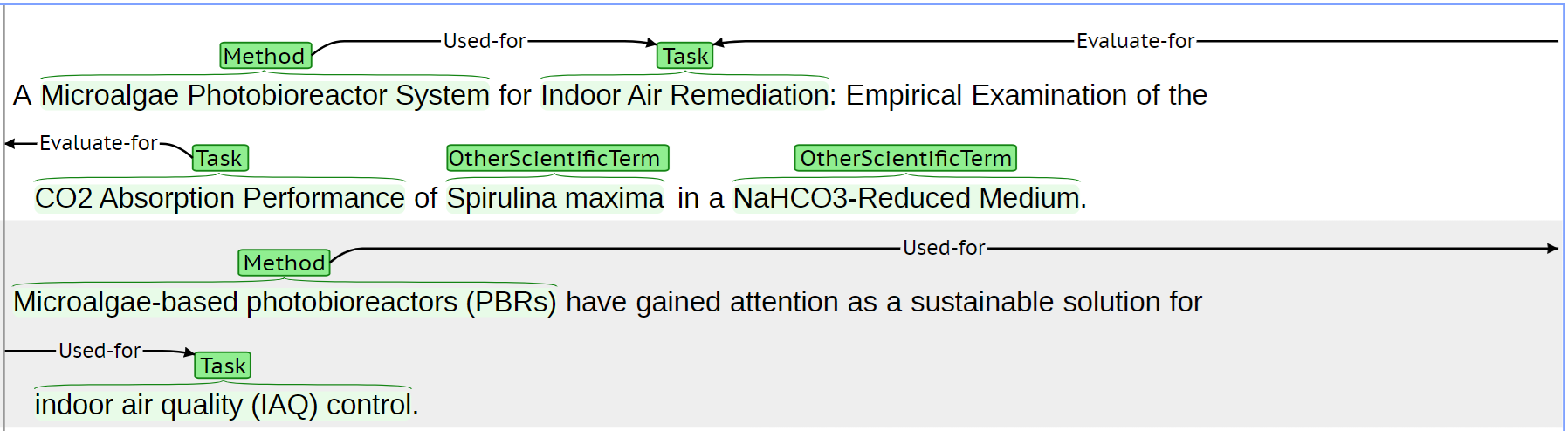}
    \\(b)
\end{minipage}
\caption{Sample Entity and Relation annotation from the SciERC dataset (a) 
and from an AECO paper abstract~\cite{app132412991} (b), following the SciERC annotation schema. The latter shows a \textit{Method} ``Microalgae Photobioreactor System'' and \textit{Task} ``Indoor Air Remediation'' extracted respectively as the head and tail of a \textit{Used-for} relation, resulting in the triple \textlangle Microalgae Photobioreactor System;\textit{Used-for};Indoor Air Remediation\textrangle, which encodes the claim that a ``Microalgae Photobioreactor System'' method is applied to solve/achieve the task of ``Indoor Air Remediation''.}
\label{fig:bratRelationVisualizations}
\end{figure*}

\color{black}

As part of a joint initiative on innovation intelligence analytics for the Architecture, Engineering, Construction and Operation (AECO) research and industry, carried out in collaboration with the Institute of Data Science and Artificial Intelligence of the University of Navarra, we designed and experimented with a hybrid workflow aimed at making sense of the recent research agenda in the AECO domain by identifying dominant research areas and tracking their evolution over time, providing quantitative grounding for insights on trending research directions and research foresight. 

The proposed method organizes the vast domain of AECO research into macro-topics: then, for each macro topic, we leverage information extraction technologies backed by deep learning models and LLMs to detect and track domain research entities such as Tasks and connect them with the Methods that have been proposed over time to solve them. To the best of our knowledge, this is the first attempt to apply semantic technologies to support data-driven innovation and technology intelligence at a large scale in the AECO domain.  

\section{Methodology Outline}
\label{methodology}

\begin{figure*}[!ht]
 \centering
 \includegraphics[width=0.9\textwidth]{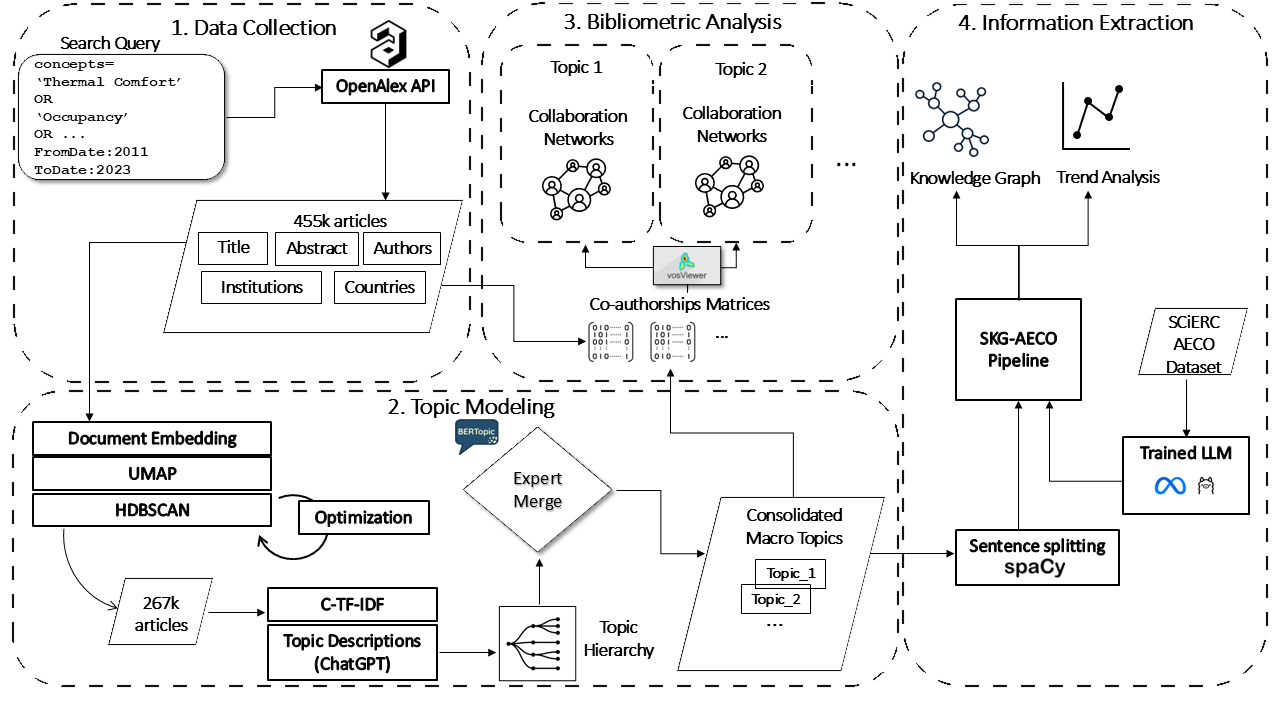}
 \caption{Flowchart of the text processing pipeline used in this research. The pipeline is structured into four main phases: 1. Data collection from the OpenAlex API; 2. Topic Modeling using optimization of BERTopic architecture and expert tuning; 3. Bibliometric analysis of collaboration networks for each consolidated topic; 4. Extraction of scientific Knowledge Graphs and generation of trend analysis for each consolidated topic, using the SKG-AECO pipeline.}
 \label{fig:pipelineFlowchart}
\end{figure*}

The methodology adopted in this study is structured into four main phases, as illustrated in Figure~\ref{fig:pipelineFlowchart}. First, we 
collect research papers from OpenAlex using its API (Data Collection). Subsequently, we use the BERTopic architecture and domain expertise to consolidate a set of optimized macro topics (Topic Modeling). Finally, for each macro topic, we carry out two parallel processes: a Bibliometric Analysis, generating collaboration network visualizations, and an Information Extraction analysis, using an adapted version of the SKG pipeline~\cite{dessi2022scicero} and producing trend analyses backed by relational graphs. We provide here a detailed description of phases 1, 2 and 4, which are more closely related to the subject of the present thesis.

A complete collection of interactive visualizations for all the analyses conducted in this study is made publicly available in a  dashboard accessible at: \url{https://huggingface.co/spaces/zavavan/AECO_Tech_Dashboard}.

Details on how to interact with and customize the single visualization panels are provided within each corresponding section in the rest of the chapter.

All the code and resources generated in this project and enabling the reproducibility of the analysis are shared in a public GitHub repository (see Appendix~\ref{sec:appendixAECO} for details).

\section{Data Collection}
\label{datacollection}
The research publications used for our analysis were sourced from the OpenAlex\footnote{\url{https://docs.openalex.org/}}, a fully open and high-coverage scientific graph database~\cite{priem2022openalex}. 

Created in 2022 as a replacement of the discontinued Microsoft Academic Graph (MAG)~\cite{mag},  OpenAlex consists of a large directed graph connecting five types of scholarly
entities, namely ``\textit{works}'' (234M) , ``\textit{authors}'', ``\textit{venues}'' (i.e., journals, conferences, etc.), ``\textit{institutions}'',  and ``\textit{concepts}''. 
The nearly 65k OpenAlex ``\textit{concepts}'' are topic labels automatically assigned by a text classifier based on titles and abstracts of the publications. They are organized into a 5-level hierarchy with 19 root-level concepts.

For our analyses, we sampled a base dataset by querying the OpenAlex API, retrieved all English-language papers published between 2010 and 2023 that were tagged with at least one concept label from a set of expert-selected terms, representing the entire AECO domain. The complete list of these concept tags, along with the corresponding coverage in terms of retrieved works, is provided in the ``\textit{resources}" folder of the code repository.

This resulted in a collection of around 466k publications. For all the following text-based analyses, we process titles and abstracts of the papers. 

Figure~\ref{fig:plotPubTrends} below shows monthly time series of the total number of publications and the number of publications per country, for the top 15 countries by number of publications in the dataset\footnote{In the OpenAlex database, the publication countries are the countries of the institutional affiliations that the publication authors claimed in the context of that work, extracted using a combination of matched institutions and parsing of the raw affiliation strings, in case specific institutional affiliation metadata are missing}. These top 15 publishing countries account for around 50\% of the total number of papers.

A complete visualization of country-based publication time series is available in the ``Publication Trends'' panel of the web dashboard. 

\begin{figure*}[!ht]
 \centering
 \includegraphics[width=0.9\textwidth]{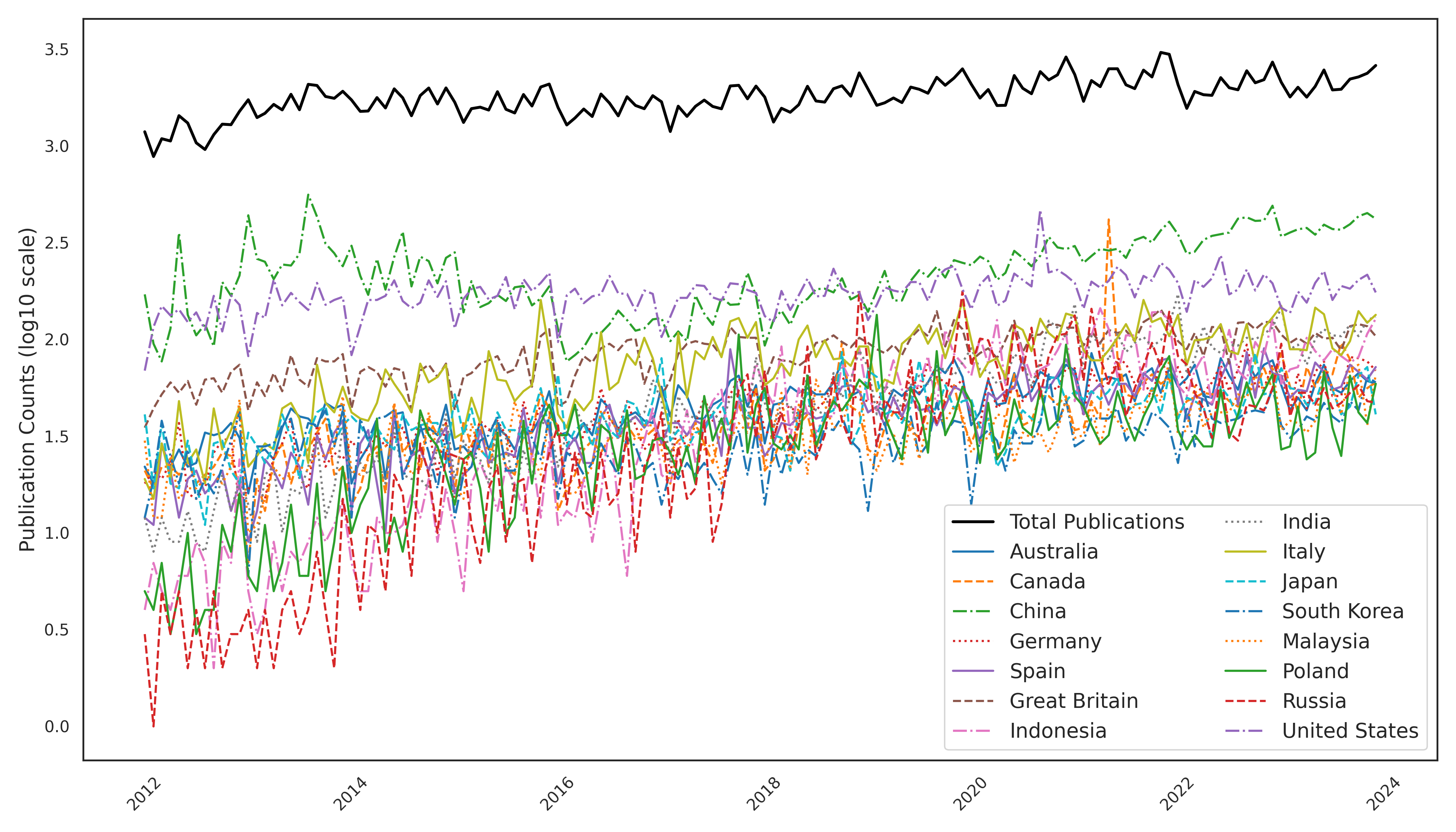}
 \caption{Log scale monthly time series of the overall publications and publications per country, for the 15 top publishing countries.}
 \label{fig:plotPubTrends}
\end{figure*}

\subsection{SciERC AECO Dataset}

From this base dataset, we sampled around 150 items (concatenated Title and Abstract), pre-processed and sentence split them using Spacy's English transformer pipeline \textit{en\_core\_web\_trf-3.6.1}\footnote{\url{https://github.com/explosion/spacy-models/releases/tag/en\_core\_web\_trf-3.6.1}} and sampled a final set of 1,016 sentences. We annotated the sentences for scientific entities and relations, based on a subset of the SciERC annotation schema from Section~\ref{sec:scientificKGs}. Namely, we annotated for scientific entities of type \textit{Task}, \textit{Method} and \textit{Metric}) and for the relations of type \textit{Used-for} and \textit{Evaluate-for}. The annotations were carried out independently by two domain experts using the Brat annotation tool~\cite{stenetorp-etal-2012-brat} and we measured the inter-annotator positive specific agreement relative to the entity detection sub-task (~\cite{HRIPCSAK2005296}\footnote{This corresponds to classical Cohen K inter-rater agreement, in tasks like NER where the number of negative cases is undefined.}), obtaining a mean F1 score of 0.73. This indicates a significant agreement between the human annotators, although some marginal task ambiguity emerged (see Section~\ref{sec:evalAECO}).

Figure~\ref{fig:bratRelationVisualizations}.b illustrates a sample visualization of SciERC AECO annotations, while Table~\ref{tab:sciercaeco} summarizes some statistics of the dataset. We publicly share the current version of the dataset (called SCIERC AECO) in the Hugging face dataset hub:
\url{https://huggingface.co/datasets/zavavan/scierc_aeco}. 

\begin{table}[!ht]
\centering
{\footnotesize
 \begin{tabular}{lll}
   \hline
 &  \textbf{Train} & \textbf{Test} \\
  \hline
Sentences &  816 &  200 \\
Negative Sentences & 536 &  124 \\
\texttt{Task} Entities & 512 & 123\\
\texttt{Method} Entities & 388 & 89 \\
\texttt{Metric} Entities & 51 & 22 \\
 \texttt{Used-for} Relations & 196 & 42 \\
 \texttt{Evaluate-for} Relations & 18 & 8 \\
 \hline
\end{tabular}
}
 \caption{Summary counts of the SciERC AECO dataset.}
 \label{tab:sciercaeco}
\end{table}

\section{Detecting Topic Clusters}
\label{sec:topicModeling}
In order to enhance the granularity of the KG-based research trend analysis we describe in Section~\ref{sec:aecoIe}, we integrate an upstream optimized topic model in the presented pipeline. This scalable approach makes the downstream IE module more sensible to low-level signals which would not emerge at the full collection level.
As we will show in Section~\ref{sec:aecoIe}, computing topic-specific statistics of the generated triples allows to extract very specific \textit{Task} and \textit{Method} entities, such as highly specialized tools and software packages.

Topic modeling is an unsupervised machine learning technique used to automatically discover abstract topics within a collection of documents.  We use the BERTopic neural topic modeling algorithm to identify the main latent topics emerging from our paper collection~\cite{grootendorst2022bertopic}.

The main feature setting apart BERTopic from more classical unsupervised topic modeling methods, such as LDA, is the use of contextual and fixed-sized dense vector representations of the input documents through embedding models, instead of static ``Bag-of-Word'' representations, which fail to account for semantic relationships among words.

The algorithm is highly modular and the sequence of processing modules can be split into two main parts. The the first one responsible for document clustering and encompasses:

\begin{enumerate}
    \item a document embedding step, creating multidimensional vector representations of documents using the encoding of pre-trained embedding models;

    \item dimensionality reduction of the document embedding vectors.

    \item clustering of the compressed embeddings and detection of groups of semantically similar documents.
\end{enumerate}

Once document clusters are created, the goal of the topic model is to detect and represent latent themes in each cluster, which is done sequentially by:

\begin{enumerate}
    \item computing a distribution with respect to document clusters of the word/terms in the whole dataset vocabulary;

    \item extracting of the best-representing terms for each cluster.

\end{enumerate}

This general BERTopic pipeline can be customized as each module is largely independent of the others. This allows us to optimize for a module configuration that achieves enhanced performance in our AECO domain dataset. 

Namely, for the document embedding step, we encode the concatenation of Title and Abstract of each research paper into a vector representation generated by the Sentence Transformer model ``all-mpnet-base-v2"\footnote{\url{https://huggingface.co/sentence-transformers/all-mpnet-base-v2}. This embedding model maps sentences and paragraphs to a 768-dimensional dense vector space and is optimized for semantic similarity tasks, providing strong out-of-the-box performance even on technical or semi-specialized domains.}.

We compute document clusters by optimizing the hyperparameters of a combination of the UMAP and HDBSCAN algorithms, similarly to what described in Section~\ref{relMapping}. 

In this case, hyperparameter grid-search is carried out optimizing with respect to the average coherence of the resulting topics. Assuming a topic is defined as a set $T = \{ t_1,t_2,...t_n \}$ of terms (tokens or n-grams) that best describe the latent theme of a document cluster, the topic coherence we used here estimates the reciprocal support between topic terms as the average cosine similarity between the probability vector of each term $t_i\in T$ and the probability vector of all other terms $t_j\in T$ in the topic, that is:

\[\mathrm{Coherence}_{c\_v}(T) = \frac{\sum_{i=1}^{n}\tilde{m}_\mathrm{cos}(\{t_i\},\{t_j\in T \mid t_j \neq t_i\})}{n}\]

where for each $t_i,t_j$:
\[\tilde{m}_\mathrm{cos}(t_i,t_j) = \mathrm{cosine\_sim}(\Vec{\mathbf{v}}_\mathrm{prob}(t_i),\Vec{\mathbf{v}}_\mathrm{prob}(t_j))\]
and the probability vector of a term subset is calculated as:
\[ \Vec{\mathbf{v}}_\mathrm{prob}(W') = \{ \sum_{w_i \in W'} \mathrm{prob}(w_i,w_j) \}_{j=1,2..,|N|}\]
where $\mathrm{prob}(w_i,w_j)$ is estimated based on term co-occurrence counts over a moving sliding window in the dataset corpus~\cite{Rudiger2022}. This method of assessing topic coherence has been shown to correlate fairly well with human judgment~\cite{Sandhiya2022305}.

Based on this metric, we proceed as follows. First, to discard marginal topic clusters in the original paper collection (possibly due to inaccuracy of the openalex concept tag filtering), we run BERTopic with default settings on the whole dataset, manually inspect the result topics descriptions, discard out-of-domain topics and filter out all papers belonging to discarded topics\footnote{Larger discarded clusters were mostly related to biological ecosystems, transportation, signal processing and textile industry.}, ending up with 267,319 items.

Subsequently, we find an optimal hyperparameter configuration for a 10,000 random document sample in order to estimate the optimal level of granularity of the clustering. 
Figure~\ref{fig:plotClusterNumCoherenceScore} plots the average topic coherence against the number of clusters for a subset of best-performing UMAP and DBSCAN hyperparameter configurations, showing that optimal coherence scores are reached for a configuration distributing the data sample into slightly more than 30 clusters.

\begin{figure*}[!ht]
 \centering
 \includegraphics[width=0.8\textwidth]{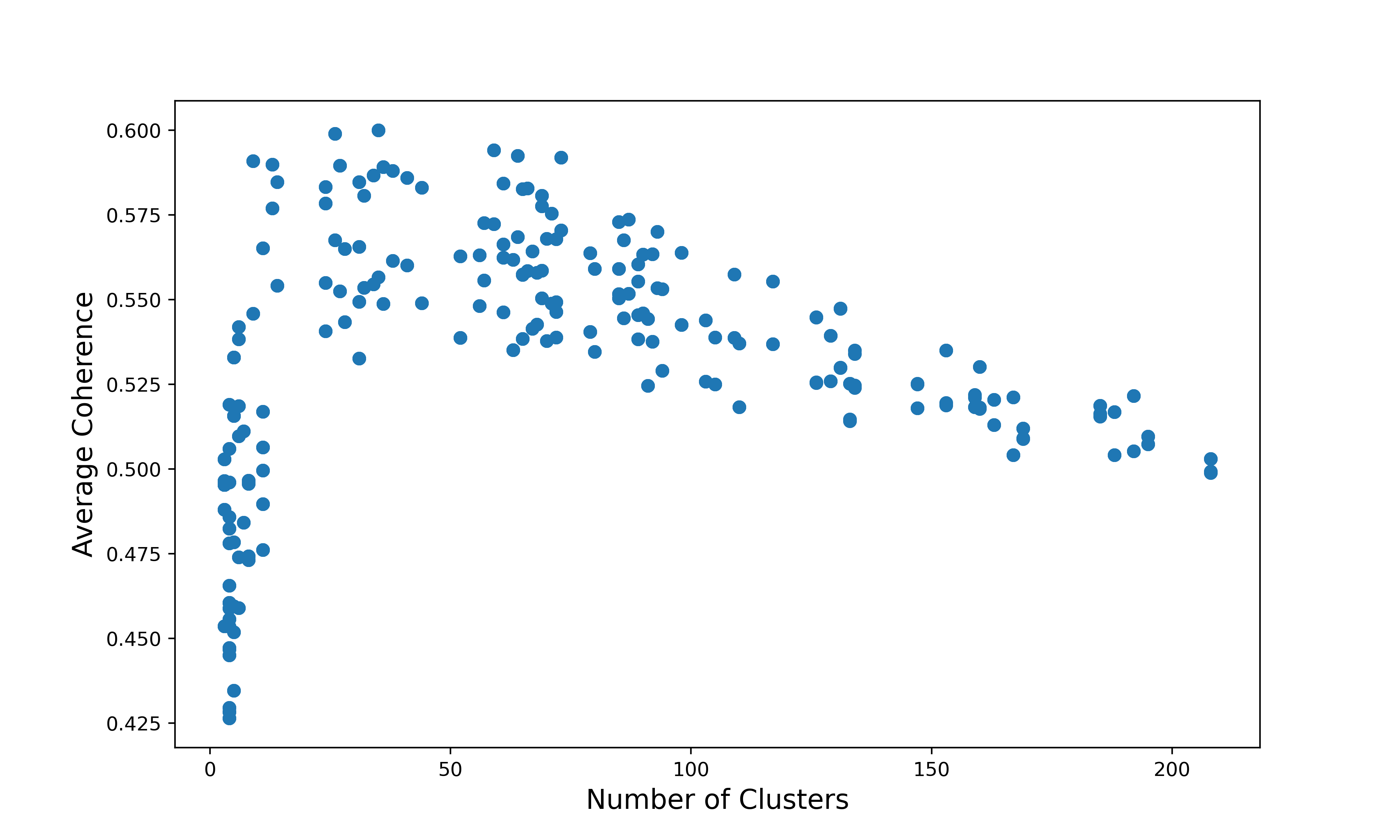}
 \caption{Average topic coherence values against the number of clusters for a subset of best-performing UMAP-HDBSCAN hyperparameter settings.}
 \label{fig:plotClusterNumCoherenceScore}
\end{figure*}

Using this parameter settings on the entire dataset results in a total of 52 clusters, which are visualized in a reduced 2-dimensional space in Figure~\ref{fig:plotClusterDataMap}, with the descriptive labels for each cluster being generated by fine-tuning the topic representations via the use of a LLM (see further down in the Section for details)\footnote{For simplicity, we only show the descriptive labels for the 10 largest clusters and use numerical labels for the others. The full list of topic labels is stored in the \textit{OptimizedAECOTopicLabels.tsv} file in the folder ``\textit{resources}" folder of the code repository.}. An interactive visualization of the topic modeling data map is made available in the ``Topic Map'' panel of the web dashboard, where the titles of article data points from the entire dataset can be accessed by hovering over them, while clicking on single data points redirects to the corresponding OpenAlex paper entry page\footnote{A search function also allows to retrieve single papers based on keywords in the Title and Abstract.}.

\begin{figure*}[!ht]
 \centering
 \includegraphics[width=0.8\textwidth]{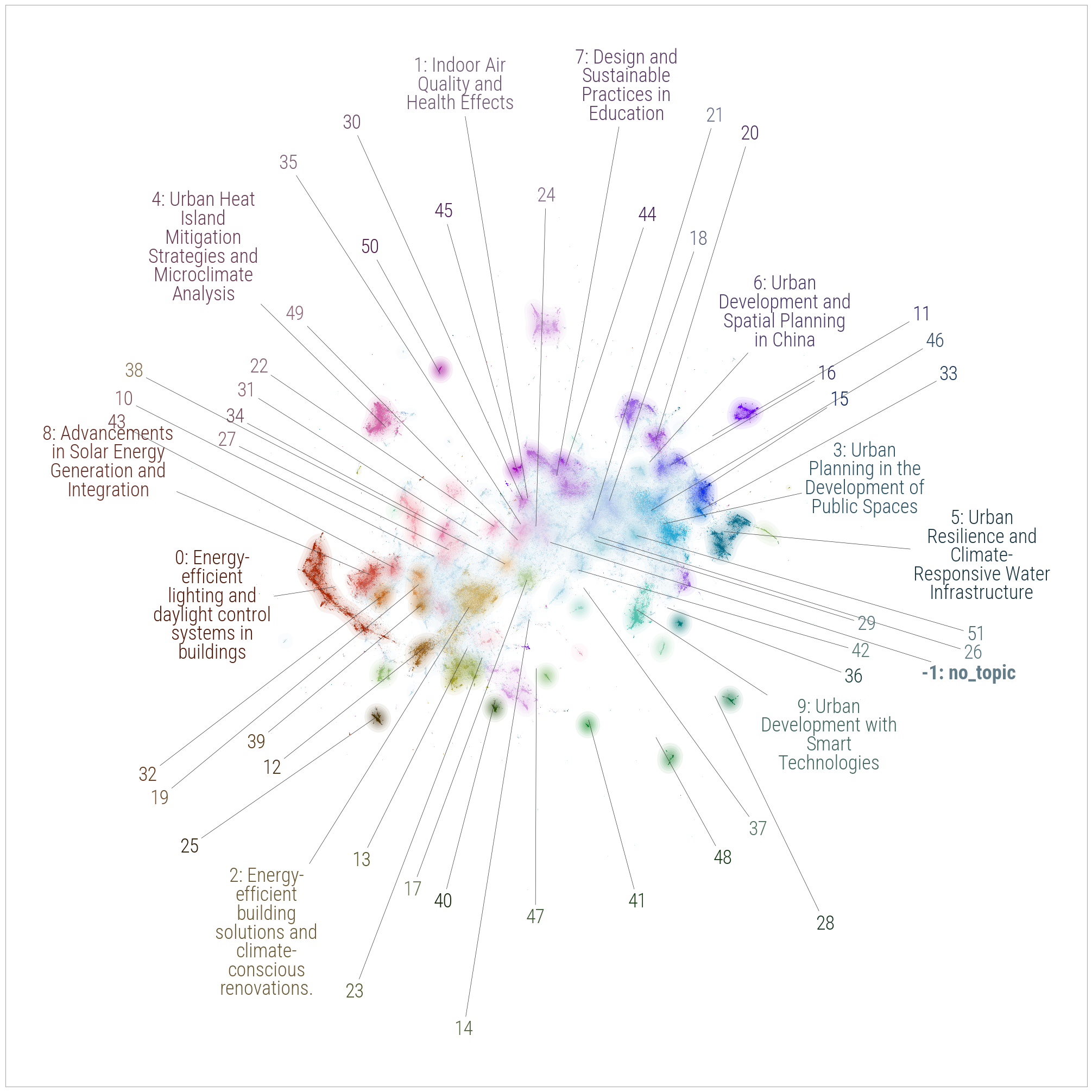}
 \caption{Reduced 2-dimensional visualization of the optimized 52 topic clusters of research papers, embedded using a Sentence Transformer model. The indicator lines originating from cluster labels (laid out here in rings around the data map, for clarity) point to each cluster's medoids. The \textit{-1:no\_topic} label denotes the set of outlier articles.}
 \label{fig:plotClusterDataMap}
\end{figure*}

The BERTopic's default topic representation is based on the c-TF-IDF weighting schema, a class-based adaptation of the classical TF-IDF that estimates the statistical relevance of a term $t$ for a class $c$ as:
\[W_{t,c} = \mathrm{TF}_{t,c} \cdot \log{(1 + \frac{A}{\mathrm{TF}_{t}} )} \]
where $\mathrm{TF}_{t,c}$ is the frequency of the term $t$ in the concatenation of all documents comprising the topic class $c$, $A$ is the average number of terms per class and $\mathrm{TF}_{t}$ is the total frequency of $t$ over all classes. 

In our settings, the vocabulary comprises the 1,000 most frequent unary terms (excluding English stop words) and an initial topic representation for each class is generated, that includes 10 vocabulary terms with the highest c-TF-IDF score. These terms are used to sample a subset of 10 most representative documents per class. Then, following the KeyBERT keyword extraction  technique (~\cite{grootendorst2020keybert}), both base terms and representative documents are embedded with the Sentence Transformer model, and vectors are compared to generate a 10-term fine-tuned topic representation. Finally, for each class, the 10 fine-tuned keywords and 10 documents are passed in as variables for a text generation prompt to the OpenAI ``gpt-3.5-turbo'' model (aka ChatGPT,~\cite{radford2018improving}) API, using parameters $temperature = 0.0$ and $diversity=0.6$.
The prompt is shared at the URL:
\url{https://github.com/zavavan/AECO_KG_Pipeline/blob/main/resources/gpt3.5_topic_representation_prompt.txt}.

Figure~\ref{fig:heatMap} in Appendix~\ref{sec:appendixAECO} shows a heat map of the coherence scores for the optimized topics, labeled using ChatGPT-generated descriptions. 

The average topic coherence score is $0.613$, with values ranging from 0.46 to 0.74 (standard deviation 0.067), which typically denotes fairly consistent topics~\cite{10.1145/2684822.2685324,lau-etal-2014-machine}. However, by embedding the c-TF-IDF representations of the topics and projecting them onto a reduced 2-dimensional embedding space (as shown in Figure~\ref{fig:2d-topic-representation} in Appendix~\ref{sec:appendixAECO}), one can notice that there are macro groups of clusters that share several semantic features and stand apart from each other.

We verified this observation by carrying out a manual inspection of the hierarchical structure of the generated clusters. Figure~\ref{fig:topicHierarchy} in Appendix~\ref{sec:appendixAECO} shows the dendrogram representation of the optimized clustering. 
One can notice, for example, that the sub-tree colored in green at the top of the dendrogram contains rather contiguous topics (indexes 0,2,4,12,13,14 and 25) all related to energy efficiency technologies for building thermal comfort.
We leverage such a hierarchical representation to merge several topic subsets placed close to each other at higher levels of the dendrogram, resulting in a number of macro clusters (16) very close to the one emerging from Figure~\ref{fig:2d-topic-representation}. Therefore, we consolidate the topic modeling using these 16 macro clusters, which are listed in Table~\ref{tab:consolidatedTopics} in Appendix~\ref{sec:appendixAECO} together with cluster size, natural language description labels generated by ChatGPT, and the topic base representations comprising the most relevant 10 terms.

\subsection{Analysis}

Sixteen dominant macro topics in AECO research emerged from the topic-modeling analysis, reflecting a diverse set of research interests centered around sustainability, urban planning, and technological advancements. The prominence of energy efficiency, smart cities, and sustainable construction indicates a strong focus on climate-aware design. The inclusion of BIM, solar integration environments, and acoustics suggests growing interest in digitalization and environmental quality in built spaces. The resilience of cities to climate change and the preservation of traditional architecture highlight a balance between innovation and heritage conservation. Finally, topics like urban agriculture and heat transfer show interdisciplinary expansions beyond conventional AECO research.

Figure~\ref{fig:topicOverTime} visualizes the evolution over time of the 16 identified AECO macro topics by showing semi-annual time series of the absolute numbers of publications per topic, in the time range 2012- early 2024. Energy Efficiency and Thermal Comfort remains the dominant topic, showing continuous growth until 2023, reflecting global sustainability efforts. Indoor Air Quality and Sustainable Air Conditioning Systems follows closely, likely due to increasing worldwide air quality concerns in cities. Between 2018 and 2020, Smart City Development and Urban Resilience overtook Child-Friendly Urban Spaces, indicating a broader shift from socially oriented to technology-focused urban planning. Landscape Planning declined from 2013 to 2020, with a notable spike in interest in 2020. Meanwhile, Topics 10–15, including BIM and Sustainable Materials, show static trends, indicating fields with steady but limited research interest.

An interactive version of the plots can be accessed in the ``AECO Macro Topic Trends'' panel of the web dashboard.

\begin{figure*}[!ht]
 \centering
 \includegraphics[width=0.9\textwidth, height=0.58\textheight]{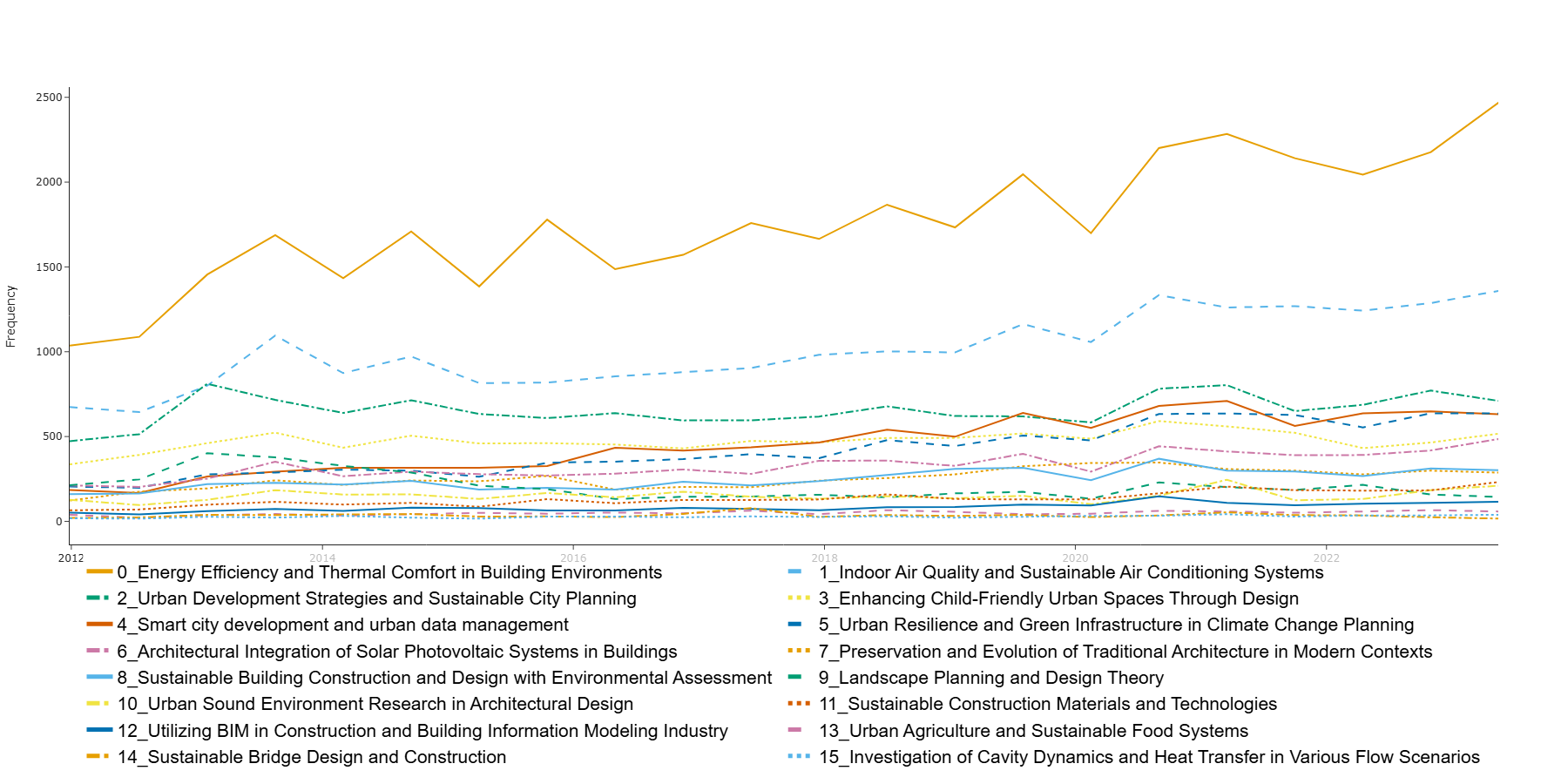}
 \caption{Evolution over time of AECO macro topics.}
 \label{fig:topicOverTime}
\end{figure*}

\section{Information Extraction Pipeline}
\label{sec:aecoIe}

Once the entire research paper dataset is partitioned into topic-coherent clusters, a semantically richer approach to exploring the AECO research landscape can be achieved by explicitly representing the content of the papers in each cluster through a KG of scientific entities~\cite{10.1145/3227609.3227689}, using the SciERC schema presented in Section~\ref{sec:scientificKGs}.

As we mentioned earlier, SciERC data schema is backing the implementation of NLP pipelines for the automatic generation of accurate, large-scale scientific KGs, such as SCICERO~\cite{dessi2022scicero}. We describe in the following the steps we carried out for adapting SCICERO's pipeline, originally designed for the Computer Science domain, to extract (a subset of) SciERC ontology entities and relations from our collections of AECO paper abstracts. The resulting pipeline is named \textit{SKG-AECO}. Moreover, we show how the resulting graph data infrastructure can be queried to generate analytical insights and trend analyses on the AECO domain. We will use macro topics 0, 6 and 12 as illustrative examples.

\subsection{SCICERO}
\label{scicero}
SCICERO is a hybrid NLP pipeline using a blend of deep learning trained models, rule-based heuristics and semantic web techniques to generate a large-scale KG~\cite{dessi2022cs} of 41M statements among 10M unique entities from a dataset of 6.7M  abstracts in Computer Science\footnote{The full ontology underlying this KG includes entity types \texttt{Task}, \texttt{Method}, \texttt{Material}, \texttt{Metric} and \texttt{OtherEntity} and 179 object properties.}. It comprises three main components: 

\begin{enumerate}
\item A set of extraction modules responsible for generating candidate entities and candidate relations connecting them; 

\item Entity and relationship handling, where entities are normalized and merged by linking to external knowledge bases. 

\item Triple validation, where final output triples are selected based on consistency with the most reliable ones (high support triples\footnote{Support refers to the number of papers from where a triple was extracted, or the number of methods that generated it.}) and an ontological schema.
\end{enumerate}

For further details on the SCICERO pipeline, refer to the original paper~\cite{dessi2022scicero}. 
The main source of candidate triples are models that have been trained in scientific text collections in the Computer Science domain\footnote{Namely, the Dygiepp model (\cite{wadden-etal-2019-entity}) and a number of relation extraction modules building on top of Dygiepp-generated entities.}. However, state-of-the-art models for scholarly Relation Extraction, trained on the SciERC dataset comprising AI/ML articles, do not perform well on new target domains, such as AECO.

For example, in a previous study (\cite{zavarella2024few}) we experimented with the SOTA architecture SpERT (see Section~\ref{sec:dlmethods}), by re-training it on SciERC train split (1861 sentences) using SciBERT (cased) embeddings~\cite{beltagy-etal-2019-scibert} and testing it on our small out-of-domain gold standard SciERC AECO. We observed that the performance degrades drastically from around 70\% to as little as 18\% for NER, and from 50\% to 7\% for RE. In the same study, we also indirectly proved that few-shot learning methods for ChatGPT generate low quality data for the SciERC task. In other words, a domain adaptation process is needed for the KG extraction models to work effectively on AECO data.

Therefore, we enhance the original pipeline with an Information Extraction module trained on target entity and relation instances from the SciERC AECO dataset.

\subsection{LLMs for structured information extraction}

Typically, deep learning RE models require a large number of training data, whose annotation requires costly domain expertise~\cite{zhao2024comprehensivesurveyrelationextraction}.
However, more recently, decoder-only LLMs, pre-trained on massive open-domain data volumes, have proven able to learn complex information extraction tasks in scientific domains from a small number of examples~\cite{NEURIPS2020_1457c0d6}.
Here, we adapt a method successfully applied in the material science domain (\cite{dagdelen2024structured}) to the fine-tuning of an LLM for a joint NER/RE task using only a few hundred structured prompt examples and no explicit schema definition. 
For each of the 816 SciERC AECO training set instances, an instruction tuning prompt-completion pair is generated as shown in Figure~\ref{fig:instrTuningPrompt}, where the LLM is prompted to extract target entities and relations from an input sentence and it is simply presented with sample output in the form of json-style annotation.

\begin{figure*}[!ht]
\centering
\footnotesize
\begin{minipage}{.9\textwidth} 
\begin{tcolorbox}[colback=yellow!5!white,colframe=yellow!50!black,
 colbacktitle=yellow!75!black,fonttitle=\ttfamily\small, title=\textbf{Zero-Shot Prompt}]
\begin{verbatim}
{  "sentence_text": "Water bio-remediation through a probiotic
                        layer system.",        
   "Tasks": { "T1": "Water bio-remediation" },                
   "Methods": { "T2": "probiotic layer system"},                
   "Metrics": {},        
   "Used-for": [ 
      { "T2": "T1"  } ],                
   "Evaluate-for": [],       
   "relevant": "true" 
}
\end{verbatim}

\noindent\rule{\linewidth}{0.4pt} 

\begin{verbatim}
{  "role": "user",
   "content": "From the following Text, extract non-overlapping
   entities of type Tasks, Methods and Metrics and extract
   Used-for relations between Methods and Tasks
   and Evaluate-for relations between Metrics and Methods.
   Text: Water bio-remediation through a probiotic layer
   system."}, 
 {"role":"assistant",
 "content": {"Tasks": {"T1": "Water bio-remediation"}, 
                    "Methods": {"T2": "probiotic layer system"},
                    "Metrics": {}, 
                    "Used-for": [{"T2": "T1"}], 
                    "Evaluate-for": []
                    }                    
}
\end{verbatim}
 \label{fig:instructionFineTuning}
\end{tcolorbox}
 \end{minipage}

\caption{Sample conversion of a SciERC AECO training instance into a prompt-completion pair for instruction fine-tuning using json-style annotation formalism. $T1$, $T2$ are entity indexes. Notice that no definitions of Entity and Relation semantics are provided in the instruction part of the prompt.}
 \label{fig:instrTuningPrompt}
\end{figure*}

Using the resulting dataset, we perform instruction fine-tuning on the smallest release of the Llama 2 foundation model (\url{meta-llama/Llama-2-7b-hf}~\cite{touvron2023llama2openfoundation}) using Hugging Face's \texttt{Trainer} class and \texttt{PeftModel} class implementation for wrapping the base Llama model to trainable PEFT weight decomposition matrices. Training is done on 6 epochs with a 8 batch size. We reached a minimum validation loss of 0.162, with an average of 0.3\% unparsable completions, indicating that the trained model achieves good performance on the task and provides structurally consistent output.

\subsection{SKG-AECO}
\label{sec:skg-aeco}

Figure~\ref{fig:skgAeco} depicts in detail the \textit{SKG-AECO} processing pipeline.

\begin{figure*}[!ht]
 \centering
 \includegraphics[width=0.9\textwidth]{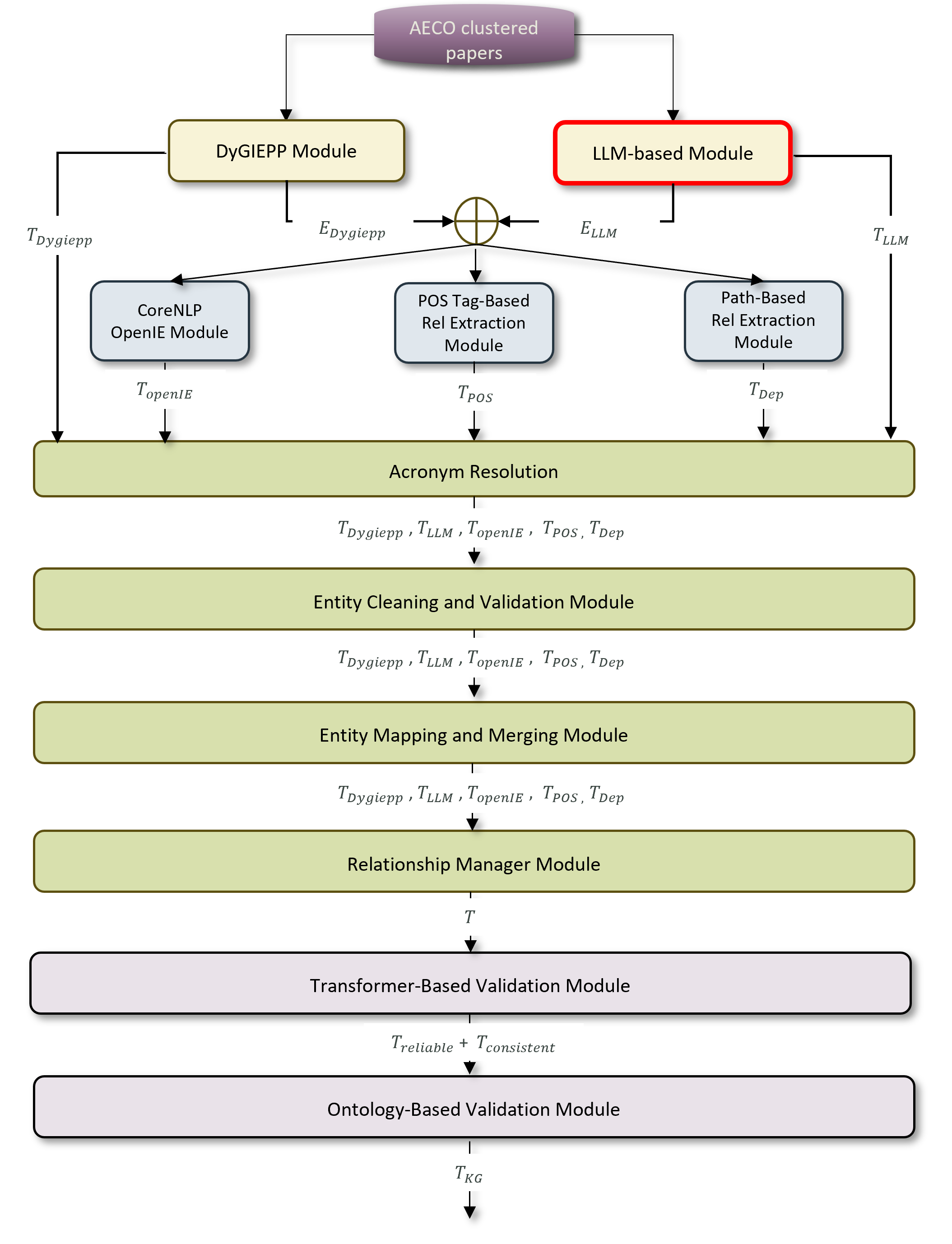}
 \caption{A detailed flowchart of the \textit{SKG-AECO} pipeline.}
 \label{fig:skgAeco}
\end{figure*}

On the total sample of 48,972 articles from macro-clusters 0, 6, and 12, the \textit{SKG-AECO} pipeline extracts a total of around 700k raw triples, with a contribution of 22.9\% of triples coming solely from the added LLM module (referred to as $E_{LLM}$ in the Figure). The triples undergo a multi-step process of validation, filtering, and merging, so that entities and relation predicates that refer to the same concepts are mapped to a unique representation.

\paragraph{Cluster-level acronym resolution} At the article level, acronyms are solved and mapped to a standard form by exploiting the regular expression patterns where they appear together with their extended forms, as in:
\textit{computational fluid dynamics (CFD)} or \textit{ building information modeling (BIM)}. However, this limits the recall as highly standardized acronyms in a domain (like \textit{BIM}) typically appear by themselves, outside any of those expanded pattern. By processing documents per topic cluster, we can take advantage of the acronyms' low ambiguity level and safely apply the acronym mappings collected globally, at the cluster level. In this way, we raise by 21.1\% the number of acronym-mapped entities, resolving 19,798 acronyms out of a total of 706,614 raw entities (2.8\%).

\paragraph{Linking to external knowledge bases} By linking to external knowledge bases, different candidate entities bearing the same meaning can be merged into a single entity. For example, \textit{``building insulation material''} and \textit{``insulating building material''} are both linked to the same Wikidata\footnote{\url{https://www.wikidata.org/wiki/Wikidata:Main_Page}} entry \url{http://www.wikidata.org/entity/q28942423} and thus merged together.
The mapping is performed in two parallel ways: first, we collect the Wikidata entries mapped from all OpenAlex \textit{concept} metadata associated with our entire research papers dataset, generating a dictionary resource of 12,700 AECO domain entries. Then, if a candidate entity string matches the main label or alternative label (i.e., labels linked by the property \textit{rdfs:altLabel}) of a Wikidata entry in the dictionary, it is mapped to that entry. Finally, the longest variant of all entities mapped to the same Wikidata entry is chosen as the representative label, and all other variants are replaced by it in the generated triples. Secondly, we run a SPARQL query retrieving Wikidata entities corresponding (via \textit{owl:sameAs} link) to DBpedia entities that belong to any of the subject categories \textit{dbc:Architecture}, \textit{dbc:Building}, \textit{dbc:Construction}, \textit{dbc:Engineering}, \textit{dbc:Operations} or their sub-categories (via \textit{skos:broader} relation). 

Out of 222,633 candidate entities, 3,790 (1.7\%) and additional 765 (0.34\%) are mapped to Wikidata through the first and second method, respectively\footnote{As computational methods are pervasive across the AECO domain, we also keep from the original SCICERO pipeline the module mapping entities onto the Computer Science Ontology (CSO~\cite{salatino2019cso}). This generates an additional 1,129 (0.5\%) entity mappings.}.
  
\paragraph{Transformer-based merging} In the SCICERO pipeline, entities that are not linked to external resources undergo a merging process based on cosine similarity of their corresponding vector representations, derived from \textit{paraphrase-distilroberta-base-v2} embedding model\footnote{\url{https://huggingface.co/sentence-transformers/paraphrase-distilroberta-base-v2}}. After some tests, we empirically lowered the similarity merging threshold to 0.75, while keeping the remaining mechanism unchanged.

\paragraph{Triple selection} By the entity merging above and relation ontology mapping mechanisms (as detailed in the SCICERO paper), the raw triples get reduced to 307,627 generalized triples. As a last step before graph generation, triples get further refined by taking: 
\begin{itemize} 
\item the set $T_{reliable}$ of reliable triples, that is triples extracted from a minimum number of articles $s_1$ and by a minimum number of extractor tools $s_2$, with $s_1$ and $s_2$ empirically set to 3 and 2, respectively;
\item an additional subset $T_{consistent}$ of non-reliable triples which are classified as consistent with $T_{reliable}$ by a transformer-based binary classifier\footnote{Based on \textit{scibert\_scivocab\_uncased}~\cite{beltagy2019scibert}.} fine-tuned on around 16,000 positive examples sampled from $T_{reliable}$ and 16,000 synthetically generated negative examples. 
\end{itemize}

The union of $T_{reliable}$ and $T_{consistent}$ forms the set of 208,462 final triples $T_{KG}$, about 29\% of the initial raw triple set.

\section{Evaluation}
\label{sec:evalAECO}

In order to assess the reliability of our IE approach, we performed an evaluation of the precision of the generated triples.
Out of the entire range of triple output by the original SCICERO pipeline, $\langle$\textit{Method; Used-for; Task}$\rangle$ and $\langle$\textit{Method; Used-for; Method}$\rangle$ are the types that we will use for trend analysis in see Section~\ref{sec:trends}. Therefore, we restrict the evaluation to these two types of SciERC triples.

We randomly sampled 300 triples, evenly distributed from the high-support and low-support groups (support $\geq 5$ and $\leq 2$, respectively), and assigned them to a total of 12 independent, AECO domain expert annotators. Each triple was assessed by three annotators as True or False, where a True label was assigned when: 1) the triple Subject and Object entities are linked by a \textit{Used-for} relation in the text sentence; 2) the triple Subject and Object are fully extracted, i.e. they do not contain less information than what is stated in the text; 3) both the triple Subject and Object are correctly classified as Task/Method.

We calculated inter-rated agreement using Gwet’s AC1 coefficient~\cite{gwet2008computing}. AC1 generalizes to annotation tasks where each item is assessed by only a subset of the annotators pool, while additionally using a corrected baseline of expected (chance) agreement that is more robust to label class imbalance\footnote{In our case, 58.7 and 41.2\% True and False, respectively.}, than the standard Cohen's $\kappa$ and Krippendorff’s $\alpha$. The resulting agreement score is 0.361 (overall observed agreement 0.670, Gwet chance agreement 0.484), which is commonly viewed as a 
``fair" agreement level but still signals a residual element of unreliability of the annotators' judgments.

Consequently, we performed an outlier analysis by computing per-annotator Cohen's $\kappa$ agreement with respect to the majority vote labels and identified a subgroup of 3 unreliable annotators with lower $\kappa$ values in the range 0.35-0.38, (versus $\geq 0.54$ values for most of the annotators). By removing this subgroup, the resulting AC1 score rises to 0.507, which is typically interpreted as ``moderate" agreement. 

Overall this indicates that, while not outstanding in absolute terms, the validation reliability is driven down by poor compliance to annotation instructions by a few outlier annotators, rather than by an inherent difficulty or subjectivity of the task.

Table~\ref{PrecisionEval} displays the Precision score values for the overall 300 manually triples, as well as for the subsets of high support, low support, and for the ones generated solely by the AECO-customized LLM.

\begin{table}[!ht]
\centering
  \begin{center}
 \begin{small}
 \begin{tabular}{|c|c|} 
 \hline
 \textbf{Extraction Method} & \textbf{Precision} \\ \hline
 Random Triples & 0.695\\
 High Support & 0.758\\
 Low Support & 0.637 \\
 LLM-generated & 0.683\\
 \hline
 \end{tabular}
 \end{small}
 \end{center}
 \vspace{-0.4cm}
 \caption{Triple evaluation over a set of 300 triples.} 
 \label{PrecisionEval}
\end{table}

Overall, the precision of the method is in line with the results from the original SCICERO pipeline and robust across all configurations. 

Analogously to the original SCICERO pipeline, triple precision correlates with triple support, with high support triples reaching a 0.76 precision level~\cite{dessi2022scicero}. 

Through error analysis, we found that a significant part of the error rate was due to the Transformer-based entity merging method. In fact, we boosted entity merging by lowering the similarity threshold from 0.9 to 0.75, in order to enhance entity generalization and mitigate data sparseness for the subsequent trend analysis, given the relatively low size of our cluster datasets. However, this accounted for a significant share of invalid triples, such as for example $\langle urban \, heat \, island;uses;urban\, green\,space\rangle$ which was wrongly extracted from the sentence
\textit{``Urban greening through local to landscape management is a proposed strategy to combat UHI and improve environmental justice in city neighborhoods.''}. In this case, the entity \textit{combat UHI} was overgeneralized to \textit{urban heat island} via acronym resolution and cosine similarity-based merging.

Finally, the novel LLM component shows a Precision level comparable to the SCICERO native extraction modules. Consequently, from the integration of the module, one can estimate a significant addition of true positive triples and a consequent recall gain.

\section{Results}

\subsection{AECO Research Knowledge Graph}
\label{sec:graph} 

From the entire triple set generated by deploying the \textit{SKG-AECO} pipeline, we select the subset of 255,764 triples of type $\langle$\textit{Method; Used-for; Task}$\rangle$ and $\langle$\textit{Method; Used-for; Method}$\rangle$ and make it publicly available as a preliminary version of the \textit{AECO} knowledge graph.

The \textit{AECO Research Landscape Knowledge Graph} (hereafter \textit{AECO-KG}) connects 15,037 and 22,240 unique \textit{Task} and \textit{Method} entities via the object property \textit{usesMethod} defined in~\cite{dessi2022scicero}.
The ontology describing \textit{AECO-KG} is an adaptation from SCICERO's\footnote{\url{https://scholkg.kmi.open.ac.uk/cskg/ontology\#}}, where classes and properties are defined in the namespace \url{http://aeco-research.org/aecokg/ontology\#} (prefix \textit{aeco-ont}), instances in the namespace \url{http://aeco-research.org/aecokg/resource/} (prefix \textit{aeco}) and the source papers from where triples originated are typed as \textit{aeco-ont:OpenAlexPaper} entities with a data property \textit{aeco-ont:hasOpenAlexURL} associating them with their unique URL in the OpenAlex platform. 

Each claim encoded by one \textit{AECO-KG} triple is reified into an instance of the \textit{aeco-ont:Statement} class and associated with the collection of entities of type
\textit{aeco-ont:OpenAlexPaper} it was generated from (using the property \textit{provo:wasDerivedFrom}\footnote{PROV-O - \url{https://www.w3.org/TR/prov-o/}}) 
and the cardinality of such collection (\textit{aeco-ont:hasSupport}). 

The \textit{AECO-KG} graph together with its ontology definitions are accessible as a single RDF format 
serialization within the European Data portal\footnote{https://data.jrc.ec.europa.eu/dataset/996d2a1b-69c9-4b27-b9b1-e0913f7f2d77}. Furthermore, we have set up a Virtuoso SPARQL endpoint where \textit{AECO-KG} can be queried, and analytical information on target entities,
attributes, and relations can be retrieved in user-specified data formats\footnote{The service is accessible at \url{https://api-vast.jrc.service.ec.europa.eu/sparql/}.}. 
As an example, a SPARQL query like the one in Figure~\ref{fig:sparqlQuery} returns the list of Tasks that are claimed to be solved using the green roof architectural solution (\textit{aeco-ont:green\_roof}), according to 1395 extracted triples. The query additionally extracts the URLs of the (899) research papers generating those triples. 

\begin{figure*}[htbp]
\centering
\begin{minipage}{.9\textwidth} 
\begin{flushleft}
\begin{tcolorbox}
\begin{small}
\begin{verbatim}
PREFIX aeco-ont: <http://aeco-research.org/aecokg/ontology#>
PREFIX aeco: <http://aeco-research.org/aecokg/resource/>
PREFIX rdf: <http://www.w3.org/1999/02/22-rdf-syntax-ns#>
PREFIX provo: <http://www.w3.org/ns/prov#>

SELECT ?subject ?url
WHERE {
   ?stmt aeco-ont:subject ?subject ;
   aeco-ont:predicate aeco-ont:usesMethod ;
   aeco-ont:object aeco:green_roof .
   ?subject rdf:type aeco-ont:Task .
   ?stmt provo:wasDerivedFrom ?paper . 
   ?paper aeco-ont:hasOpenAlexURL ?url . }
\end{verbatim}
\end{small}
\end{tcolorbox}
\end{flushleft}
 \end{minipage}
 \caption{Sample SPARQL query returning \textit{aeco-ont:Task} entities that are claimed to be applying the Method \textit{aeco:green\_roof} according to the \textit{AECO} graph statements, together with the URLs of the papers supporting the claims.}
 \label{fig:sparqlQuery}
\end{figure*}


\subsection{Trend Analysis}
\label{sec:trends} 

For each target macro-cluster, we perform trend analysis based on the output triples of type $\langle$\textit{Method; Used-for; Task}$\rangle$ and $\langle$\textit{Method; Used-for; Method}$\rangle$ extracted from the research papers belonging to that cluster. A complete visualization of trend analysis time series is available in the ``Research Tasks and Methods Trends'' panel of the web dashboard.


Figure~\ref{fig:trends_tasks-topic-0} shows the historical trends of the most frequently occurring \texttt{Task} entities in triples from micro-cluster 0, representing the evolution of key applications in this AECO topic area. \textit{Energy performance} has consistently been the most researched task throughout the period. The second most prominent task is \textit{thermal comfort}, showing a steady upward trend that reflects the sustained focus on occupant-centric performance. While \textit{daylighting system design} was initially among the top three, in recent years it has been surpassed by emerging tasks such as \textit{urban heat island} and \textit{outdoor thermal comfort}.

\begin{figure*}[!ht]
 \centering
 \includegraphics[width=0.9\linewidth]{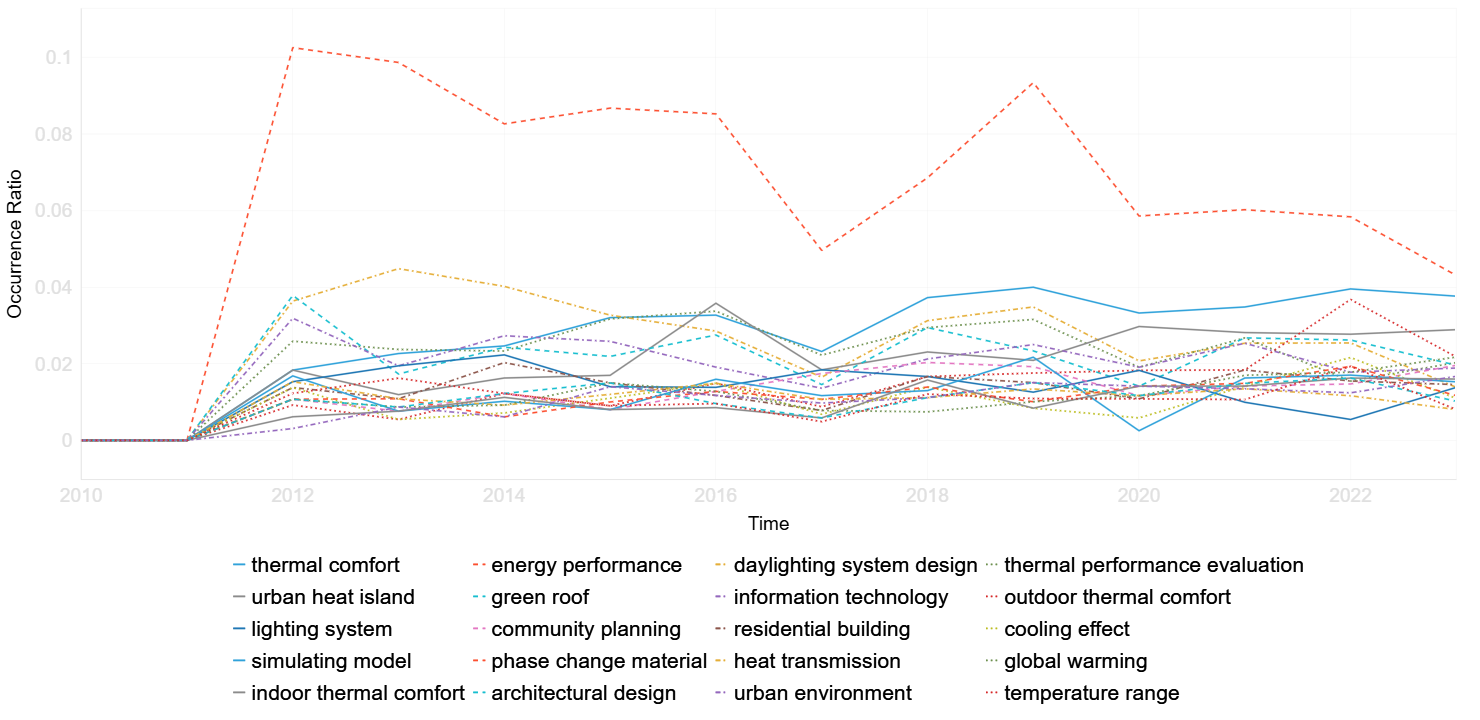}
 \caption{Trend analysis of the top 20 Tasks for macro-cluster 0, with the Tasks listed in the legend at the bottom of the plot. The y-axis measures the ratio of articles mentioning the Tasks to the overall number of articles in the cluster.}
 \label{fig:trends_tasks-topic-0}
\end{figure*}

Analogously, Figure~\ref{fig:trends_methods-topic-0} shows the historical trends of the most frequent \textit{Method} entities in cluster 0 triples, capturing the evolution over time of key technological solutions in this AECO topic area. \textit{Information technology} remains the most widely used method across the entire period, with peaks around 2016 and continued relevance thereafter. It often appears in sentences where it functions as a tool or enabler, representing how the study was conducted (e.g., use of sensors, data systems, digital tools). From 2016 onward, \textit{urban heat island} methods gained prominence, temporarily overtaking all others between 2018 and 2020. It often becomes a framing method, a contextual variable that structures simulations, models, or measurements. \textit{Global warming} also shows a steady increase, becoming the third most used method in recent years. It acts as a guiding context for scenario-based analysis or long-term forecasting, especially in energy demand, comfort evaluation, or mitigation strategies. Additional methods, such as \textit{daylight} and \textit{urban green space}, display consistent but lower usage over time.

\begin{figure*}[!ht]
 \centering
 \includegraphics[width=0.9\linewidth, height=\textheight, keepaspectratio]{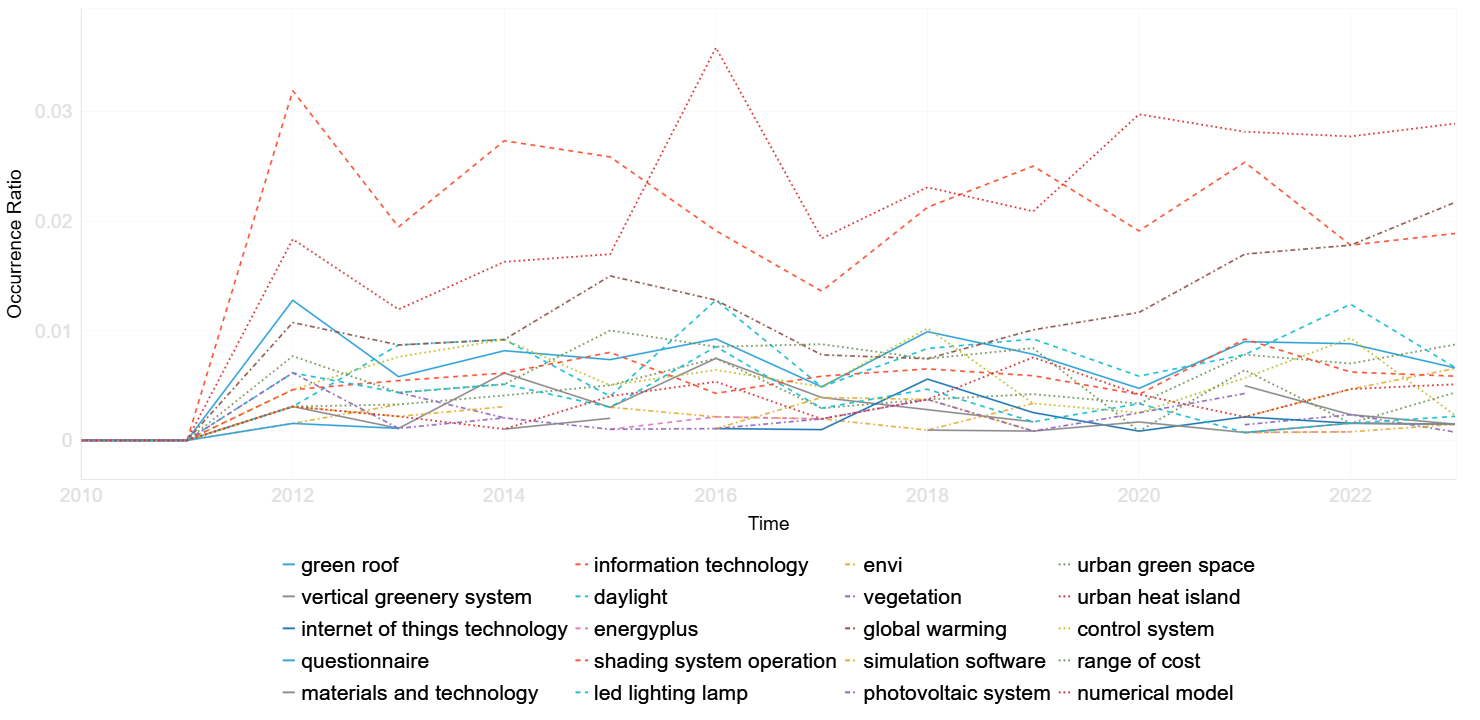}
 \caption{Trend analysis of the top 20 Methods for macro-cluster 0, with the Methods listed in the legend at the bottom of the plot. The y-axis measures the ratio of articles mentioning the Method to the overall number of articles in the cluster.}
 \label{fig:trends_methods-topic-0}
\end{figure*}

\chapter{Building Causality Graphs from Biomedical Text}
\label{sec:causalRE}
\section{Causality in Biomedical Text}

Causal reasoning is a core aspect of human cognition, and extracting cause-effect relationships from text has long been recognized as a critical task in NLP, particularly for building causal networks that support prediction and decision-making~\cite{yang2022survey}. 

\paragraph{\textbf{Quantitative vs. Qualitative Approaches}} Approaches to causal analysis can be broadly classified into two categories: \textit{qualitative} and \textit{quantitative} methods. Qualitative approaches typically frame the problem as a task of Causal Relation Extraction (CRE), where a single-label, binary or multi-class causal relation between candidate entities is detected as claimed in an input text, optionally together with some additional variables affecting the causal effect. On the other hand, quantitative approaches aim to assess the strength of causal links and manage the uncertainty inherent in causal inference~\cite{cui2024odysseycommonsensecausalityfoundational}. These latter approaches often build on theoretical models such as the Potential Outcome Framework~\cite{rubin74} and Graphical Models~\cite{pearl2018book}. 

For example, in an observational study with $K$ units indexed as $i = 1,...,K$, with units partitioned by the values of a treatment assignment variable $Z_i$ (where $Z_i=1$ in the treatment group and $Z_i=0$ for a control group), the Potential Outcome model aims to estimate the difference between these groups on an outcome variable $Y_i$, that is the Average Treatment Effect:
\begin{equation}
ATE = \mathbb{E}(Y_i(1) - Y_i(0))   
\end{equation}
where $Y_i(1)$ is the $Y$ value for the unit $i$ if $i$ is in the treatment group and $Y_i(0)$ is the $Y$ value for the unit $i$ if $i$ is in the control group. In other words, as the difference between the potential outcomes for each single unit $i$, denoted as:
\begin{equation}
\Delta_i = Y_i(1) - Y_i(0)    
\end{equation}
can never be directly observed (one patient can either receive the treatment or not), one can resort to estimating the average difference for a population sample. 

In a settings where one is not directly designing and managing a clinical trial but is relying instead on passively collected textual observational data such as Electronic Health Records (EHRs), generated through routine care and monitoring, the population sample will consist of text units $T_i$, where each $T_i$ must be mapped to a vector of $p$ pre-treatment covariates $X_i = (X_{i1},X_{i2},...,X_{ip} )$\footnote{Such as patient age, gender, previous conditions, etc.}, by applying some text analysis tools~\cite{mozer2024}.

In contrast, qualitative approaches bypass the uncertainty estimate and aim to directly identify potential causal relations between variables, such as entities or events in text.
This can result in  oversimplification of the causal effect estimate. For example, by default crucial intervening factors are not taken into account such as \textit{confounders}\footnote{A confounder is a variable that causally influences both the treatment and the outcome and can generate a spurious association between the two, if not accounted for.}. While techniques have been recently proposed to derive more complex causal variables from text analysis (see~\cite{mozer2024}), this is outside the scope of this study, which entirely falls within the framework of a qualitative representation of causal relationships.

Annotation schemas for qualitative Causal Relation in NLP commonly draw from the cognitive theory of force dynamics~\cite{10.1093/oxfordhb/9780199399550.013.13}, which conceptualizes causality as dynamic force interactions between entities, aligning better with human intuition.

In the biomedical domain, the ability to accurately identify causal relationships between events or entities in text is critical to advance knowledge discovery~\cite{sohag2025,Behaviormetrika2022}. Causal relationships underlie fundamental biomedical insights, forming clinical decision-making for example in treatment recommendations.
However, manually analyzing vast amounts of biomedical literature and clinical texts is infeasible and has triggered the development of automated approaches for the extraction of causal relationships, enabling researchers and healthcare professionals to identify potential risk factors, understand disease progression, and assess treatment effectiveness more efficiently and accurately~\cite{AKKASI2021103820,VANSCHAIK2023,Pawar2023157,Sriram2025}.

\paragraph{\textbf{Text Corpora}} A range of diverse unstructured observational data, including EHRs, clinical notes, and online drug reviews, can serve as valuable sources for causal inference experiments. These data sources capture real-world patient experiences and medical events, offering opportunities to generate inexpensive causal effect estimates when Controlled Randomized Trials (CRT) are not practically or ethically feasible~\cite{Shen2021,mozer2024,Fernainy2024}.

At the same time, the potential of leveraging such unstructured textual data is hindered by the complexity and variability of biomedical texts, which often contain diverse linguistic patterns and domain-specific terminologies, as well as implicit causal statements that are not immediately obvious without context~\cite{Kilicoglu201646,Lease200558,AKKASI2021103820}. The variety of text categories that can be used as observational data include:

\begin{enumerate}
    \item \textit{Electronic Health Records:} these include admission, progress and discharge notes, summarizing hospitalizations, stays, diagnoses, interventions, prescribed medications, and follow-up plans. 
    They are characterized by an highly technical language, specialized medical terminology, abbreviations, acronyms (even institution-specific), as well as telegraphic style including fragmented sentences and shorthand forms (e.g. \textit{``pt c/o cp x3d"} for \textit{``patient complains of chest pain for 3 days
    "}). Finally, they contain non-standardized Protected Health Information tokens.
   \item \textit{Drug Reviews}, as gathered from online health forums or social media platforms, are typically non-curated text featuring layperson vocabulary (e.g. \textit{``heartburn"} for \textit{``gastroesophageal reflux"}), non-standard spelling/grammar, commercial or slang drug names, subjective language, together with irrelevant anecdotal comments and vague temporal references (\textit{``after a few weeks I felt dizzy"}).
\end{enumerate}

The diversity in how causal information is expressed across these text collections makes it difficult to develop models that can consistently perform well across diverse datasets and scenarios~\cite{tan-etal-2023-recess,10.1007/s10115-022-01665-w,10.1162/tacl_a_00511,lishuang2024}. Additionally, the scarcity of comprehensive, labeled corpora specific to causal relationships 
limits the ability to train models effectively~\cite{AKKASI2021103820,Xu20201519,Yerkhassym20221549,VANSCHAIK2023}. 

The goal of this study is to evaluate the effectiveness of advanced open-source LLMs for CRE in the biomedical domain. While CRE can be approached using general RE techniques, identifying causal relationships presents distinct semantic challenges, as it often requires capturing implicit reasoning, temporal order, and contextual nuance beyond surface-level associations. Therefore, we systematically compare a range of LLM architectures and learning/inference strategies in the context of CRE. We benchmark these methods against two strong baselines using BERT and ClinicalBERT encoders, conducting experiments on the MIMICause dataset~\cite{khetanmimicause}, a well-established resource for CRE in clinical text. To assess the cross-domain adaptability of these models, we also evaluate them on two additional datasets with partially different text characteristics and causal relationships classification schemas.

\color{black}

\subsection{Task Definition}
Regardless of the specific characteristics of the input text, our CRE task of identifying causal relations can be formulated as a single-label multi-class relation classification problem:

\begin{equation}
f: (X,e_1,e_2) \to Y
\end{equation}

where $Y = \{y_1,...y_n\}$ is the relation label, $X$ is an input text sequence, $e_1$ and $e_2$ are non-overlapping, continuous token subsequences of $X$ representing the entities between which the causal relation is to be identified, that is:

\color{black}
\begin{equation}
 X=[x_1,x_2,...x_{n-1},x_n] ,
 \end{equation}
\begin{equation}
 e_1 = X[i:j] \,\,\text{with} \,\,i \leq j \,\,\text{and}\,\, i,j \in [1..n] , 
 \end{equation}
\begin{equation}
e_2 = X[k:l] \,\,\text{with}\,\, k \leq l \,\,\text{and}\,\, k,l \in [1..n] ,
\end{equation}
\begin{equation}
j < k \,\,\text{or}\,\, l < i ,
\end{equation}

where the last statement asserts that the entities must not overlap and one must precede the other.

In the following, we present a comprehensive empirical analysis of how the various LLM-based learning strategies presented in Section~\ref{llmMethods} perform in the CRE task, testing across a range of SOTA, open-source decoder-only models, providing practical guidance and actionable insights for researchers and practitioners working with biomedical texts and applications. Moreover, we gauge on the portability of CRE-capable fine-tuned models across different types of textual collections within the biomedical domain.

\section{Datasets}
\subsection{MIMICause}

We benchmark various models and learning strategies to identify causal narratives within clinical notes using the MIMICause dataset~\cite{khetanmimicause}, using its Hugging Face distribution\footnote{\url{https://huggingface.co/datasets/pensieves/mimicause}}.
MIMICause is derived from a collection of de-identified discharge summaries sourced from the MIMIC-III (Medical Information Mart for Intensive Care-III) clinical database~\cite{johnson2016mimic}\footnote{Harvard's DBMI Data Portal: \url{https://portal.dbmi.hms.harvard.edu/projects/n2c2-nlp/}}, which are further annotated for the nine types of biomedical entities illustrated in Table~\ref{tab:bioEnts} with corresponding examples.

The MIMICause annotation schema states that \textit{``a causal relationship/association exists when one or
more entities affect another set of entities''}~\cite{khetanmimicause}.
Eight directed relation types between pairs $e_1, e_2$ of entities are defined, where the order of the entity tags determines the direction of causality: $Cause(e_1,e_2)$, $Cause(e_2,e_1)$, $Enable(e_1,e_2)$, $Enable(e_2,e_1)$, $Prevent(e_1,e_2)$, $Prevent(e_2,e_1)$, $Hinder(e_1,e_2)$, $Hinder(e_2,e_1)$. Additionally, an $Other$ relation class encompasses instances where either a non-causal interaction or no relationship at all exists between a given pair of biomedical entities. Consequently, MIMICause benchmark models CRE as a 9-class relation classification problem, i.e. $Y = \{0,...,8\}$.

The semantics of the four relation labels is the following:
\begin{itemize}
 \item $Cause$: $e_i$ is the direct and primary reason for the occurrence of $e_j$, like in the sample sentence: \textit{``He was treated with $<$e1$>$ATRA$<$/e1$>$ and subsequently suffered from $<$e2$>$ATRA syndrome$<$/e2$>$ with an acute elevation in his WBC to $>$50''}, where the entity ATRA is the causative agent of ATRA syndrome.
 \item 
 $Enable$: $e_i$ facilitates or contributes to the occurrence of $e_j$ in combination with other factors. This relation reflects uncertainty or partial involvement, like in the sample sentence: \textit{``$<$e2$>$Rash$<$/e2$>$- scattered papules on arm and swollen eyelids. Unclear if from levaquin or $<$e1$>$morphine$<$/e1$>$''}, where morphine is suspected to be one of multiple agents that may have led to the rash.
 \item $Prevent$:
 the emergence or application of $e_i$ leads to the eradication, prevention or decrease of $e_j$, like in the sample sentence: \textit{``$<$e2$>$Docusate Sodium$<$/e2$>$ 100mg Capsule Sig: One (1) Capsule PO BID (2 times a day) as needed for $<$e1$>$constipation$<$/e1$>$''}, where the entity Docusate Sodium is intended to prevent the onset of constipation. 
 \item $Hinder$: $e_i$ weakens, delays, or limits the severity or frequency of $e_j$, in combination with other factors. For example, in the sample sentence, \textit{``She was treated with home fentanyl 25mcg patch for pain control, home lidocaine patch with $<$e1$>$morphine$<$/e1$>$ for $<$e2$>$breakthrough pain$<$/e2$>$ Medications on Admission''} a combination of lidocaine and morphine is administered to reduce the severity or frequency of breakthrough pain.
\end{itemize}

\begin{table*}
 \caption{The entity types annotated in the \textit{n2c2 2018 shared task}.}
 \label{tab:bioEnts}
\centering
 \begin{tabular}{l|l}
 \hline
\textbf{Entity Type} & \textbf{Examples} \\
 \hline
 \hline
Drug & \textit{morphine, ibuprofen} \\
Adverse Drug Event (ADE) & \textit{nausea, seizures} \\
Reason & \textit{vitamin K deficiency} \\
Dosage & \textit{2 units, stress dose} \\
Strength & \textit{10 mg, 60 mg/0.6 mL} \\
Form & \textit{capsule, syringe, tablet} \\
Frequency & \textit{daily, twice a day, Q4H} \\
Route & \textit{transfusion, oral, gtt} \\
Duration & \textit{for 10 days, chronic, 2 cycles} \\
 \hline
 \hline
\end{tabular}

\end{table*}

For more details on the definitions of the causal relation schema, refer to the original paper~\cite{khetanmimicause}.

Each row in MIMICause dataset comprises a text, a pair of entities explicitly mentioned in the text, and a causal relation label. Causal relations can link entity pairs within the same sentence or, in rare cases, spanning a few sentences in the input text. These relationships may be explicitly signaled by lexical causal connectives, such as ``due to'', or they may be implicit, requiring inference from the broader context. 

The MIMICause dataset comprises 2,714 examples, with the train-dev-test split in the Hugging face distibution summarized in Table~\ref{tab:labelDistr}.

\begin{table}[htp!]
\caption{Distribution of causal relation labels over train, eval and test splits of the MIMICause dataset. The second column contains the numerical equivalents of relation labels in the adopted Hugging Face distribution.)}

 \centering
 \fontsize{10}{10}\selectfont
 \begin{tabular}{ccccc}
 \hline
MIMICause Relation & HF label & \# Train Instances & \# Eval Instances & \# Test Instances \\
 \hline
 \hline
Cause(e1,e2) & 0 & 254	& 64 &	36 \\
Cause(e2,e1) & 1 & 266		& 67 &37 \\
Enable(e1,e2) & 2 & 126	& 31 &	17 \\
Enable(e2,e1) & 3 & 126	& 32 &	18 \\
Prevent(e1,e2) & 4 & 188	& 47 &	26 \\
Prevent(e2,e1) & 5 & 179	& 45 &	25\\
Hinder(e1,e2) & 6 & 111	& 28 &	15 \\
Hinder(e2,e1) & 7 & 133	& 33 &	19\\
Other & 8 & 570	& 142 & 79 \\
 \hline
 \hline
\end{tabular}
 \label{tab:labelDistr}
\end{table}

\subsection{Adverse Drug Event Dataset}
\label{sec:ade}
A second dataset we use to validate the generalization of our evaluation results is the Adverse Drug Event corpus (hereafter ADE), a popular benchmark of medical case reports, sourced from MEDLINE abstracts and annotated with mentions of drugs and adverse drug events (conditions)~\cite{GURULINGAPPA2012885}. Namely, the distribution we used is the subset named \textit{Ade\_corpus\_v2\_drug\_ade\_relation} from the Hugging Face \text{ADE-Corpus-V2} repository\footnote{\url{https://huggingface.co/datasets/ade-benchmark-corpus/ade\_corpus\_v2}}, which contains 6,821 case report sentences, each annotated with \textit{drug} and \textit{effect} entities, together with their character position within the text.

ADE assumes a binary classification schema of the relations between drugs and adverse events: the events are either effects caused by the drug, or they are not related, with no further sub-categorization of the relation. Moreover, ADE only contains positive instances, that is, examples of causal relation from the annotated \textit{drug} to the annotated \textit{effect}.

To align this binary classification setup with the causal relation classification schema of the fine-tuned models, trained on MIMICause, a relation label mapping is necessary. Specifically, MIMICause defines two types of directed positive causal links from $E_1$ to $E_2$: $Cause(E_1,E_2)$ and $Enable(E_1,E_2)$. The latter refers to situations in which a drug contributes to the occurrence of a condition in combination with other factors, according to MIMICause’s definition of $Enable$. This is treated as a positive causal instance in the ADE dataset, as illustrated in the first two rows of Table~\ref{tab:adeExamples}, where bupivacaine and lidocaine are both contributing factors to methemoglobinemia. 
Conversely, $Prevent(E_1,E_2)$ and $Hinder(E_1,E_2)$ are interpreted as negative links, indicating the absence of a causal relationship between $E_1$ and $E_2$, and so are all inverse-direction relations (e.g., $Cause(E_2,E_1)$, $Enable(E_2,E_1)$, $Prevent(E_2,E_1$)), as well as the $Other$ class.

We will show in Section~\ref{sec:generalization} how inference on ADE examples is performed using prompt-based conversion. Here, we just observe that inference output labels are mapped to 1, if they are equal to $Cause(e_1,e_2)$ or $Enable(e_1,e_2)$, and to 0 otherwise.

\begin{table*}
 \caption{Positive examples of drug reaction case reports from the ADE dataset and two synthetic negative instances. The negative instances in rows four and five are generated by entity pair sampling from the sentence in row three.}
\centering
\fontsize{11}{10}\selectfont
 \begin{tabularx}{\linewidth}{Xccc}
 \hline
\textbf{text} & \textbf{drug} & \textbf{effect} & \textbf{label}\\
 \hline
 \hline
Methemoglobinemia after axillary block with bupivacaine and additional injection of lidocaine in the operative field. & bupivacaine & Methemoglobinemia & 1 \\
Methemoglobinemia after axillary block with bupivacaine and additional injection of lidocaine in the operative field. & lidocaine & Methemoglobinemia & 1 \\
Cutaneous sarcoidosis during interferon alfa and ribavirin treatment of hepatitis C virus infection: two cases. & ribavirin & cutaneous sarcoidosis & 1 \\
Cutaneous sarcoidosis during interferon alfa and ribavirin treatment of hepatitis C virus infection: two cases. &	ribavirin & infection & 0 \\
Cutaneous sarcoidosis during interferon alfa and ribavirin treatment of hepatitis C virus infection: two cases. &	interferon & hepatitis & 0 \\
 \hline
\end{tabularx}
 \label{tab:adeExamples}
\end{table*}

From the original ADE corpus, we generate two test sets. First, we use a 20\% fixed random split of the full ADE\footnote{Notice that ADE does not provide an official train-test split. The index range of the collected split referencing the original ADE dataset is shared at the link \url{https://drive.google.com/file/d/1dIEhzoq_DN1_chKDxQbP2eUA5d0nN7hH/view?usp=drive_link} for full reproducibility.}. 
Additionally, in order to evaluate across both positive and negative classes, we generate a synthetic set of negative examples using the following procedure. First, we collect the sets of all \textit{Drug} and \textit{Effect} entities in the entire ADE corpus. Subsequently, for each example in ADE, we generate all candidate combinations of all \textit{Drug}-\textit{Effect} entities matched within the example's text and compute the set difference with all occurrences of the same \textit{Drug}-\textit{Effect} combination for the same sentence in ADE (positive set). As ADE annotation can not be guaranteed to follow a strict Closed World Assumption, unannotated entity pairs in a sentence can not be assumed \textit{a priori} to be true negatives. Therefore, we manually validated a small fraction of this initial negative sampling and removed error-prone entities\footnote{Namely, entities that do not appear in positive combinations, but are sub-strings of entities that do.}. We found out that upon removal of these spurious cases, ADE annotation is quite systematical and entities in a sentence not tagged in any causal link are actually not linked and therefore usable in negative samples. Consequently, we consolidated a set of 400 silver standard negative examples and finally created a balanced test sample of 800 instances
. Rows 3 through 5 in Table~\ref{tab:adeExamples} illustrate a positive example ($Cause$ or $Enable$ relationship) from ADE and two synthetic negative examples generated from the same sentence, respectively.

\subsection{Drug Review Dataset}
\label{drugReviewDatset}
We created a drug review test set building upon the open-source \textit{Drug Reviews (Druglib.com)} dataset, available within the UCI Machine Learning Repository\footnote{UCI Machine Learning Repository, available at \url{https://archive.ics.uci.edu/}}.

The \textit{Drug Reviews} dataset\footnote{\textit{Drug Reviews}: \href{https://archive.ics.uci.edu/dataset/461/drug+review+dataset+druglib+com}{https://archive.ics.uci.edu/dataset/461/drug+review+dataset+druglib+com}
}, introduced by Gr\"{a}\ss{}er et al.~\cite{10.1145/3194658.3194677} to study sentiment analysis of drug experience, contains patient reviews on specific drugs along with related conditions and was obtained by crawling online pharmaceutical review sites\footnote{The dataset is distributed under a CC BY 4.0 license, which allows using the data for research purposes as well as sharing and adapting it for any purpose.}.

\begin{table*}
 \caption{Sample reviews with target entity metadata from the \textit{Drug Reviews} dataset.}
\centering
\fontsize{11}{10}\selectfont
 \begin{tabularx}{\linewidth}{ccX}
 \hline
\textbf{Drug} & \textbf{Condition} & \textbf{Review} \\
 \hline
 \hline
ciprofloxacin &	urinary tract infection & i had a urinary tract infection so bad that when i pee it smells but when i started taking ciprofloxacin it worked it's a good medicine for a urinary tract infections.\\
ziana & acne & when i first started using ziana, i only had acne in between my eyebrows, chin, and the nose area. my acne worsened while using it and then it got better. but after about 4 months of using it, it became ineffective. so i now have acne between my eyebrows, chin, cheeks, forehead, and the nose area. its great at first but after a while it made my face even worse than before i used the product.\\
nuvaring &	birth control &	i tried the nuvaring. this was my first form of any birth control. this was very easy to put inside and very easy to take out. i didn't feel the ring ever. i thought it was amazing until i started to get huge deep pimples. they were impossible to get rid of. \\
 \hline
\end{tabularx}
 \label{tab:UseCaseExamples}
\end{table*}

In the \textit{Drug Reviews} dataset, the target Drug and Condition metadata entities are not always explicitly mentioned in the review text. For compliance with the instruction prompt setup of our benchmark evaluation in Section~\ref{sec:benchmarkEval}, we first filtered a subset of around 19,200 \textit{Drug Reviews} instances where both Drug and Condition entities are matched within the text. Table~\ref{tab:UseCaseExamples} lists a few examples of reviews from this subset. Subsequently, for our evaluation we collected a random sample of 40 reviews for each possible relation: $Cause$, $Prevent$, $Hinder$, $Enable$ and $Other$, yielding an overall set of 200 relations to be validated.

Then, to evaluate the correctness and directionality of causal relations extracted from the drug reviews by the models, we conducted an annotation exercise involving three annotators per relation type\footnote{The annotators involved in the evaluation were specialists in the Digital Health field and entirely independent and external to this study.}. The coders were instructed to simply read the drug review text and assess whether the output relationship was correct, with options to mark it as \textit{True} if the relation \textit{(E1 causal\_rel E2)} was supported by the text, \textit{False} if not, or \textit{Swapped Entities} if the releation type was correct but direction was opposite (\textit{E2 causal\_rel E1}). 

We calculated the average pair-wise Cohen $\kappa$ inter-rater agreement~\cite{McHugh2012276} of all three raters, resulting in a value of 0.739, as well as the Fleiss $\kappa_F$ agreement~\cite{Falotico2015463}, resulting in a value of 0.728, both indicating a substantial level of agreement among the annotators. For detailed figures see Table ~\ref{tab:annotationRes} in Section~\ref{sec:generalization}.

\section{Benchmarking Learning Methods}
\label{sec:benchmarkEval}

In this section, we first introduce the baseline approaches used for comparison with the learning methods under analysis. Next, we shortly describe the range of LLMs we experiment with. Finally, we present the results obtained for all model configurations and implemented methods.

\subsection{Baselines}

We compared our implemented learning methods with the two baseline architectures employed in~\cite{khetanmimicause}, both leveraging BERT-based text encoders combined with fully connected \textcolor{black}{Feed Forward Network (FFN)} classifier layers. We will refer to them as BERT+Ent and Clinical-BERT+ENT.
In both architectures, the sentence encoding vector is augmented via vector concatenation with the token-average context vectors of the two target entities before being passed through the FFN and softmax classification layers.
Among these baselines, the architecture incorporating the domain-specific Clinical-BERT encoder, denoted as \texttt{Clinical-BERT+Ent} in Table~\ref{tab:results}, yields the best performance\footnote{Note that the Clinical-BERT performance may be slightly overestimated as its embeddings were fine-tuned on text collections including the MIMIC dataset, albeit without explicit causal relation annotations.}.

\subsection{Large Language Models}

We tested and evaluated five main families of state-of-the-art, open-source LLMs with decoder-only architectures, namely Mistral, Llama (ver 2 and ver 3), Gemma, and DeepSeek. 
In order to provide meaningful comparisons between models and to operate within the constraints of limited compute resources, we opted for using small-to-mid-range models across the various architectures. We also applied quantization throughout our experiments, as described in Table~\ref{tab:hyperparametersInference} in Appendix~\ref{sec:appendixCausalityGraphs}. All the models tested in this study are released as open-source (with varying usage licenses) and accessible through Hugging Face via the Transformer library, hereby making our study fully reproducible from a user perspective.
\color{black}

\paragraph{Mistral:}

We used Mistral-7B-v0.1\footnote{\url{https://huggingface.co/mistralai/Mistral-7B-v0.1}}, a pre-trained generative text model with 7 billion parameters, released under Apache 2.0 license. By leveraging \textcolor{black}{Grouped-Query Attention (GQA)~\cite{Ainslie20234895} and Sliding Window Attention mechanisms (SWA)~\cite{beltagy2020longformerlongdocumenttransformer}}, Mistral-7B-v0.1 provides increased inference speed and reduced memory requirements for the decoding of longer token sequences~\cite{jiang2023mistral7b}, while outperforming the Llama-2 13B model with almost half its parameters.
As a base model with no instruct-tuning, we used Mistral-7B-v0.1 for fine-tuning experiments only.

Moreover, we used MistralOrca\footnote{\url{Open-Orca/Mistral-7B-OpenOrca}} for all of our in-context learning evaluations. MistralOrca is a 7 billion parameters model based on Mistral 7B architecture and further instruction-tuned by OpenOrca on a reproduction of the Orca dataset~\cite{mukherjee2023orca}, which focuses on imitating step-by-step reasoning and explanations from larger teacher models~\cite{lian2023mistralorca1}. This makes it particularly suitable for tasks requiring explanation, reasoning, and multi-step problem-solving, such as the Chain-of-Thought and Tree-of-Thought methods.

\paragraph{Llama 2:}
The Meta Llama 2 family is a collection of pre-trained and fine-tuned generative text models, ranging in scale from 7 billion to 70 billion parameters, all with a 4k tokens context length~\cite{touvron2023llama}.
We experimented with instruction fine-tuning on the smallest-sized release of the Llama 2 foundation models, Llama-2-7b\footnote{\url{meta-llama/Llama-2-7b-hf}}, which was pre-trained on 2 trillion tokens of data from publicly available sources.

\paragraph{Llama 3:}

The Meta Llama 3.1 family encompasses a collection of multilingual, pre-trained, and instruction-tuned decoder-only transformer models in 8B, 70B and 405B sizes, all featuring support for extended context lengths (128k tokens) and Grouped-Query Attention.

The Llama-3.1-8B-Instruct model we adopted in our experiments\footnote{\url{https://huggingface.co/meta-llama/Llama-3.1-8B-Instruct}} is optimized for instruction-following tasks and has undergone \textcolor{black}{Supervised Fine-Tuning (SFT) and Reinforcement Learning with Human Feedback (RLHF)~\cite{10.5555/3294996.3295184}} to align with human preferences for helpfulness and safety. 

JSL-MedLlama-3-8B-v2.0 (hereafter referred to as \textit{MedLlama}\footnote{\url{https://huggingface.co/johnsnowlabs/JSL-MedLlama-3-8B-v2.0}}), built upon the Llama 3 8B architecture, is an advanced model developed by the John Snow Labs specifically tailored for medical and healthcare applications, having undergone fine-tuning on extensive medical literature and datasets. By testing this model, we wanted to gauge the impact of domain-specific fine-tuning data on the CRE performance.
\paragraph{Gemma:}
Gemma is a family of lightweight SOTA open models from Google, well-suited for a variety of text generation tasks, including question-answering, summarization, and reasoning~\cite{gemma_2024}.
The gemma-2-9b model we tested\footnote{\url{https://huggingface.co/google/gemma-2-9b}} was trained on an 8 trillion tokens dataset including Web Documents, code, and mathematical texts.
\paragraph{DeepSeek-Qwen-Distill:}
The DeepSeek-R1-Distill-Qwen-32B (referred to as \textit{DeepSeek-Qwen-Distill} in the results Table~\ref{tab:results}), is an open-source, dense LLM based on the Qwen2.5-32B architecture and trained by \textcolor{black}{Reinforcement Learning (RL)} using the output reasoning data of the DeepSeek-R1 model~\cite{deepseekai2025deepseekr1incentivizingreasoningcapability}. DeepSeek-R1 is a Mixture-of-Experts architecture (totaling 671B parameters) that recently gained popularity for reaching outstanding reasoning capabilities using cost-effective RL. By directly fine-tuning the smaller Qwen2.5-32B model on 800k reasoning examples curated with DeepSeek-R1, the reasoning patterns of the larger model have been successfully distilled into the smaller one, achieving new state-of-the-art results for dense models in reasoning tasks like MATH-500 and LiveCodeBench~\cite{deepseekai2025deepseekr1incentivizingreasoningcapability}.

By testing on DeepSeek-Qwen-Distill we aim to assess how powerful, general reasoning capabilities might affect LLM's performance on domain-specific \color{black}CRE \color{black} tasks.

\section{Inference and Learning Strategies}
\label{sec:inferenceStrategies}

To evaluate the ability of the LLMs to detect causal relationships, we designed a set of experiments based on five distinct learning paradigms, namely: i) instruction prompting, ii) in-context learning, iii) prompt chaining, iv) chain-of-thought, and v) instruction fine-tuning.

Prompt-based approaches have become a de facto standard for evaluating pre-trained LLMs across NLP tasks, as are known to effectively leverage the implicit knowledge encoded in LLMs. Moreover, prompt engineering supports high flexibility and reproducibility and allows rapid exploration of task-specific capabilities of LLMs, particularly for structured tasks like CRE~\cite{liu2023pre}. On the other hand, we test on instruction fine-tuning to assess the potential performance gains when adapting model parameters on task-specific data.


A consistent set of inference parameters are used across all inference strategies and models, with the exception of the \textit{max\_new\_tokens} which is adapted to each prompting technique to accommodate intermediate reasoning steps. Inference parameters settings are listed in Table~\ref{tab:hyperparametersInference} in Appendix~\ref{sec:appendixCausalityGraphs}. Essentially, we use greedy decoding to enforce deterministic token generation.

All the scripts for running inference, training and evaluating the LLMs considered in this benchmarking study are made publicly available in a code repository, described in the Appendix~\ref{sec:appendixCausalityGraphs}.

\paragraph{Instruction Prompting}
\label{sec:instPromtp}

We first employed an instruction prompting method that builds upon zero-shot learning for CRE~\cite{Ma_2023}. Our approach aims to harness the background knowledge of the LLM by explicitly defining the semantics of the relation labels in the task. Specifically, we structured the prompt into four key components:
\begin{itemize}
\item \textit{Task Instruction}: This section outlines the task to be solved.
\item \textit{Semantic Definition of Labels}: Here, the numerical labels from MIMICause are mapped to natural language definitions to clarify their meanings.
\item \textit{Output Formatting}: This part specifies the expected format of the output.
\item \textit{Input Data}: The input sentence, along with the target entities E1 and E2, whose relation is to be classified, is provided here.
\end{itemize}

This structured approach ensures that the LLM can effectively interpret and perform the RE task by leveraging both explicit guidance and its inherent knowledge.

Figure~\ref{fig:labelDefinitionsPrompt} provides a detailed illustration of the full prompt used in this method, along with a sample output generated by the model. To further encourage the model to contextualize its reasoning, we conclude the prompt with the instruction, \textit{``Please explain your response''}. This approach aims to elicit the model's understanding of the input context and its decision-making process.

\begin{figure}[!ht]
\begin{tcolorbox}[colback=yellow!5!white,colframe=yellow!50!black,
 colbacktitle=yellow!75!black,fonttitle=\ttfamily\small, title=\textbf{Zero-Shot Prompt}]
\begin{tcolorbox}[fontupper=\ttfamily\footnotesize,fontlower=\ttfamily\footnotesize, sidebyside, sidebyside align=center,lefthand width=0.03cm,frame hidden, colback=yellow!5!white, boxrule=0mm, boxsep=0mm, sharp corners, colframe=white]
\begin{minipage}[t][1.6cm][t]{0.3cm}
 \rotatebox{90}{\parbox[c]{1.5cm}{%
 \centering
 \textbf{Task}
 }}
\end{minipage}

\tcblower
Given a text enclosed in triple quotes and a pair of entities E1 and E2, classify the relation holding between E1 and E2.
\end{tcolorbox}
\begin{tcolorbox}[fontupper=\ttfamily\small,fontlower=\ttfamily\footnotesize, sidebyside, sidebyside align=center,lefthand width=0.03cm,frame hidden, colback=yellow!5!white, boxrule=0mm, boxsep=0mm, sharp corners, colframe=white]
\begin{minipage}[t][1.7cm][t]{0.3cm}
 \rotatebox{90}{\parbox[c]{1.5cm}{%
 \centering
 \textbf{Labels}
 }}
\end{minipage}

\tcblower
The relations are identified with 9 labels from 0 to 8. The meaning of the labels is the following:

0 means that E1 causes E2

1 means that E2 causes E1

2 means that E1 enables E2

3 means that E2 enables E1

4 means that E1 prevents E2

5 means that E2 prevents E1

6 means that E1 hinders E2

7 means that E2 hinders E1

8 means that E1 and E2 are in a relation different than any of the previous ones.
\end{tcolorbox}
\begin{tcolorbox}[fontupper=\ttfamily\small,fontlower=\ttfamily\footnotesize, sidebyside, sidebyside align=center,lefthand width=0.03cm,frame hidden, colback=yellow!5!white, boxrule=0mm, boxsep=0mm, sharp corners, colframe=white]
\begin{minipage}[t][1.5cm][t]{0.3cm}
 \rotatebox{90}{\parbox[c]{1.4cm}{%
 \centering
 \textbf{Format}
 }}
\end{minipage}

\tcblower
Given X the label that you predicted, for the output use the format LABEL: X
\end{tcolorbox}
\begin{tcolorbox}[fontupper=\ttfamily\small,fontlower=\ttfamily\footnotesize, sidebyside, sidebyside align=center,lefthand width=0.03cm,frame hidden, colback=yellow!5!white, boxrule=0mm, boxsep=0mm, sharp corners, colframe=white]
\begin{minipage}[t][2cm][t]{0.3cm}
 \rotatebox{90}{\parbox[c]{1.9cm}{%
 \centering
 \textbf{Data}
 }}
\end{minipage}

\tcblower
Text: ```mg po BID prn agitation in an effort to avoid sedation associated with benzos.'''

E1: ```benzos''', E2: ```sedation''

Please explain your response.
\end{tcolorbox}
\end{tcolorbox}
\begin{tcolorbox}[fonttitle=\ttfamily\small,fontupper=\ttfamily\footnotesize,colback=yellow!5!white,colframe=yellow!50!black,
 colbacktitle=yellow!75!black,title=\textbf{Model Sample Output}]
LABEL: 4 The text suggests that the benzos (a type of medication) cause or lead to sedation, which means that E1 (benzos) prevents E2 (sedation). Therefore, the relation is labeled as 4.
\end{tcolorbox}
\caption{Sample zero-shot prompt with explicit semantic definition of the MIMICause relation labels (top) and a LLM-generated response (bottom).}
 \label{fig:labelDefinitionsPrompt}
\end{figure}

\paragraph{In-context Learning}
\label{sec:iCL}

We implemented the iCL paradigm by integrating the same task instructions used in the instruction prompting and additionally generating in-context examples by randomly sampling $K$ instances from the MIMICause training split\footnote{To ensure robust performance and avoid overfitting to specific training examples, the sampling of in-context examples is performed independently for each test instance inference, using a fixed random seed (42) to ensure reproducibility.}. For a sample illustration of a few-shot prompt, please refer to the Appendix~\ref{sec:appendixCausalityGraphs}.

\paragraph{Prompt Chaining}
\label{sec:promptChainApplication}
We implemented two variants of prompt chaining. The first adapts the SumAsk prompting technique proposed in~\cite{li-etal-2023-revisiting-large}, decomposing the RE task into a 3-step prompt pipeline. This pipeline guides the LLM to:
\begin{enumerate}
\item Generate a summary of the relationship(s) between the target entities;
\item Reformulate each relation label into a Yes/No question about the target entities;
\item Use the summary from step 1 to answer the questions from step 2, returning the label corresponding to the most probable ``Yes'' answer.
\end{enumerate}

In the second prompt chaining variant, we used a simpler 2-step prompt chaining approach. First, we ask the model to identify the relationship between the target entities. Then, we pass the model's output as an input variable to a second prompt, which asks the LLM to return the label that best describes the relationship. 

Links to the full illustrations of both variants are provided in Appendix~\ref{sec:appendixCausalityGraphs}.

\paragraph{Chain-of-Thought}
\label{sec:CoTApplication}

For zero-shot CoT, we append the usual \textit{``Let's think step-by-step''} instruction at the end of a standard prompt.

For few-shot CoT, we tested two variants:
\begin{itemize}
\item \textit{One-shot CoT}: A single reasoning example is randomly sampled from the 9 relation types.
\item \textit{9-shot CoT}: One reasoning example is provided for each of the 9 relation types.
\end{itemize}

To mitigate example bias, we manually curated a pool of 3 reasoning examples per relation category based on the following criteria: i) each example contains a clearly annotated causal relation between entities; ii) sentence structures vary in complexity and length to simulate realistic clinical data diversity; and iii) the reasoning path between entities is interpretable and aligned with the expected relation label. All examples were independently reviewed by two clinical NLP experts to ensure quality and label consistency. During inference, examples were randomly sampled from this pool to ensure diversity.

For the 9-shot CoT, we randomized the order of the reasoning examples, as their presentation order is known to influence prompt performance~\cite{lu-etal-2022-fantastically}.

A sample one-shot CoT prompt illustration is pointed to in Appendix~\ref{sec:appendixCausalityGraphs}.

\paragraph{Instruction Fine-Tuning}
\label{sec:ift}

For instruction fine-tuning, we first transformed the MIMICause training split into instruction prompts, as described in Figure~\ref{fig:istrPrompts}.

\begin{figure}[!htbp]
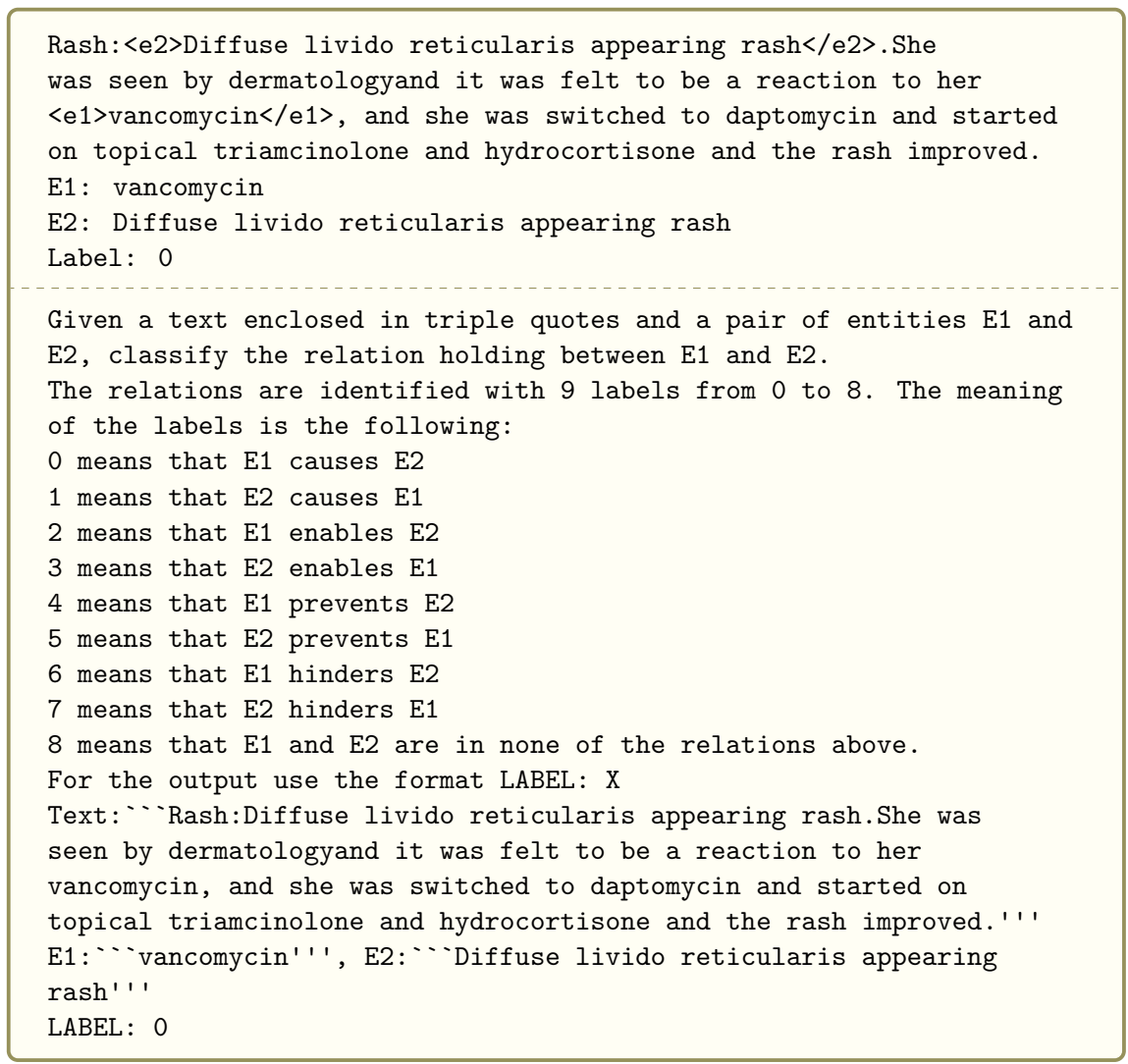

\begin{tcolorbox}[fonttitle=\ttfamily\small,fontupper=\ttfamily\small,fontlower=\ttfamily\small, colback=yellow!5!white,colframe=yellow!50!black]

Rash:<e2>Diffuse livido reticularis appearing rash</e2>.She was seen by dermatologyand it was felt to be a reaction to her <e1>vancomycin</e1>, and she was switched to daptomycin and started on topical triamcinolone and hydrocortisone and the rash improved.

E1: vancomycin

E2: Diffuse livido reticularis appearing rash

Label: 0

\tcblower

Given a text enclosed in triple quotes and a pair of entities E1 and E2, classify the relation holding between E1 and E2.

The relations are identified with 9 labels from 0 to 8. The meaning of the labels is the following:

0 means that E1 causes E2

1 means that E2 causes E1

2 means that E1 enables E2

3 means that E2 enables E1

4 means that E1 prevents E2

5 means that E2 prevents E1

6 means that E1 hinders E2

7 means that E2 hinders E1

8 means that E1 and E2 are in none of the relations above.

For the output use the format LABEL: X

Text:```Rash:Diffuse livido reticularis appearing rash.She was seen by dermatologyand it was felt to be a reaction to her vancomycin, and she was switched to daptomycin and started on topical triamcinolone and hydrocortisone and the rash improved.'''

E1:```vancomycin''', E2:```Diffuse livido reticularis appearing rash'''

LABEL: 0

\end{tcolorbox}

\caption{Transformation of a MIMICause datapoint into an instruction prompt for model fine-tuning.}
 \label{fig:istrPrompts}
\end{figure}

Each generative model was then fine-tuned on this instruction dataset using the \texttt{trainer} class from Hugging Face\footnote{\url{https://huggingface.co/docs/transformers/en/main_classes/trainer}}~\cite{wolf-etal-2020-transformers}. We maintained a consistent set of hyperparameters across all experiments, with the primary settings listed in Table~\ref{tab:hyperparameters} in Appendix~\ref{sec:appendixCausalityGraphs}. 
Since our primary focus is to evaluate CRE performance across various learning methods and pre-trained models, we have not performed extensive hyperparameter optimization.

Given the computational limitations of fully fine-tuning large generative models, we employed the Low-Rank Adaptation (LoRA) technique for Parameter-Efficient Fine-Tuning~\cite{Yu_2023}. Using Hugging Face's \texttt{PeftModel} class, we wrapped the base LLMs into trainable PEFT models. The same LoRA hyperparameter configuration was applied across all experiments and reported in Table~\ref{tab:hyperparameters} in Appendix~\ref{sec:appendixCausalityGraphs}. 
This resulted in a trainable parameter ratio of 0.04 relative to the original model size.

\section{Results}
\label{sec:results}

Table~\ref{tab:results} presents the F1 scores on the test split of the MIMICause dataset \color{black} across all model configurations, in-context example settings, and learning strategies. Given the relatively poor performance of most techniques compared to instruction fine-tuning, we report F1 scores only for the following representative models: 
\begin{enumerate}
 \item \textit{MistralOrca}, the best-performing general-purpose model not specialized in domain-specific text; 
 \item \textit{DeepSeek-Qwen-Distill}, a next-generation model that has established a new state of the art in LLM reasoning capabilities;
 \item \textit{MedLlama}, evaluated using two strategies (9-shot CoT and Prompt Chaining). 
\end{enumerate}

\begin{table}[h!]
\centering
\color{black}
 \caption{F1 performance values on the test split of the MIMICause dataset of various combinations of models, parameter sizes, learning examples, and learning methods, against the two encoder model \textcolor{black}{Baselines (BL)}. The reported Macro F1 values are averaged over the nine MIMICause relation categories. In the Method column, FT denotes Fine-Tuning, SumAsk and 2-Chain correspond to the two implementations of Prompt Chaining, iCL represents In-context Learning and CoT refers to Chain-of-Thought. The best-performing configurations within each general learning category are highlighted in bold.}

{\footnotesize 

\begin{tabularx}{\textwidth}{lp{4cm}XXX|cc}
 \hline
& \textbf{Model} & \textbf{Size} & \textbf{Method
} & \textbf{N examples} & \textbf{Micro F1} & \textbf{Macro F1} \\
 \hline
 \hline
\parbox[t]{2mm}{\multirow{2}{*}{\rotatebox[origin=c]{90}{BL}}} & BERT+Ent & 110M & FT & 1953 & - & 0.54\\

 & Clinical-BERT+Ent & 110M & FT & 1953 & - & 0.56 \\
 
 \midrule
 
 \parbox[t]{2mm}{\multirow{9}{*}{\rotatebox[origin=c]{90}{zero-shot}}} & MistralOrca & 7B & InstPrompt & - & 0.096 & 0.064 \\
 & DeepSeek-Qwen-Distill & 32B & InstPrompt & - & 0.267 & 0.172 \\
 & MistralOrca & 7B & SumAsk & - & 0.449 & 0.343 \\
 & DeepSeek-Qwen-Distill & 32B & SumAsk & - & 0.338 & 0.163 \\
& MistralOrca & 7B & 2-Chain & - & 0.591 & 0.348 \\
& DeepSeek-Qwen-Distill & 32B & 2-Chain & - & \textbf{0.60} & \textbf{0.450} \\
& MedLlama & 8B & 2-Chain & - & 0.304 & 0.210 \\
& MistralOrca & 7B & 0-shot CoT & - & 0.342 & 0.273 \\
& DeepSeek-Qwen-Distill & 32B & 0-shot CoT & - & 0.541 & 0.387 \\
\midrule
\parbox[t]{2mm}{\multirow{5}{*}{\rotatebox[origin=c]{90}{few-shot}}} & MistralOrca & 7B & iCL & 1 & 0.055 & 0.033 \\
& MistralOrca & 7B & iCL & 3 & 0.074 & 0.048 \\
& MistralOrca & 7B & iCL & 9 & 0.089 & 0.073 \\
& MistralOrca & 7B & iCL & 27 & 0.237 & 0.177 \\
& MistralOrca & 7B & 9-shot CoT & 9 & 0.405 & 0.278 \\
& DeepSeek-Qwen-Distill & 32B & 9-shot CoT & 9 & \textbf{0.557} & \textbf{0.489}\\
& MedLlama & 8B & 9-shot CoT & 9 & 0.230 & 0.203\\
\midrule
\parbox[t]{2mm}{\multirow{5}{*}{\rotatebox[origin=c]{90}{fine-tuning}}} & Mistral-7B-v0.1 & 7B & FT & 1953 & 0.817 & 0.780 \\
& MistralOrca & 7B & FT & 1953 & 0.843 & 0.817 \\
& Llama-2-7b & 7B & FT & 1953 & 0.788 & 0.747 \\
& Llama-3.1-8B-Instruct & 8B & FT & 1953 & 0.788 & 0.751 \\
& MedLlama & 8B & FT & 1953 & \textbf{0.847} & \textbf{0.829} \\
& gemma-2-9b & 9B & FT & 1953 & 0.825 & 0.810 \\
 \hline
\end{tabularx}
}
 \label{tab:results}
\end{table}

Overall, fine-tuned models largely outperform encoder-only baselines, achieving substantially higher F1 scores even when starting from less domain-specific pretraining. This highlights their comparatively higher robustness and adaptability to the specialized clinical language.

Moreover, fine-tuned models significantly outperform all other LLM methods. Notably, even the lowest-performing fine-tuned model (\textit{Llama-2-7B}) achieves a macro F1 score that is over 0.25 higher than the best-performing in-context learning configuration. This result is expected, given the domain-specific nature of the causal relation labels. Supporting this, we observed that many responses generated via CoT reasoning attempted to elaborate on the semantics of the relation but still produced incorrect labels, like for the entities \textit{E1:4 days} and \textit{E2:antibiotics} in the text:

\begin{quotation}
 \noindent\textit{``Following 4 days of IV antibiotics, the patient was narrowed to PO amoxicillin for enterococcus plus cipro for serratia.''} 
\end{quotation}

where one model response was:

\begin{quotation}
 \noindent\textit{``The relation holding between 4 days and antibiotics is that 4 days of antibiotics were given to the patient. The closest relation that best describes it is: 4 days enables antibiotics. The label associated to this closest relation is 2. The answer is: LABEL 2.''} 
\end{quotation}

Here clearly the correct label is 8, as there is no relation between the entities \textit{E1:4 days} and \textit{E2:antibiotics}.

As expected, zero-shot methods generally perform poorly. Among them, simple Instruction Prompting yields the lowest micro F1 score (0.096). However, more advanced prompting techniques such as Prompt Chaining and CoT provide substantial improvement.

Specifically, the 2-Chain approach achieves the highest performance among zero-shot techniques ($F1 = 0.45$), outperforming both CoT (0.387) and SumAsk ($0.343$). This suggests that decomposing the classification task into multiple reasoning steps helps  activate latent knowledge from pretraining, leading to more accurate semantic labeling. These findings align with observations from~\cite{jimenez-gutierrez-etal-2022-thinking} and~\cite{li-etal-2023-revisiting-large}.

While CoT shows a similar benefit, its effectiveness is specifically pronounced in models capable of longer reasoning chains and self-corrections, such as~\textit{DeepSeek-Qwen-Distill}.

For instance, given the entities \textit{E1: Oxycodone-Acetaminophen} and \textit{E2: pain} in the following input:

\begin{quotation}
 \noindent\textit{``Oxycodone-Acetaminophen 5-325mg Tablet Sig: One(1) Tablet PO Q6H (every 6 hours) as needed for pain.''}
\end{quotation}

a zero-shot CoT prompt to \textit{MistralOrca} led to this reasoning:

\begin{quotation}
 \noindent\textit{``The relation holding between Oxycodone-Acetaminophen and pain is that Oxycodone-Acetaminophen is a medication used to treat pain. The closest relation that best describes it is: Oxycodone-Acetaminophen prevents pain. The label associated with this closest relation is 4. The answer is: LABEL 4.''}
\end{quotation}

However, in other instances, only~\textit{DeepSeek-Qwen-Distill} was able to infer the correct label through complex reasoning. This is illustrated in Figure~\ref{fig:errorAnalysis} in Appendix~\ref{sec:appendixCausalityGraphs}.

Other observed patterns of erroneous responses include cases where the reasoning sequence is sound, but eventually the mapping to the target label fails, or cases where a label for the inverse of the target relation is output. Examples 1. and 2. below, with their corresponding entities and CoT responses from the \textit{Mistral-7B-OpenOrca} model, illustrate the first and second pattern, respectively:
\begin{enumerate}
\item{E1:\textit{Pneumonia}, E2:\textit{levofloxacin}

Text: \noindent\textit{``Multifocal Pneumonia : She was admitted with multifocal pneumonia. She was started on levofloxacin and vancomycin.''}

CoT: \noindent\textit{``The text says that the patient was admitted with multifocal pneumonia and was started on levofloxacin and vancomycin. This means that levofloxacin is a treatment for pneumonia.So, the correct label is:LABEL: 0''}}

\item{E1:\textit{post-ATN diuresis}, E2:\textit{fluids} 

Text:\noindent\textit{``Patient was given continuous IV fluids and developed post-ATN diuresis complicated by one episode of hypernatremia.''}

CoT: \noindent\textit{``The text says that the patient was given continuous IV fluids and developed post-ATN diuresis.So, the relation between post-ATN diuresis and fluids is that fluids cause post-ATN diuresis.Therefore, the correct label is 0.LABEL: 0''}
}

\end{enumerate}

Furthermore, we find that manually crafting artificial reasoning chains for few-shot CoT prompting provides performance gains primarily for highly capable models such as~\textit{DeepSeek-Qwen-Distill}. For other models, 9-shot CoT does not yield meaningful improvements over the zero-shot setting.

Few-shot in-context learning shows modest gains as the number of examples increases, with the overall performance remaining low. This indicates that LLMs struggle to generalize causal relations from limited in-context examples in the absence of task-specific parameter tuning.

In contrast, fine-tuning allows models to internalize the causal structure of the dataset, moving beyond reliance on general pretraining knowledge. Among all fine-tuned models, \textit{MedLlama}, extensively pre-trained on medical literature and datasets, achieves the best performance. This result suggests that pretraining on domain-specific content helps build stronger representations of clinical language, including abbreviations and acronyms.

Nevertheless, pretraining alone does not suffice for effective \color{black}CRE\color{black}. As shown in Table~\ref{tab:results}, \textit{MedLlama}'s performance in the 2-Chain and 9-shot CoT configurations remains limited.

To assess the statistical significance of these findings, we run a paired random permutation test~\cite{Pitman1937} comparing each fine-tune model with the best non-tuned options, namely DeepSeek-R1-Distil-Qwen-32 B with chain of thoughts (DeepSeek-R1-Distill-Qwen-32B 9-shot CoT) and zero-shot (DeepSeek-R1-Distill-Qwen-32B-chain)\footnote{We use Holm method~\cite{Holm1979ASS} to correct $p$ values for multiple comparisons.}. We found that all fine-tuned models produce significantly different predictions from those of the two DeepSeek models ($p < .01$ for all pairs) while differences between individual fine-tuning models were not statistically significant. Moreover, the two DeepSeek-based methods are not significantly different from each other ($p = 0.3177$), confirming that including in-context examples does not positively impact the model's performance in this specific task.

\subsection{Relation Analysis}
\label{secanal}

Examining more in-depth the classes of extracted MIMICause relations, Figure~\ref{fig:confusionMatrix} in  Appendix~\ref{sec:appendixCausalityGraphs} presents the confusion matrix across the nine classes for the best-performing model (the fine-tuned \textit{MedLlama}). The matrix reveals that most errors stem from semantically related or directionally reversed relations. In particular, $Prevent(e2,e1)$ and $Hinder(e1,e2)$ are occasionally interchanged, reflecting their semantic proximity as inhibitory causal types. Directional pairs (e.g., $Cause(e1,e2)$ vs. $Cause(e2,e1)$) also show moderate confusion, suggesting that interpreting syntactic cues of argument order remains partially challenging for decoder-only LLMs.
Finally, the broad $Other$ category attracts a small fraction of ambiguous instances across all labels.

The box plots in Figure~\ref{boxplot_clssbytrinfq} summarize the distribution of F1 scores (Y-axis) against class frequencies (X-axis), grouped by all the fine-tuned models\footnote{Two classes, \texttt{Enable($E_{1}$, $E_{1}$)} and \texttt{Enable(E2, E1)}, have the same number of instances (126), which explains why there are two plots corresponding to the 126 instances count.}.

\begin{figure*}[!ht]
\centering
\includegraphics[width=0.7\textwidth]{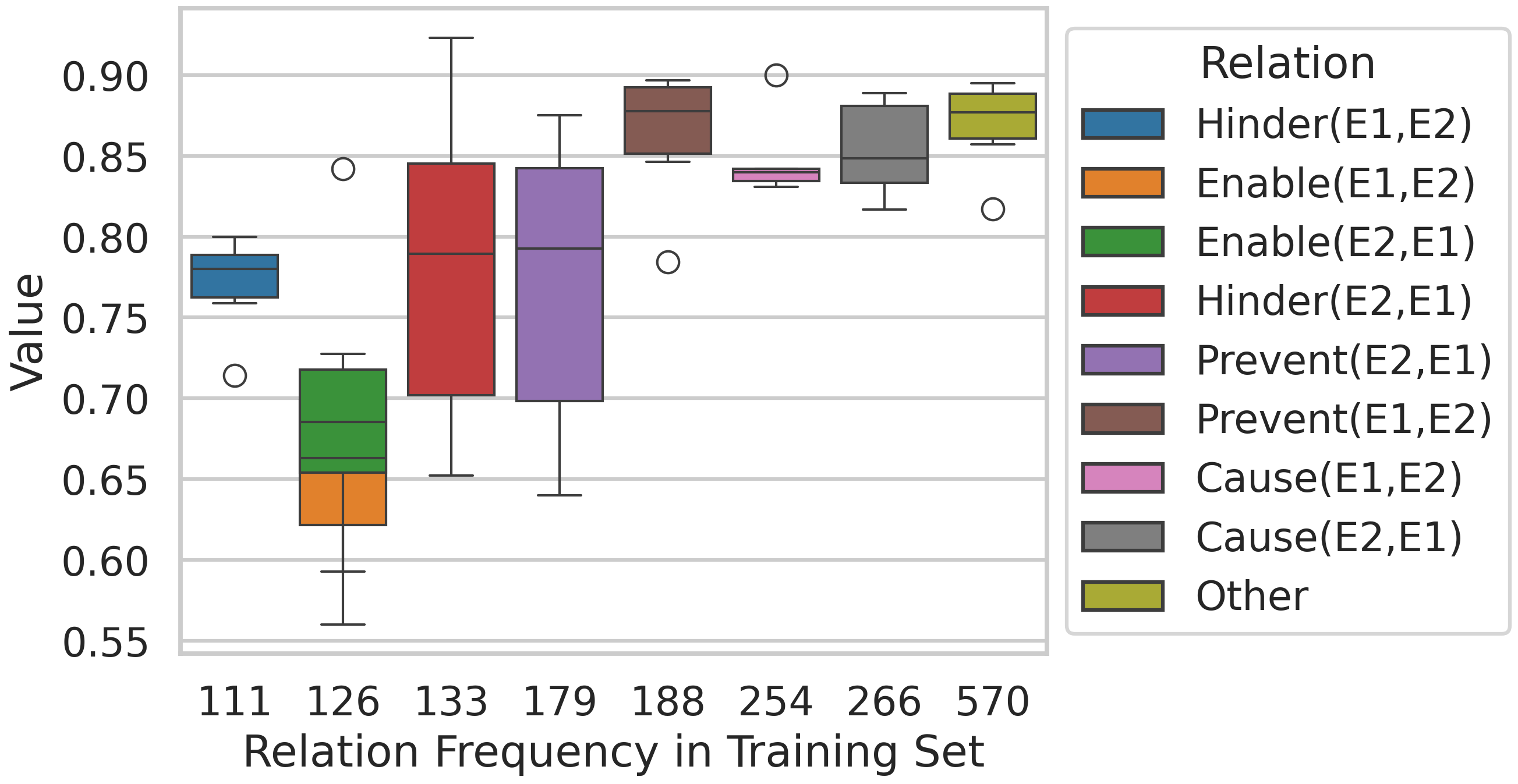}
\caption{Training instance frequency analysis. Distributions of the F1 scores (Y-axis), collapsed by all fine-tuned models, on each of the nine classes in MIMICause, as a function of the each class' occurrence in the train split of the MIMICause dataset (X-axis).}
\label{boxplot_clssbytrinfq}
\end{figure*}

Overall, the plot shows that performance tends to be either low or highly variable before reaching a threshold of approximately 188 instances, possibly reflecting a more significant influence of differing pre-training data on the fine-tuning phase across the different models. Conversely, the low variance and high average performance observed for the classes $Prevent(E_1,E_2)$, $Cause(E_1,E_2)$, $Cause(E_2,E_1)$, and $Other$ on the right-side of the plot (higher frequency relation classes) are coherent with some previous research showing how even simpler models like LSTMs could learn to generalize a combination of lexemes (in this case, the \texttt{($E_{1}$, relation, $E_{2}$)} triple) after observing it for a specif amount of times in the training data~\cite{valvoda-etal-2022-benchmarking}.

In the Figure~\ref{linplot_clssbytrinfq_bymodel} below, each diagram disaggregates the data from Figure~\ref{boxplot_clssbytrinfq} by individual model, with each point in the curve representing the F1 score obtained by a model for different classes, here indicated solely by frequency\footnote{Note that the error bar refers to the \texttt{Enable} relation, which is the only relation in MIMICause for which $|$\texttt{relation($E_{1}$, $E_{2}$)}$|$ == $|$\texttt{relation($E_{2}$, $E_{1}$)}$|$ (see Figure~\ref{boxplot_clssbytrinfq}).}. 

\begin{figure*}[!ht]
\includegraphics[width=0.9\textwidth]{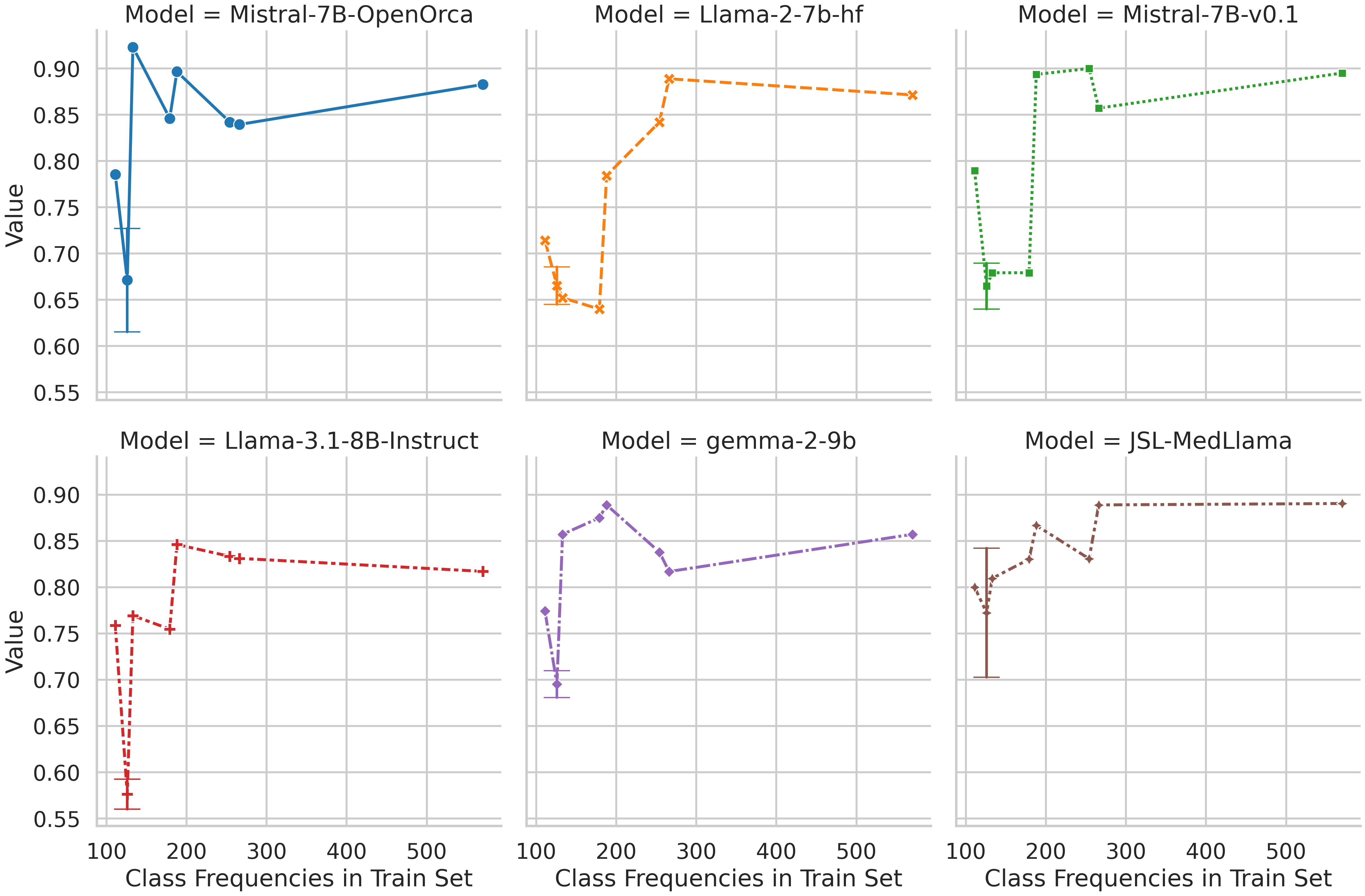}
\caption{Training instance frequency analysis per model: F1 scores (Y-axis) obtained by the different fine-tuned models on each of the nine classes in MIMICause, as a function of the each class' occurrence in the train split of the MIMICause dataset (X-axis).}
\label{linplot_clssbytrinfq_bymodel}
\end{figure*}

The figure helps to better understand how each model exhibits a distinct response pattern, with model performance stabilizing after exposure to a varying threshold centered between 200-300 instances of training data, after which the task becomes notably easier. 

Among the evaluated models, MedLlama demonstrates greater stability and consistency across all classes of MIMICause, with performance levels centered around 0.8 for harder relations such as $Enable(E_1,E_2)$ or $Hinder(E_1,E_2)$, and around 0.9 for relations $Enable(E_1,E_2)$, $Prevent(E_1,E_2)$ and $Other$.

Overall, our analysis suggests that model performance is not directly correlated with the frequency of a specific label in the training set, but rather by the interaction among components within each \texttt{($E_{1}$, relation, $E_{2}$)} triple, as highlighted by the fact that classes sharing the same relation (e.g., $Prevent(E_1,E_2)$ and $Prevent(E_2,E_1)$) exhibit remarkably different results. Additionally, the results do not indicate pure memorization, as performance does not follow a binary 0–100\% pattern but instead fluctuates within the mid-to-high F1 score range.

\subsection{Prompt Design Sensitivity}
\label{promptDesignSensitivity}
 
Additionally, we wanted to explore how the performance within each learning strategy is robust across variations on prompt design choices. Therefore, we conducted a sensitivity analysis 
applied to the instruction prompting and few-shot methods defined in Section~\ref{sec:inferenceStrategies}. For each such method, a set of significant prompt design dimensions is identified and, for each dimension, we apply a range of four diverse variations from the base instruction prompt illustrated in Figure~\ref{fig:labelDefinitionsPrompt}. For each prompt variation, the MistralOrca model is evaluated and micro F1 scores are computed over the MIMICause test split. We also report standard deviation over the range of F1 scores.

While instruction prompting turned out to be the least effective method in Section~\ref{sec:results}, testing on it enables us to gauge on the model robustness to prompt design variations independently from confounding factors characterizing the more complex reasoning chain methods. 
Moreover, the base instruction prompt is the core component of the iCL, CoT and instruction fine-tuning methods implemented in Section~\ref{sec:inferenceStrategies}, so that the present robustness analysis arguably projects to those approaches as well\footnote{Among the two models evaluated with instruction prompting we opted for testing on MistralOrca here given our computational infrastructure constraints.}.

\begin{table}[!ht]
\centering
 \caption{Micro F1 performance scores on the test split of the MIMICause dataset of five variants of the base instruction prompt illustrated in Figure~\ref{fig:labelDefinitionsPrompt} in Appendix~\ref{sec:appendixCausalityGraphs}, for the MistralOrca model. Each row contains prompt design variants with respect to the target dimension indicated in the left column headings. The last column reports standard deviation over those variations.}
{\footnotesize 
\begin{tabularx}{0.9\textwidth}{lp{4cm}cccccccc}
 \hline
& \textbf{Prompt Dimension} & \textbf{v0} & \textbf{v1} & \textbf{v2} & \textbf{v3} & \textbf{v4} & \textbf{Std Dev (\%)} \\
 \hline
 \hline
\parbox[t]{2mm}{\multirow{6}{*}{\rotatebox[origin=c]{90}{instr. prompt}}} & \textbf{Persona} & 0.096 & 0.100 & 0.189 & 0.141 & 0.141 & $\pm28.35$ \\
 & \textbf{Phrasing} & 0.096 & 0.059 & 0.063 & 0.052 & 0.107 & $\pm32.44$ \\
& \textbf{Definitions} & 0.096 & 0.115 & 0.0929 & 0.171 & 0.237 & $\pm43.18$ \\
 & \textbf{Format} & 0.096 & 0.130 & 0.126 & 0.167 & 0.118 & $\pm20.20$ \\
 & \textbf{Explain} & 0.096 & 0.078 & 0.163 & 0.115 & 0.089 & $\pm30.92$ \\
 & \textbf{Ordering} & 0.096 & 0.223 & 0.282 & 0.301 & 0.278 & $\pm35.37$ \\
 \\
 \hline
 \\
 \parbox[t]{2mm}{\multirow{2}{*}{\rotatebox[origin=c]{90}{few-shot}}} &\textbf{Selection} & 0.193 & 0.122 & 0.178 & 0.230 & 0.167 & $\pm22.09$ \\
 & \textbf{Ordering} & 0.078 & 0.115 & 0.089 & 0.141 & 0.223 & $\pm44.75$ \\
 &  &  &  &  &  &  &  \\
\\
\hline
\end{tabularx}
}
 \label{tab:promptVariationAnalysis}
\end{table}

Table~\ref{tab:promptVariationAnalysis} reports the performance scores, where v0 represents the base prompt (no variation), while v1 through v4 denote variations relative to a prompt design dimension. We shortly introduce in the following the meaning and rationale of such dimensions along with some examples. The complete set of prompt variations is made available in the folder \textit{prompt\_analysis} in the code repository (see Appendix~\ref{sec:appendixCausalityGraphs}).

\begin{enumerate}
 \item Persona setting (referred to as Persona in Table~\ref{tab:promptVariationAnalysis}): this is the process of defining a target role of the LLM in a prompt, typically via a system message or a sentence prefixed to the prompt instruction. v0 base prompt does not include a Persona, while we explore a few variants of domain-specific Persona setting clauses, identifying the model either as human-like assistant (e.g. \textit{``You are a clinical NLP expert specializing in medical text annotation''} or as a system (e.g. \textit{``You are a knowledgeable clinical NLP model ... ''}).
 \item Instruction phrasing (Phrasing): We vary the tone and style of the task instructions, from more formal with explicit input markup specification to more direct and informal. 
 \item Label definition phrasing (Definitions): we compare the original label definitions and numerical label mapping in the base instruction prompt with alternative formattings and orderings, as well as alternative wording using lexical synonyms. v4 prompt implements the official, verbose definition of the MIMICause relation labels from the original paper~\cite{khetanmimicause}. 
 \item Output formatting instruction (Format): we vary the instruction for output formatting, testing for json and xml variants, alternating instruction styles as well.
 \item Explanation requirement (Explain): we test with dropping the closing request for output label explanation (``Please explain your response.''), as well as with demanding different levels of detail.
 \item Instruction ordering (Ordering): here, we test alternative ordering of the component parts of the base prompt, namely Task Instruction, Label Definition, Formatting Instruction and Input Data. 
\end{enumerate}

For the few-shot method, we measure robustness with respect to example selection strategies and to the ordering of the selected examples. Throughout these evaluations, we use an optimized prompt deploying the best scoring configurations from the previous analysis and a 9-shot settings, with exactly one example per label (see the exact prompt in the \textit{prompt\_analysis} folder in the code repository). This way, we can test whether performance gain from prompt design optimization propagates also through the few-shot method. The nine examples are selected using five strategies:
\begin{enumerate}
 \item Randomized (v0): one randomized example per label type, sampled independently for each inference instance;
 \item Prototype (v1): selecting, for each relation type, the medoid example among all example embedding vectors for that relation\footnote{Using \textit{all-MiniLM-L6-v2} Sentence Transformer over a concatenation of the example's text and entity strings.};
 \item Max\_Diversity (v2): selecting the nine label representative examples that maximize the sum of pairwise embedding vector distance scores, therefore sampling for variety in the train set; 
 \item Min\_Diversity (v3): selecting the nine label representative examples that minimize the sum of pairwise embedding vector distance scores; 
\item Input\_Similarity (v4): For each inference instance, it selects the nine label instances with highest embedding similarity to it.
\end{enumerate}
Finally, in order to test for example ordering, we fix the set of 9 examples from the Min\_Diversity strategy and re-shuffle the order for each variation v0 through v5.

Overall, we observe that instruction prompt exhibits a relatively high variance across prompt design settings, with standard deviation ranging from 20\% to 43\%. This shows that prompt optimization can significantly boost the performance of this inference method. Nonetheless, the range of performance scores remains below the level of the reasoning-based zero-shot methods evaluated in Section~\ref{sec:results} for the same model.

Label definition and prompt structure ordering are the most effective dimensions of prompt optimization. We find that enriching the definitions of the MIMICause causal relations with synonym and paraphrase expressions elicits the model's pre-training knowledge enhancing its interpretation of the input text. Interestingly, we also observe that moving the input data increasingly closer to the instruction clause within the prompt improves the performance, with the best result obtained when data are prefixed to the prompt (v4). Unsurprisingly, removing the closing request for explanation of the predicted label deteriorates the performance, while strengthening the requests elicits CoT-type output, boosting the results. Finally, while the impact of Persona setting has been observed to be inconsistent for factual Question Answering tasks in an extensive study by~\cite{zheng2024helpful}, we find that it systematically improves F1 scores on our CRE task.

Few-shot method is relatively more robust to example selection strategies, with the only strategy outperforming random selection being Min\_Diversity. We observe that in-context examples do not enhance the model performance even in combination with an optimized prompt, consistently with what found in Section~\ref{sec:results}.

\subsection{Training Hyperparameters}

While fine-tuning proved to be the most effective method for our CRE task, in order to test how robust the method is to variations across hyperparameter settings in the training process, we performed a sensitivity analysis applied to \textit{MedLlama}, the best performing architecture, focusing on three main parameters controlling model instantiation and the training process, namely: LoRA rank, training learning rate and training batch size, indicated respectively as $rank$, $lr$ and $bs$ in Figure~\ref{sensitivityPlots}. The parameters ranged over the values: $rank=[4, 8, 16]$, $lr=[1e-5, 2e-5, 5e-5]$, $bs=[4, 8]$.

\begin{figure*}[!ht]
\centering
\color{black}
\includegraphics[width=0.75\textwidth, keepaspectratio]{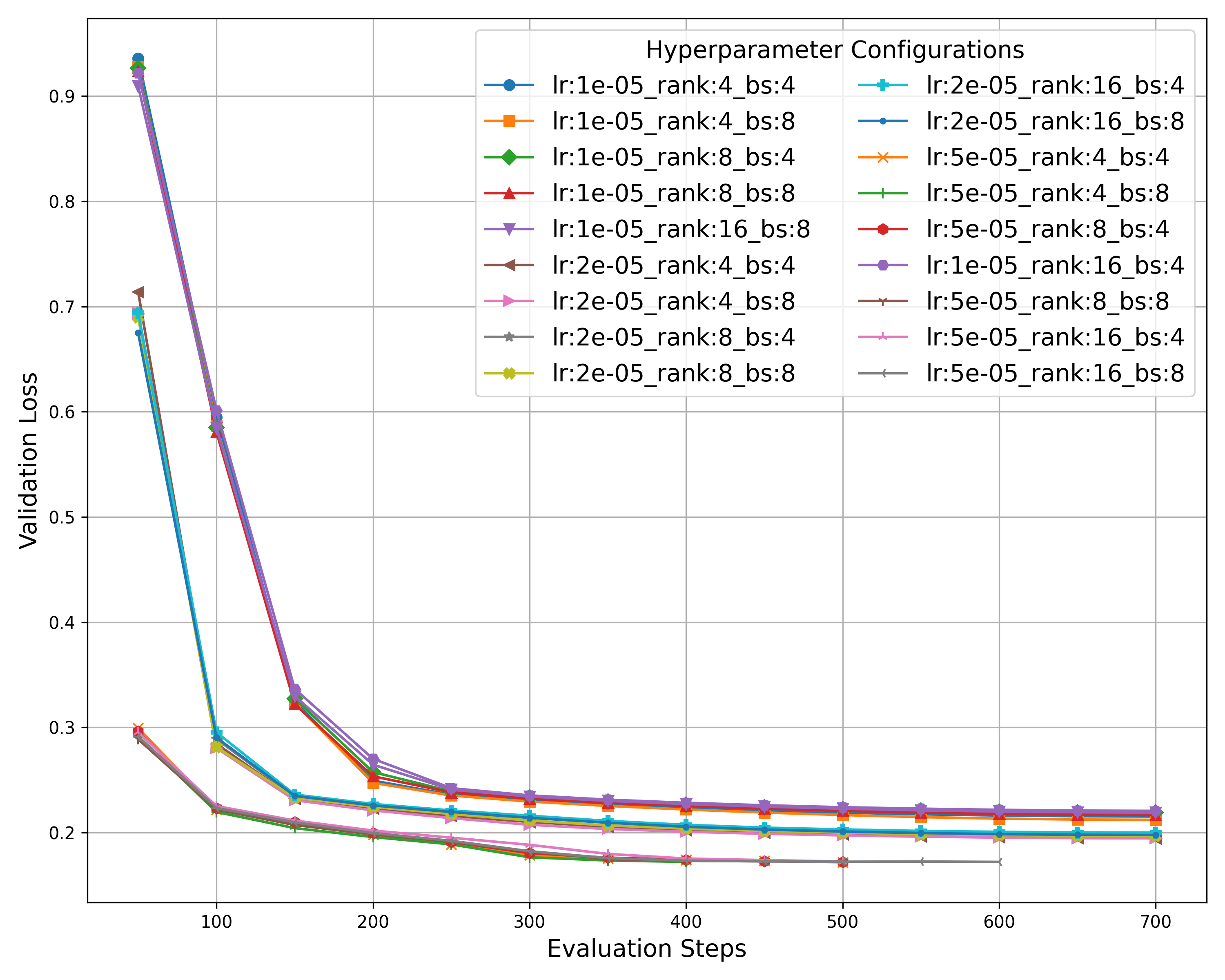}
\caption{Sensitivity analysis of the best performing fine-tuned model, \textit{MedLlama}, to variation over Lora rank, learning rate, and training batch size hyperparameters (respectively $rank$, $lr$, and $bs$ in the Figure). The plots show the validation loss values for 12 intervals of 50 evaluation steps, for a total of 18 model configurations listed.}
\label{sensitivityPlots}
\end{figure*}

To limit computational resource use, we tuned on a subset of 1,000 instances (around 50\%) of the MIMICause train split. We applied a step-based evaluation strategy with an early stopping approach and patience parameter equal to 3 and a \textit{max\_steps} cap of 700.

Figure~\ref{sensitivityPlots} shows the evolution of the loss values on MIMICause validation set (489 instances) of \textit{MedLlama} for 12 intervals of 50 steps (starting from 50 and ending with 700 steps), during training with a total of 18 parameter configurations listed on the top-right corner. We found that, while starting from three clusters of very distant performance levels after the first 50 steps, all model configurations eventually converge towards an interval of $< 5$ points of validation loss, with values ranging from 0.171 to 0.220. We also evaluated the micro F1 scores on the MIMICause test split of the best and worst performing configurations identified in the present hyperparameter search, finding a performance delta of around 0.04 (0.698 vs. 0.661). Although not fully comprehensive, these findings suggest an overall robustness of the fine-tuning method applied to the CRE task.

\section{Result Generalization}
\label{sec:generalization}

We release the best performing model resulting from our study,  renamed as CLiMA (Causal Linking for Medical Annotation) and make it publicly available as LORA adapters, with associated training scripts and hyperparameters settings, in the code repository of this study (see Appendix~\ref{sec:appendixCausalityGraphs}) as well as in the Hugging Face repository: \url{https://huggingface.co/unica/CLiMA}.


We use CLiMA to test the generalization capability of the fine-tuning method to CRE from medical note MIMICause data through the ADE and Drug Reviews datasets.

\paragraph{ADE}

In Table~\ref{tab:ADEeval} we report the F1 scores on the 20\% sample of ADE positive examples and the balanced random sample of positive and negative instances, for all the decoder-only models fine-tuned on the MIMICause benchmark. The performance level of all the models is well aligned with the evaluation on the MIMICause test set, indicating that the causal relation understanding resulting from the instruction fine-tuning generalizes well across partly differing datasets and classification schemas. A slight increase in the F1 scores of some of the models is even noticed, arguably due to the simplified classification task.

\begin{table*}
 \color{black}
 \caption{F1 scores of the fine-tuned models on a 0.2 random sample and a synthetic, 800-sized balanced sample of ADE case reports.}
 \centering

 \begin{tabular}{lccc}
 \toprule
\textbf{Model} & \textbf{20\% ADE} & \multicolumn{2}{c}{\textbf{ADE balanced}} \\
\textbf{} & \textbf{F1} & \textbf{macro F1} & \textbf{micro F1} \\
 \hline
 \hline
 \textit{Mistral-7B-v0.1} & 0.956 & \textbf{0.850} & \textbf{0.851} \\
 \textit{MistralOrca} & 0.959 & 0.766 & 0.772 \\
 \textit{Llama-2-7b} & 0.913 & 0.698 & 0.712 \\
 \textit{Llama-3.1-8B-Instruct} & 0.978 & 0.685 & 0.708 \\
 \textit{gemma-2-9b} & 0.948 & 0.789 & 0.793\\
\textit{CLiMA} & \textbf{0.981} & 0.838 & 0.84 \\

 \bottomrule
\end{tabular}
 \label{tab:ADEeval}
\end{table*}

Although not strictly comparable, all the models outperform the SOTA encoder-only architectures trained directly on ADE data~\cite{hennen-etal-2024-iter} on the positive example prediction, and a few of them also in the balanced test set. Noticeably, some of the domain-general and lower-sized models, such as \textit{Mistral-7B-v0.1}, reach the best score on the balanced test set.

\paragraph{Drug Reviews}
Table~\ref{tab:annotationRes} summarizes the Precision scores of CLiMA on the Drug Review sample described in Section~\ref{drugReviewDatset}, aggregated per causal relation group. As discussed earlier, we calculated a majority vote among groups of three annotators for all 200 causal relationships identified by our algorithm. The human evaluation setup was designed exclusively to validate the correctness of the relationships identified by the model, rather than to identify a complete ground truth of all possible relationships present in the text; therefore, we were only able to compute precision, while we lack knowledge of false negatives required to compute recall or F1-score.

\begin{table*}
 \caption{Precision scores of CLiMA on the \textit{Drug Reviews (Druglib.com)} data sample aggregated for relation groups, together with average pair-wise Cohen $\kappa$ and Fleiss $\kappa_F$ IAA coefficients among human annotators.}
 \centering
 \begin{tabular}{ccccc}
 \hline
\textbf{Relation Type} & \textbf{Cohen $\kappa$} & \textbf{Fleiss $\kappa_F$} & \textbf{Precision} \\
 \hline
 \hline
\textit{Cause} & 0.706 & 0.707 & 0.70 \\
\textit{Enable} & 0.591 & 0.576	&	0.60 \\
\textit{Prevent} & 0.831 & 0.808	&	0.78 \\
\textit{Hinder} & 0.763 & 0.746	&	0.60 \\
\textit{Other} & 0.770 & 0.741	& 0.97 \\
 \hline
\textbf{\textit{Overall}} & 0.739 & 0.728	& 0.73 \\
 \hline
 \hline
\end{tabular}
 \label{tab:annotationRes}
\end{table*}

CLiMA achieves an overall precision of 0.73. If we disregard the directionality of the extracted relationships between pairs of entities, the precision slightly increases to 0.76. In both cases, the obtained precision is quite satisfactory, as it closely aligns with the algorithm's overall performance on the original MIMICause test dataset, for which it was specifically trained. This demonstrates the robustness and generalization capabilities of the fine-tuning method.

Looking at the IAA coefficients, we notice that the raters found annotating the \texttt{Enable} and \texttt{Hinder} relations more challenging. This observation is consistent with the analysis presented in Section~\ref{secanal}.

\section{Drug Reviews Causal Graphs}

Given the successful validation of CliMA on detecting causal relations from drug reviews, we deploy it on the 19,200 instance subset of \textit{Drug Reviews} with metadata entities explicitly matched within the text, and generate a causal drugs knowledge graph (referred to as \textit{CausalDrugsKG}) comprising a total of 19,200 triples, with roughly 3,000 distinct (non-reified) triples, connecting 1,149 unique Drug entities and 322 unique Condition entities via the five causal relation categories, i.e. \textit{Cause}, \textit{Enable}, \textit{Prevent}, \textit{Hinder} and \textit{Other}.
Drug and Condition entities are described within the namespace: \url{http://causaldrugskg.org/causaldrugskg/resource/} (prefix \textit{csldrg}). 

In the ontology designed to describe the KG 
(\url{http://causaldrugskg.org/causaldrugskg/ontology} namespace, with prefix \textit{csldrg-ont}), each extracted claim is reified into an instance of the \textit{csldrg-ont:Statement} class and associated with the collection of drug reviews it was generated from (using the property \textit{provo:wasDerivedFrom}) and the number of source reviews
 (\textit{csldrg-ont:hasSupport}). A sample of generated (un-reified) triples is illustrated in Table~\ref{SampleTriplesCausal}, together with their supports, while Figure~\ref{fig:sampleStatement} illustrates a sample reified statement concerning the resource \textit{causaldrugskg:accutane}

\begin{table}[t]
\centering
 \begin{center}
 \fontsize{11}{10}\selectfont
 \begin{tabular}{|llll|} 
 \hline
 \textbf{Drug} & \textbf{Causal\_Relation} & \textbf{Condition} & \textbf{Support}\\ 
\hline
mirena &	Cause &	birth control &	60 \\
accutane & 	Cause & 	acne	 & 33 \\
viibryd	 & Enable &	depression	 & 5 \\
methotrexate & Enable	 & psoriasis & 2 \\
 lexapro & Prevent & anxiety	& 269 \\
 vyvanse &	Prevent	& adhd	 & 142 \\
pristiq	 & Hinder	 & depression	 & 17 \\
buspar &	Hinder	 & anxiety	 & 11 \\
nexplanon &	Other & birth control & 406 \\
aviane	 & Other &	birth control &	52 \\
\hline
 \end{tabular}
 \end{center}
 \caption{Sample statements for the 5 causal relation categories extracted by the MedLlama model, with their support values in the \textit{Drug Reviews} dataset.}
 \label{SampleTriplesCausal}
\end{table}

\begin{figure}
\begin{flushleft}
\begin{tcolorbox}
\begin{small}
\begin{verbatim}
csldrg-ont:statement_2 a rdf:Statement ;
 provo:wasDerivedFrom csldrg:157784,
 ...
 csldrg:157915 ;
 csldrg-ont:hasSupport 33 ;
 csldrg-ont:object csldrg:acne ;
 csldrg-ont:predicate csldrg-ont:Cause ;
 csldrg-ont:subject csldrg:accutane .
\end{verbatim} 
\end{small}
\end{tcolorbox}
\end{flushleft}
 \caption{A sample reification for the statement assessing a \textit{csldrg-ont:Cause} relation between the instances \textit{csldrg:accutane} and \textit{csldrg:acne}, extracted from 33 reviews (only 2 are shown here for the sake of simplicity.)}
 \label{fig:sampleStatement}
\end{figure}

In order to make our causal KG interoperable with other biomedical resources, we linked the triples' Drug and Condition entity metadata with ontologies from the NCBO BioPortal ~\cite{Ochs2017165}. The NCBO BioPortal is an open, community-developed repository that currently provides access to a library of as many as 1,190 biomedical ontologies and terminologies in subject matters such as pharmacology, public health, etc., and to over 93M class and property mappings across these ontologies~\cite{whetzel2011bioportal}.

The linking is performed by querying the NCBO BioPortal Annotator API\footnote{\url{https://data.bioontology.org/documentation\#nav_annotator}} with each of the Drug and Condition entity terms and limiting the search to a restricted set of reference ontologies. This \textit{ontologies} parameter was determined by intersecting the sets of ontologies returned by querying the NCBO BioPortal Recommender API\footnote{\url{https://data.bioontology.org/recommender}. This endpoint returns a ranked list of appropriate ontologies for a given input text, based on a weighted combination of coverage, acceptance, detail and specialization scores.} with the text of the drug reviews in our dataset.

Overall, 1057 Drug and 268 Condition entities (92\% and 83\%) in the KG 
are mapped onto one or many of 23 drug ontologies and 31 condition ontologies via \textit{owl:sameAs} property, with each Drug and Condition entity being mapped onto an average of 5.6 and 8.6 ontologies, respectively. Table~\ref{ontologyCoverage} shows the coverage of the top 10 ontologies, in terms of number of unique Drug and Condition entities in our causal KG. 
\color{black}
\begin{table}[t]
\centering
 \begin{center}
 \fontsize{11}{10}\selectfont
 \begin{tabular}{|lc|lc|} 
 \hline
 \textbf{Condition Ontology\hspace{3mm}} & \textbf{Coverage} & \textbf{Drug Ontology\hspace{3mm}} & \textbf{Coverage} \\
 \hline 
OCHV &	246 & RXNORM & 869 \\
IOBC &	242 & MESH & 831 \\
MEDDRA & 	232 & NCIT & 812 \\
MESH &	229 & OCHV & 789 \\
MDM &	227 & MDM & 687 \\
SNOMEDCT &	226 & CHEBI & 589 \\
NCIT &	212 & IOBC & 453 \\
NIFSTD &	184 & SNOMEDCT & 353 \\
DOID &	178 & NDDF & 351 \\
MONDO &	178 & LOINC & 294 \\
\hline
 \end{tabular}
 \end{center}
 \caption{Coverage of the top 10 ontologies for Drug and Condition entities in the causal KG. } 
 \label{ontologyCoverage}
\end{table}

We made publicly available\footnote{Under Creative Commons Attribution 4.0 International (CC BY 4.0).} the automatically generated causal KG in Turtle and RDF serialization format within the European Data portal\footnote{\url{https://data.jrc.ec.europa.eu/dataset/acebeb4e-9789-4b5c-97ec-292ce14e75d0}}. 
The direct link is: \url{https://jeodpp.jrc.ec.europa.eu/ftp/jrc-opendata/ETOHA/ETOHA-OPEN/CausalDrugsKG.ttl}. Furthermore, we have set up a Virtuoso SPARQL endpoint where \textit{CausalDrugsKG} can be queried, and analytical information on target entities, attributes, and relations can be
retrieved in user-specified data formats\footnote{\url{https://api-vast.jrc.service.ec.europa.eu/sparql/}}. 
As an example, a SPARQL query like the one in Figure~\ref{fig:accutaneQuery}  returns the four statements from the graph having the target Drug entity \textit{csldrg:accutane} as subject. For each result statement, a link to its corresponding URL in a Virtuoso Faceted Browser endpoint\footnote{\url{https://api-vast.jrc.service.ec.europa.eu/fct/}}
is returned, allowing further navigation. 

\begin{figure}
\begin{flushleft}
\begin{tcolorbox}
\begin{small}
\begin{verbatim}
PREFIX csldrg: <http://causaldrugskg.org/causaldrugskg/resource/>
PREFIX csldrg-ont: <http://causaldrugskg.org/causaldrugskg/ontology#>
SELECT ?statement
FROM <CausalDrugsKG> 
WHERE { ?statement a rdf:Statement .
 ?statement csldrg-ont:subject csldrg:accutane . }
\end{verbatim}
\end{small}
\end{tcolorbox}
\end{flushleft}
 \caption{Sample SPARQL query returning all \textit{CausalDrugsKG} statements with the graph entity \texttt{csldrg:accutane} as \textit{csldrg-ont:subject}.}
 \label{fig:accutaneQuery}
\end{figure}

\paragraph{Data Analytics Dashboard}
\label{sec:dashboard}
We provide aggregated analyses of \textit{CausalDrugsKG} through an interactive visualization dashboard accessible as an Hugging Face Spaces page at 
\url{https://huggingface.co/spaces/zavavan/CausalDrugsKG_Dashboard}. The dashboard's ``Top Key Entities'' panel shows the 30 most frequently occurring Drug and Condition entities in the graph, with over 15\% of the extracted triples (out of the 19,200) having \textit{birth control} as Condition, followed by \textit{pain}, \textit{depression} and \textit{anxiety}. Clicking on the drug and condition names on the right-side legend redirects to the corresponding \textit{CausalDrugsKG} entity page in the Virtuoso Faceted Browser.

The next plot illustrates the distribution of entity types within the KG, specifically focusing on conditions and drugs. This provides an overview of how frequently each entity type appears in the dataset. Another plot highlights the ontology coverage, showing the proportion of entities, again categorized as drugs and conditions, that have been successfully linked to biomedical concepts in external bio-ontologies. This helps assessing which standardized terminologies are represented within the KG and to what extent.

Finally, the ``Causal Relation Chord Diagrams'' panels for each of the five causal relations present the most connected entity pairs within the KG through chord illustrations, with the size of the chords encoding the frequency support of the depicted relations. Figure~\ref{fig:chord} contains a condensed visualization where we pre-filtered the relations with a minimum number of 70 occurrences. In the KG, the two relations with such a support are \textit{Prevent} (in blue) and \textit{Other} (in orange).

\begin{figure}[htp!]
\centering
\includegraphics[width=0.8\textwidth]{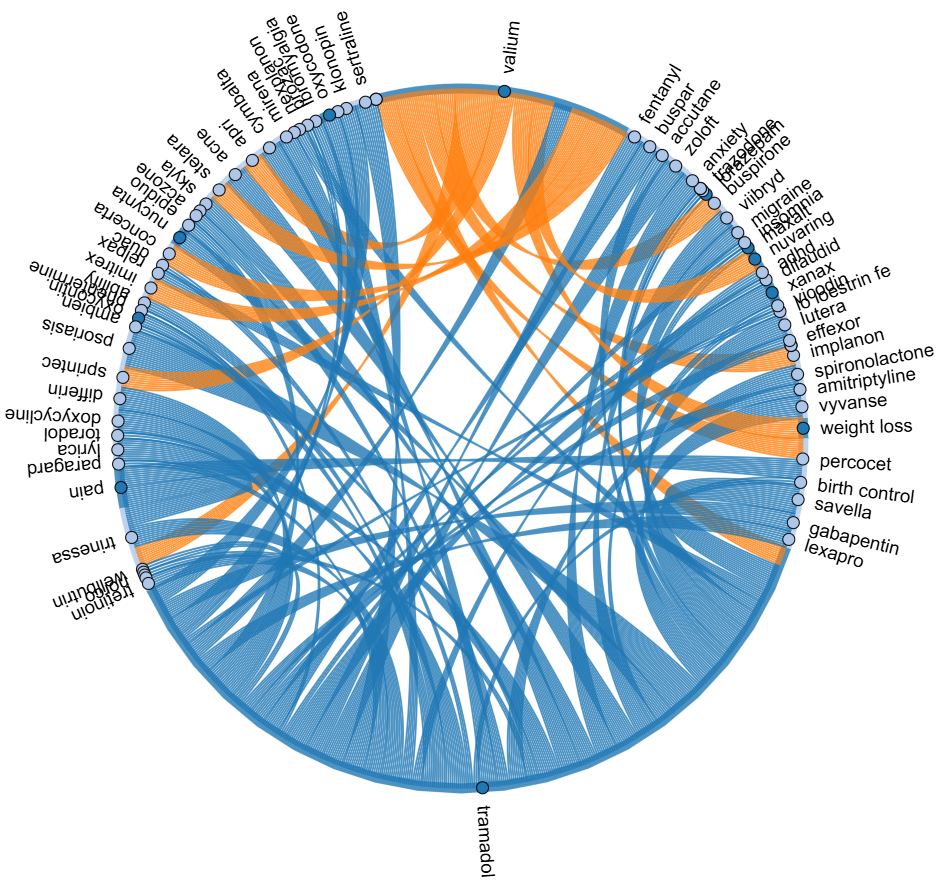}
\caption{Chord diagram illustrating the strength (statement support) of relationships among top Drug and Condition entities in \textit{CausalDrugsKG}.} \label{fig:chord}
\end{figure}

\chapter{Conclusions}
\label{sec:conclusions}

\section{Research results}

The increasing availability of unstructured data in natural language has opened unprecedented opportunities for automatic KG generation systems to extract complex knowledge structures and support actionable data analysis services for a wide range of domains and application scenarios. In this thesis, we experimented with the application of extractive techniques from NLP, ML and generative AI, coupled with SW data linking best practices, to the construction of interoperable knowledge infrastructures, supporting fine-grained data analytics and trend analysis.

For DT monitoring, we presented an unsupervised information extraction pipeline optimized to generate open-domain KGs from micro-blogging text, without relying on a target domain ontology schema in the extraction process.
In a test tweet collection the pipeline proved to outperform some of the state-of-the-art methods, generating highly accurate triples, with around 12\% of entities linked to DBpedia entries. We show the potential usefulness of the generated KG infrastructure for tracking relevant entities in the DT ecosystem, for example as a knowledge plug-in for RAG interfaces. When transferred to a news corpus, the algorithm has shown to scale linearly with the document set size and enabled the generation of a large scale KG of Digital Health, with around 8\% of the 86k extracted entities linked to DBpedia entries of domain relevant types like \textit{dbpedia:Disease}, \textit{dbpedia:Company} and \textit{dbpedia:Drug}. A preliminary data analytics interface to the graph highlighted its potential for generating insights on trends and key players in the digital health sector.

In the AECO research use case, we proved that leveraging deep text understanding abilities of instruction-tuned LLMs enables to customize an existing Information Extraction pipeline to out-of-domain data, with minimal annotation effort. Moreover, when applied to a large dataset of scientific papers, partitioned by an optimized neural topic model, the resulting KG generation pipeline allows the detection of insightful trends on predominant Tasks and Methods, specific to major macro-areas in the domain.
The integration of an upstream topic model is beneficial for a fine-grained research trend analysis, as it makes it more sensible to low signals which otherwise would not emerge at the full collection level. As an example, computing triple statistics for the macro-topic ``Energy Efficiency and Thermal Comfort in Building Environments"  allows to detect as predominant methods highly specialized tools such as the energy simulation package EnergyPlus™ or the 3D software model Envi-met, along with more generic concepts such as Information Technology. To the best of our knowledge, this establishes a novel framework that could be transferable to the large scale exploration of other research fields. 

Finally, by extensively benchmarking LLM architectures and learning paradigms, we demonstrated that instruction fine-tuned models can reach strong performance levels on detecting causal relations from multi-type entities in biomedical text. We found that, while domain-specific pretraining enhances model capabilities compared to general purpose LLMs, medical LLMs show improved performance when they can leverage the complex causal structures specific to the train datasets. At the same time, these models seem to generalize a notion of causality across varying contexts, as it is observed by the robust performance across clinical notes, medical case reports and drug reviews. This highlights their potential for diverse biomedical applications. In fact, we designed an end-to-end pipeline for constructing a KG of drug-condition causal relationships mined from patient-authored drug reviews, deploying an LLM trained on the MIMICause dataset and off-the-shelf entity linking libraries.

Overall, these results help us addressing the research questions from Section~\ref{challenges}:

\textbf{Q1}:\textit{How NLP and Semantic Web technologies can be combined to extract knowledge from noisy user-generated text collections and represent it in interoperable formats?} 

The DT monitoring use case showed that a suitable combination of lightweight pre-processing, standard NLP tools (like dependency parsing) with a word embedding-based clustering approach, enables to generate and link valid and informative triples from noisy social media posts, using no training data and minimal system engineering.

\textbf{Q2}:\textit{Which NLP and ML techniques better fit different application scenarios?}

Use cases 2 and 3, while quite heterogeneous, both suggest that, in scenarios when a target entity-relation schema is provided, tuning pre-trained LLMs on labeled data, no matter how sparse, is the most effective option. In challenging domains such as clinical notes, this method largely outperforms SOTA encoder architectures (BERT) and in-context learning techniques\footnote{As LLM training techniques, particularly for reasoning-intensive inference, are evolving fast, we underline that our conclusions are to be taken ``as of" the time of our latest evaluation run, i.e. January 2025.}. In low-resource use cases like number 2, manually annotating a small training set is cost-effective compared to relying on zero- or few-shot model capabilities.

However, in open domain scenarios with no schema definition and no labeled data like use case 1, syntactical NLP methods are a scalable solution to bootstrap an emerging set of entities and relations.

\textbf{Q3}: \textit{How can KG representations enhance the analysis of trends within a specific domain?}

As we illustrate in the use case 2, KGs support the creation of trend analysis that are grounded on semantically transparent relations and can be tracked back to the text documents that generated them, like the \textit{Task}-\textit{Method} time series in Chapter~\ref{sec:aeco}.

\textbf{Q4}:\textit{Can Generative AI techniques support the construction of scientific knowledge graphs of very abstract research concepts in technical domains, with only limited customization?}

The evaluation of the \textit{SKG-AECO} pipeline in Section~\ref{sec:skg-aeco} proved that, by tuning a small- to mid-sized foundational LLM on sparse, high-quality labeled data, and by linking to suitable sections of large multi-domain KBs like Wikidata and DBpedia, a large scale graph of research concepts in a multidisciplinary and highly technical domain such as AECO can be built and made partially interoperable.

\textbf{Q5}:\textit{Which Generative AI techniques and LLM architecture and training methods better adapt to the generation of causal graphs from medical texts?}

As we observed already, instruction fine-tuning shows stronger performance over few-shot learning or reasoning-intensive prompt techniques like CoT or prompt chaining, with also higher stability across relation labels.  

\section{Limitations and future developments}
\label{futureDevs}

We finally discuss a few general limitations affecting some of presented use cases, outlining also directions of future developments.

\subsection{Digital Transformation Monitoring}

The task configuration of the DT monitoring use case limits a full exploitation of the generated data. As the generation pipeline for the \textit{DTSMM\_KG} graph does not rely on an ontology specification of the target domain for tailoring the entity and relation extraction process, the extracted entities\footnote{Namely, the \textit{emerging} entities that are not linked to DBpedia and can not inherit DBpedia type categories.} are currently not type-classified, which prevents the execution of more structured queries on the graph.

Therefore, we aim to work on an enhanced version of the pipeline that builds upon the entity and relation spans generated with the current approach and further categorizes them into a fine-grained schema tailored to the specific domain, leveraging contextual embedding vector representations.

For the \textit{DHNEWS\_KG}, we are considering extending the entity linking to technology KGs in the health domain, such as the recently released PKG 2.0~\cite{xu2025pubmed} that integrates data sources like research papers, patents and research projects. The analytical goal here is to detect insights about the societal and market impact of trending technologies, as mirrored in the news discourse, and reveal interesting patterns in the innovation process.

\subsection{Scientific Knowledge Graphs}

A recurrent limitation of KG generation pipelines is  triple data sparseness.

We illustrate this in  Figure~\ref{fig:tripleSupportDistribution}, showing the log scale triple distribution over triple support levels, i.e. the number of documents from which a triple was extracted. One can notice as most of the data points cluster around the average support value (1.15) with a long tail stretching towards higher values.

Triple data sparseness turns out to be a significant bottleneck for the research trend analysis in Section~\ref{sec:trends}, which requires the triples extracted by the IE module to be not only semantically correct but also abstract enough to detect a representative sample of triple instances over time.

\begin{figure*}[!ht]
 \centering
 \includegraphics[width=0.9\linewidth]{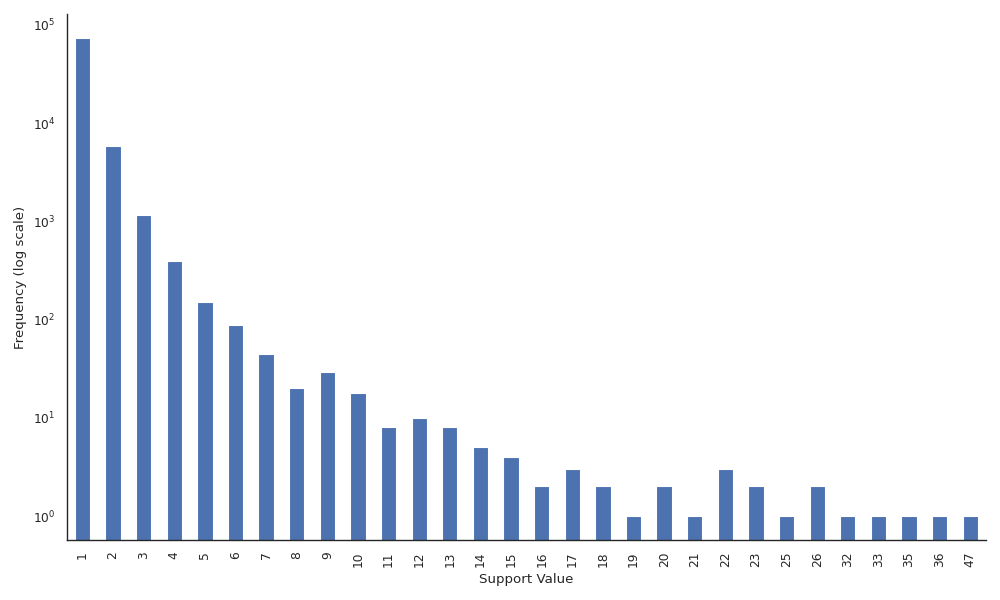}
 \caption{Log scale triple distribution over support, for the $\langle Method;Used-for;Task\rangle$ and $\langle Method;Used-for;Method\rangle$ triple set used in the trend analysis.}
 \label{fig:tripleSupportDistribution}
\end{figure*}

This right-skewed distribution is mainly due to poor entity generalization, with triple subject and object entity average support ranging between 4.43 and 4.91.
In the case of the \textit{SKG-AECO} pipeline, this stems from a sub-optimal performance of the entity merging mechanism described in Section~\ref{sec:skg-aeco}. In fact, to reduce computational complexity, the Transformer-based merging only attempts to merge candidate entities that share at least one token, which prevents to merge near-equivalent concepts such as, e.g., \textit{``photovoltaics"},\textit{``solar pvs"} and \textit{``photovoltaic system"}. On the other hand, the lowered similarity threshold we applied to boost entity merging generates inaccurate merging, as discussed in Section~\ref{eval}. 
An alternative solution we are currently exploring consists of: 1. fine-tuning the underlying \textit{paraphrase-distilroberta-base-v2} entity embedding model to our target domain by contrastive learning (using a subset of highly reliable merge pairs as training data); 2. computing an optimized clustering on the entire set of candidate entities based on this domain-tuned model.

Another possible reason lies in the low recall of the LLM-based triple extraction module we integrated. In fact, the SciERC AECO dataset we curated for training the LLM module is based on a simplified SciERC schema that does not include entity co-reference resolution links. Given that we only process title and abstract of the research papers, this might significantly reduce the recall rate of the triple extraction process. As an example, in the abstract passage below the link between the Method \textit{``Intelligent Lighting Control System"} and Task \textit{``corridor lighting"} is missed because the cross-sentence co-reference to the Method entity (\textit{``The system"}) is not recognized:

\begin{example}
\textit{``Design of Intelligent Lighting Control System Based on Vivado Environment. The system can be applied to corridor lighting inside and outside, which includes [...]''}.    
\end{example}

We plan to extend the annotation of the SciERC AECO dataset with entity co-reference and re-train the LLM module on this new release, testing for the tradeoff between the recall gain and the error rate that this might introduce.  

Upstream of the \textit{SKG-AECO} pipeline, one source of the triple sparseness issue is the relatively low size of the input clustered data.
One limitation of our approach is that, differently than for relation clustering in Section~\ref{relMapping}, we optimize topic clustering based on topic coherence metrics only, without penalizing for the ratio of outlier data points discarded by HDBSCAN, therefore ending up by filtering out a significant fraction of the original data.

In order to characterize the relation between topic modeling quality and data representativeness, Fig.~\ref{fig:plotClusteredRatioCoherenceScore} in Appendix~\ref{sec:appendixAECO} shows the variation of average topic coherence with respect to the ratio of clustered data points\footnote{That is, 1 - outlier\_ratio.} for the 10,000 random document sample of Section~\ref{sec:topicModeling}, over variations of the HDBSCAN hyperparameters. One can notice that the most coherent topics are stably discovered within a 25-40\% range of outlier ratio, while coherence starts degrading when outlier ratio drops towards 0. This indicates a clear trade off between discovering compact, high-quality clusters and clustering higher portions of the data.

For application scenarios where accurate topic-based analysis is prioritized, a baseline option of simply discarding outlier documents may be applied, which can lead to data sparseness problems for the downstream cluster-based processing, depending on the size of the source dataset. The methodology employed in our experimental study may consequently limit the representativeness of the downstream analyses with respect to the original dataset.

However, this limitation can be mitigated by for example re-assigning outlier articles to discovered topics without re-computing the optimized topic model, by performing outlier topic imputation methods (e.g. based on best matching c-TF-IDF topic representations or embedding vector cosine distance). Further empirical testing is required to assess the impact of either approach on the topic representativeness of the downstream bibliometric and information extraction analyses.


Finally, a major line of development of the present AECO research graph is a further integration of heterogeneous graph representations of the same domain.  For instance, while not described in this thesis, we performed a bibliometric co-authorship analysis on the same AECO paper dataset, resulting in a graph of weighted collaboration triples between research institutions (see journal publication ii. in the Dissemination section). Integrating these sub-graphs would allow for example to selectively retrieve co-authorship relations relative to subject matters like the Method entities extracted from co-authored papers, hence supporting advanced queries like: \textit{``Retrieve institutions collaborating on the application of method X for solving Task Y''}. Full integration of these relations into a comprehensive AECO research graph will contribute to a more fine-grained picture of the research innovation process.

\subsection{Causality Graphs from biomedical text}

The main limitation of the causal KG experiments relate to the limited generalization of the models' causal understanding.

While the fine-tuned MedLlama model demonstrated strong performance across both the ADE and Drug Reviews datasets, it was primarily trained on the low-sized MIMICause dataset, which may not encompass the full spectrum of linguistic nuances present in broader biomedical texts. Additionally, the model's lower performance on the less frequent \texttt{Enable} and \texttt{Hinder} relations, indicates a need for larger and more diverse annotated datasets to improve model robustness and the learning of these complex relations. Furthermore, we fine-tuned models on a narrow domain due to the scarcity of available gold-standard datasets for training. This specificity might limit the model's applicability to broader biomedical contexts. To enhance generalization, future research could focus on developing gold-standard datasets from a wider array of health domains, allowing for the inclusion of more nuanced and fine-grained causal relations. This expansion would enable models to better capture the complexity and diversity of causal relationships in various biomedical fields.

We also acknowledge that our approach is primarily focused on the classification of causal relations rather than on deeper causal reasoning. While our models effectively extract causal links from biomedical texts, they do not yet engage in causal reasoning, which involves understanding the underlying mechanisms and implications of these relations. Expanding our work to include causal reasoning would enable the models to provide insights into the causal pathways and potential outcomes, offering a more comprehensive understanding of biomedical phenomena.

Additionally, our exclusive focus on relation classification problem left unsolved the task of biomedical entity detection and linking, which is a crucial building block of an end-to-end causal KG construction pipeline, and a challenging tasks due to the variety of technical terminologies, abbreviations, commercial labeling and even slang expressions found in clinical and patient-authored text collections~\cite{wei2016assessing}. We plan to test with integrating existing biomedical NER algorithms for scaling the coverage of our prototype \textit{CausalDrugsKG} graph. 

In particular, we are currently benchmarking the applicability of causal graph generation pipelines for a use of pharmacovigilance on Adverse Drug Events from crowdsourced patient-authored reviews on off-label drug prescriptions.

Finally, we would like to conclude with a consideration on the security and privacy protection of the used data and the released models. Please note that all datasets used in this study are publicly available and have been de-identified in accordance with privacy regulations prior to their release. Namely, MIMICause, derived from MIMIC-III, underwent rigorous de-identification procedures including removal of protected health information (PHI) following HIPAA Safe Harbor guidelines\footnote{\url{https://www.hhs.gov/hipaa/for-professionals/special-topics/de-identification/index.html}}, while the ADE corpus was constructed from anonymized MEDLINE case reports, and the Drug Reviews dataset contains only user-generated content without personal identifiers.

Regarding our released fine-tuned models, we acknowledge potential memorization risks inherent in large language models trained on clinical text. To mitigate these concerns, we recommend that users of our released checkpoints: (1) apply additional privacy-preserving techniques when deploying models in production environments, (2) avoid training on or exposing the models to any non-de-identified patient data, and (3) implement appropriate access controls and usage monitoring. Furthermore, our models should be used strictly for research purposes and require proper ethical review before any clinical deployment. We emphasize that while our training data is de-identified, users must remain vigilant about potential re-identification risks and ensure compliance with local data protection regulations (e.g. GDPR\footnote{\url{https://eur-lex.europa.eu/eli/reg/2016/679/oj/eng}}) in their specific deployment contexts.

\bibliographystyle{alpha}
\bibliography{biblio}

\clearpage\newpage

\appendix
\addcontentsline{toc}{chapter}{Appendices}
\renewcommand{\thesection}{\Alph{section}}
\renewcommand{\thesubsection}{\Alph{section}.\arabic{subsection}}

\section{KNOWLEDGE GRAPH CONSTRUCTION METHODS}
\label{sec:appendixKGConstruction}

\begin{table*}[!ht]
\centering
\small
 \begin{tabular}{cm{3cm}m{5cm}m{3cm}}
 \toprule
  &  \textbf{Hyperparameter} &  \textbf{Explanation}  & \textbf{Value Range}\\
  \toprule
 \multirow{3}{*}{\rotatebox[origin=c]{90}{UMAP}} & \textit{n\_neighbors}  & the number of neighboring data points UMAP will look at when attempting to learn the manifold structure of the data & [5,15,30]\\

& \textit{min\_dist} &  the minimum distance apart that points are allowed to be in the low dimensional representation & [0.0,0.1]\\

&  \textit{n\_components}  &  the dimensionality of the reduced space & [2,5,10,30] \\
    \hline
\multirow{3}{*}{\rotatebox[origin=c]{90}{HDBSCAN}} & \textit{min\_cluster\_size} & affects the final number of generated clusters & [5,10,20,50,100,200]\\
  & \textit{min\_samples}   &  the minimum number of points required in the neighborhood of a point for it to be considered a core point & [None,1,2,5]\\
   &  \textit{metric}  &  used to calculate the distances & ["euclidean","cosine"] \\
    \bottomrule
\end{tabular}
 \caption{Main hyperparameters searched for upon optimizing the UMAP-HDBSCAN interaction.}
 \label{tab:UMAPHDBSCANHyperparameters}
\end{table*}

\clearpage\newpage
\section{DT MONITORING} 
\label{sec:appendixDigitalTransforamation}

\paragraph{Code Repositories:}

\url{https://github.com/zavavan/dtm_kg}
\\
\textbf{Main Programming Languages}: Python
\\
\textbf{External Libraries}: SpaCy, Scikit-learn, NetworkX, hdbscan, umap, Stanford CoreNLP, NLTK, Pandas 
\\ 
\noindent\rule{\textwidth}{0.4pt}

\begin{table}[htp!]
\centering
 \begin{center}
 \begin{small}
 \begin{tabular}{|c|c|c|} 
 \hline
 Embedding Model & Silhouette $\cdot$ Clustered Ratio & Num Clusters \\ \hline
 BERT & 0.9387 & 1107 \\
BERT & 0.9287 & 918 \\
 BERT & 0.9171 & 1063 \\
 Sentence-BERT & 0.6852 & 869 \\
 Sentence-BERT & 0.6794 & 978 \\
 Sentence-BERT & 0.6767 & 1050 \\
 GloVe & 0.6505 & 327 \\
 GloVe & 0.6362 & \textbf{332} \\
 GloVe & 0.6345 & 511\\
\hline
 \end{tabular}
 \end{small}
 \end{center}
 \vspace{-0.4cm}
 \caption{\color{black}The table presents clustering score values and the number of output clusters for the top three performing UMAP-HDBSCAN configurations across three tested embedding models. It's worth noting that the dataset comprises a total of 29,335 relation instances for contextualized BERT and Sentence-BERT embeddings. In contrast, for static GloVe embeddings, we consolidated single occurrences of each relation form, resulting in a final set of 2,539 relations due to their context-independent vector representations.} 
 \label{alternativeEmbeddings}
\end{table}

\clearpage\newpage
\section{AECO}
\label{sec:appendixAECO}

\paragraph{Code Repository:}

\url{https://github.com/zavavan/AECO_KG_Pipeline}.
\\
\textbf{Main Programming Languages}: Python
\\
\textbf{External Libraries}: BERTopic; Hugging Face's Transformers, PEFT, Datasets, Evaluate; VOSViewer; SpaCy, Pandas, Brat
\\
\noindent\rule{\textwidth}{0.4pt}

\begin{figure*}[!ht]
 \centering
 \includegraphics[width=0.7\textwidth]{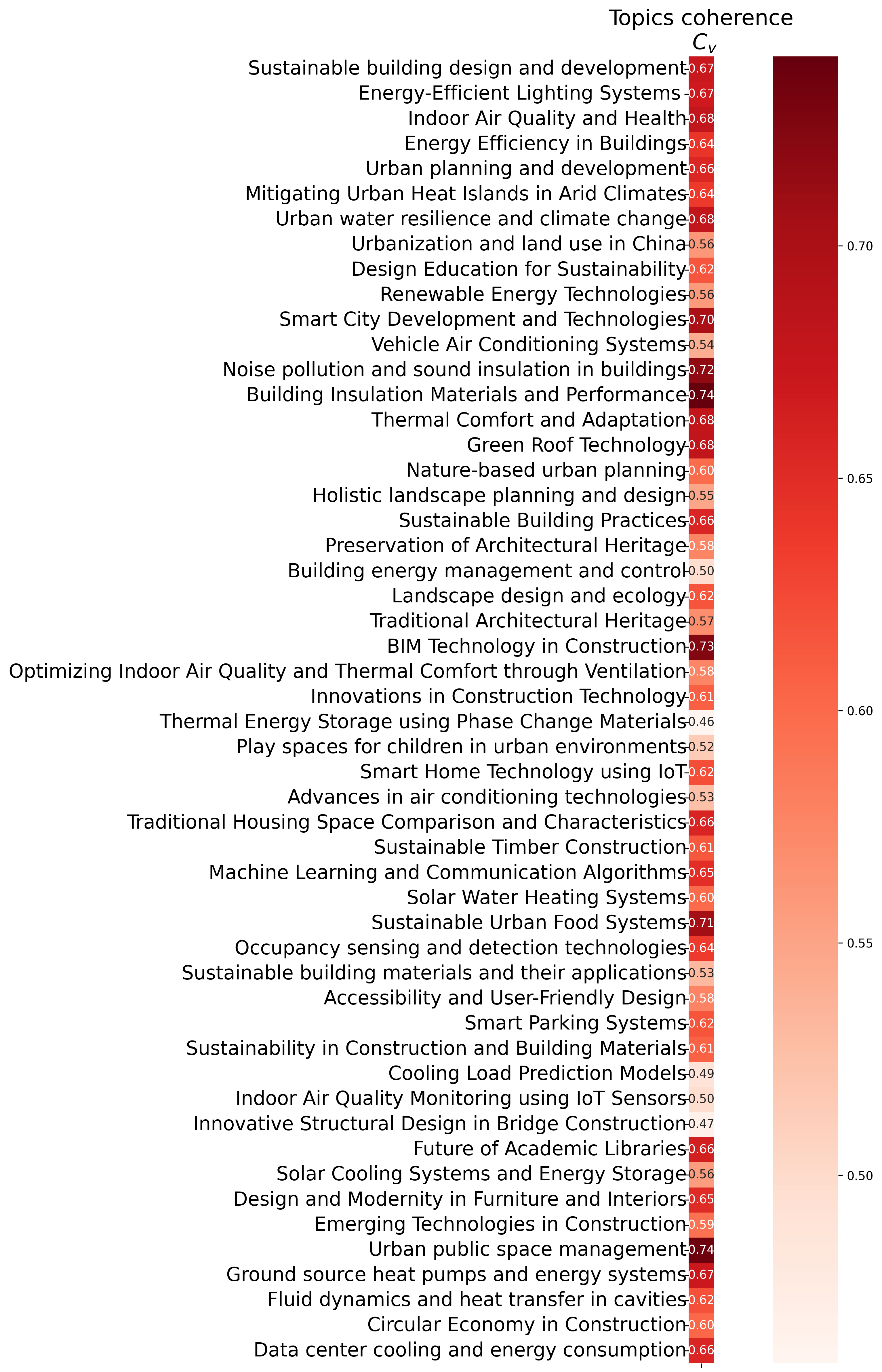}
 \caption{Topic Coherence score heat map for the optimized topic model.}
 \label{fig:heatMap}
\end{figure*}

\begin{figure*}[!ht]
 \centering
 \includegraphics[width=0.8\textwidth, height=0.7\textheight, keepaspectratio]{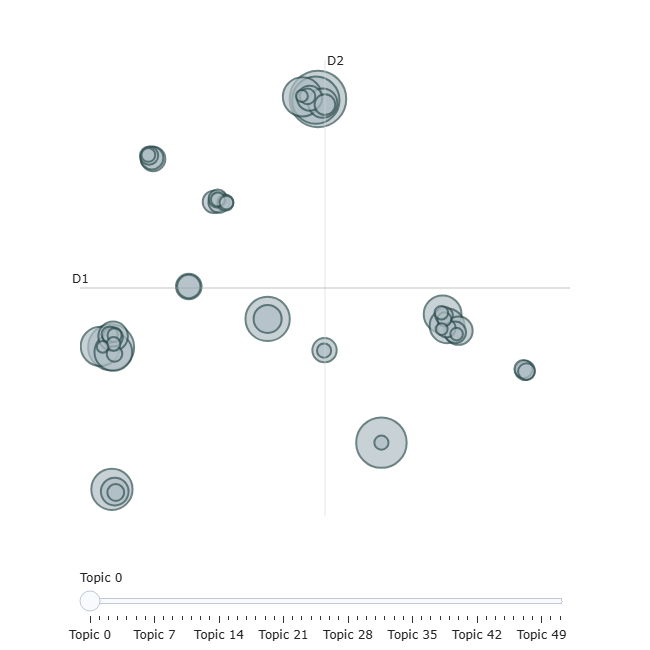}
 \caption{Optimized topics displayed in a reduced 2-dimensional embedding space, showing inter-topic distances.}
 \label{fig:2d-topic-representation}
\end{figure*}

\begin{figure*}[!ht]
 \centering
 \includegraphics[width=0.7\textwidth, height=0.92\textheight]{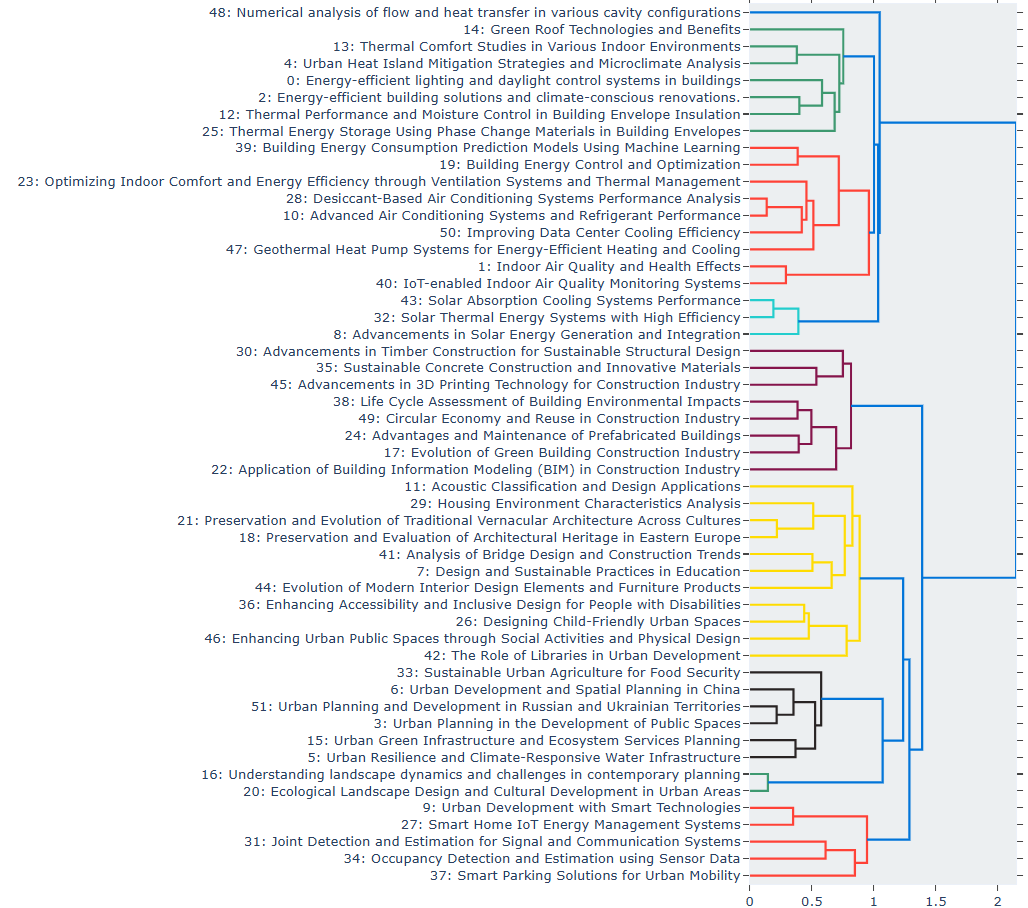}
 \caption{Dendrogram representation of the optimized topics' hierarchical clustering. The leaves of the tree represent the 52 clusters, the intermediate nodes represent merged clusters, and the height of the merging (distance from the leaves) indicate topic similarity as based on the cosine distance matrix between topic embeddings.}
 \label{fig:topicHierarchy}
\end{figure*}

\begin{table*}[!ht]
 \caption{The consolidated set of macro-topics resulting from topic merging, with their document counts, LLM-generated descriptions, and term based representation.}
 \label{tab:consolidatedTopics}
 \renewcommand{\arraystretch}{1.0} 
 { \footnotesize
\begin{tabularx}{\textwidth}{rrXX}
\toprule
Topic & Count & Representation & Terms \\
\midrule
0 & 39619 & Energy Efficiency and Thermal Comfort in Building Environments & ['energy', 'thermal', 'lighting', 'building', 'comfort', 'buildings', 'temperature', 'heat', 'study', 'performance'] \\
1 & 22740 & Indoor Air Quality and Sustainable Air Conditioning Systems & ['air', 'indoor', 'ventilation', 'conditioning', 'cooling', 'quality', 'temperature', 'energy', 'concentrations', 'control'] \\
2 & 14635 & Urban Development Strategies and Sustainable City Planning & ['urban', 'land', 'planning', 'development', 'city', 'spatial', 'cities', 'growth', 'public', 'expansion'] \\
3 & 10650 & Enhancing Child-Friendly Urban Spaces Through Design & ['design', 'space', 'product', 'architectural', 'public', 'architecture', 'students', 'library', 'research', 'study'] \\
4 & 10244 & Smart city development and urban data management & ['smart', 'city', 'parking', 'cities', 'data', 'occupancy', 'urban', 'information', 'development', 'based'] \\
5 & 9518 & Urban Resilience and Green Infrastructure in Climate Change Planning & ['urban', 'resilience', 'infrastructure', 'water', 'climate', 'green', 'planning', 'cities', 'flood', 'change'] \\
6 & 7452 & Architectural Integration of Solar Photovoltaic Systems in Buildings & ['solar', 'pv', 'energy', 'photovoltaic', 'bipv', 'power', 'systems', 'electricity', 'renewable', 'building'] \\
7 & 5632 & Preservation and Evolution of Traditional Architecture in Modern Contexts & ['architectural', 'architecture', 'heritage', 'traditional', 'house', 'historical', 'houses', 'buildings', 'study', 'cultural'] \\
8 & 5600 & Sustainable Building Construction and Design with Environmental Assessment & ['building', 'green', 'construction', 'buildings', 'assessment', 'life', 'environmental', 'cycle', 'industry', 'design'] \\
9 & 4619 & Landscape Planning and Design Theory & ['landscape', 'landscapes', 'design', 'garden', 'architecture', 'cultural', 'ecological', 'planning', 'rural', 'research'] \\
10 & 3368 & Urban Sound Environment Research in Architectural Design & ['noise', 'sound', 'acoustic', 'design', 'insulation', 'floor', 'building', 'level', 'environment', 'study'] \\
11 & 3039 & Sustainable Construction Materials and Technologies & ['concrete', 'timber', 'construction', 'wood', '3d', 'materials', 'material', 'structural', 'structures', 'building'] \\
12 & 1901 & Utilizing BIM in Construction and Building Information Modeling Industry & ['bim', 'construction', 'information', 'building', 'technology', 'design', 'industry', 'modeling', 'project', 'management'] \\
13 & 1090 & Urban Agriculture and Sustainable Food Systems & ['food', 'urban', 'city', 'cities', 'planning', 'land', 'production', 'community', 'social', 'development'] \\
14 & 767 & Sustainable Bridge Design and Construction & ['lt', 'gt', 'design', 'structural', 'construction', 'structures', 'structure', 'new', 'river', 'engineering'] \\
15 & 615 & Investigation of Cavity Dynamics and Heat Transfer in Various Flow Scenarios & ['cavity', 'flow', 'wall', 'transfer', 'heat', 'number', 'pressure', 'walls', 'numerical', 'surface'] \\
\bottomrule
\end{tabularx}
}
\end{table*}

\begin{figure*}[!ht]
 \centering
 \includegraphics[width=0.7\textwidth]{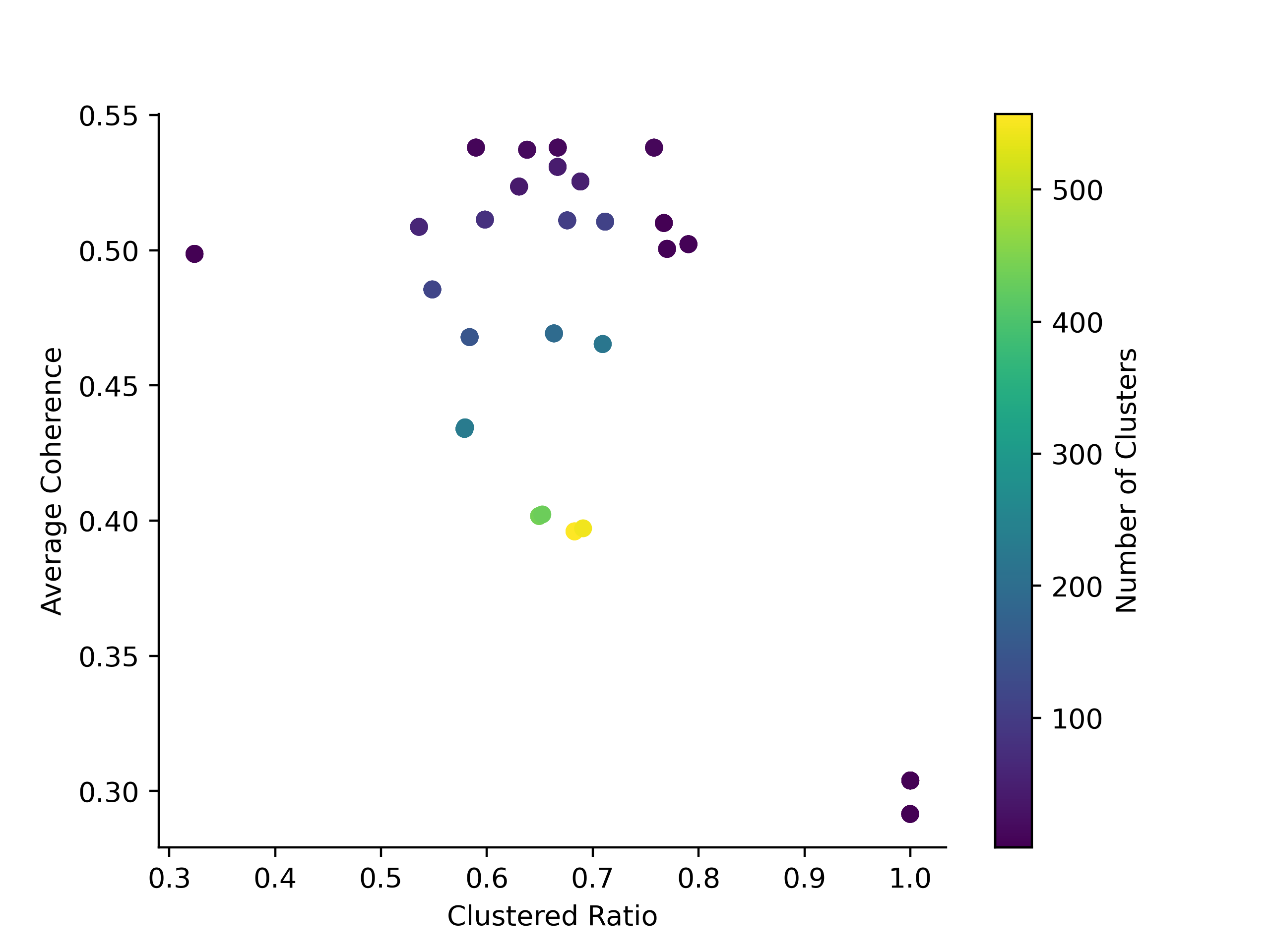}
 \caption{Average topic coherence values against the ratio of clustered datapoints for a subset of HDBSCAN hyperparameter settings. The color-coded values of the number of resulting clusters are also shown.}
 \label{fig:plotClusteredRatioCoherenceScore}
\end{figure*}

\clearpage\newpage
\section{CAUSALITY GRAPHS}
\label{sec:appendixCausalityGraphs}

\paragraph{Code Repositories:}
\url{https://github.com/zavavan/CRE_LLM_Benchmark}
\\
\textbf{Main Programming Languages}: Python
\\
\textbf{External Libraries}: Hugging Face's Transformers, PEFT, Datasets, Evaluate; SpaCy; Scikit-learn; Pandas

The full list of Prompt Variations of the Prompt Sensitivity Analysis is available at the URL: 
\url{https://drive.google.com/file/d/1mGawv1h-o-0xT6DNca4IAzVavmgrq8Ps/view?usp=sharing}
\\
\noindent\rule{\textwidth}{0.4pt}

Tables~\ref{tab:hyperparametersInference} and~\ref{tab:hyperparameters} provide the hyperparameters used across all zero-/few-shot and all instruction fine-tuning experiments, respectively. For any other unspecified parameter, we retained the default values provided by Hugging Face’s classes.

We train and run model inferences on Google Colab Pro using a single NVIDIA A100-SXM4 GPU with 80GB VRAM running Ubuntu 22.04.4 LTS with CUDA 12.4, with the exception of prompt-based methods using the \textit{DeepSeek-Qwen-Distill} model, which were run through the HF Inference API (\url{https://huggingface.co/docs/inference-providers/en/providers/hf-inference}).

The software environment used throughout this study included the following library versions: Python 3.12.12, PyTorch 2.8.0, Transformers 4.57.1, PEFT 0.17.1, Accelerate 1.11.0, and BitsAndBytes 0.48.1.

The released artifacts for the models fine-tuned using the PEFT LoRA framework contain only the LoRA adapter weight matrices and configuration files (stored in the files adapter\_model.bin  and adapter\_config.json, respectively). The base model weights are not redistributed; instead, users can reproduce the complete model by loading the corresponding base checkpoint from the Hugging Face Hub and attaching the adapter, using the code in Figure~\ref{fig:modelLoadingCode}:
\begin{figure}[htp!]
\centering
\begin{minipage}{.8\textwidth} 
\begin{footnotesize}
\begin{verbatim}
from transformers import AutoModelForCausalLM, AutoTokenizer
from peft import PeftModel
base_model_id = "johnsnowlabs/JSL-MedLlama-3-8B-v2.0"
adapter_path = "/path-to-fine-tuned-model"
bnb_config = BitsAndBytesConfig(
    load_in_4bit=True,
    bnb_4bit_quant_type="nf4",
    bnb_4bit_use_double_quant=True)
base_model = AutoModelForCausalLM.from_pretrained(
    base_model_id, 
    quantization_config=bnb_config)
tokenizer = AutoTokenizer.from_pretrained(base_model_id)
model = PeftModel.from_pretrained(base_model, adapter_path)

\end{verbatim}
\end{footnotesize}
 \end{minipage}
 \caption{Code snippet for merging the released LoRA adapter with the MedLlama base model.}
 \label{fig:modelLoadingCode}
\end{figure}

\begin{table}[htp!]
 \caption{\color{black}Model instantiation and inference parameters used across all zero-shot and few-shot experiments. \textit{max\_new\_tokens} parameter for two-step prompt methods \textit{SumAsk} and \textit{2-Chain} is specified as a pair of values, one for each inference call. \textit{max\_new\_tokens} is raised to 900 for inference with \textit{DeepSeek-Qwen-Distill} in all prompting methods, in order to accommodate for the long reasoning chains of this model.}
 \color{black}
\centering
{\footnotesize
 \begin{tabular}{llll}
 \toprule
\textbf{Parameter} & \textbf{HF parameter} & \textbf{Prompt} & \textbf{Value}\\
\hline
 \hline
Use sampling & \texttt{do\_sample} & all & False\\
4 bit quantization & \texttt{load\_in\_4bit} & all & True \\
quantization scheme & \texttt{bnb\_4bit\_quant\_type} & all & nf4 \\
 \hline
 \multirow{4}{*}{maximum num of new tokens} & 
 \multirow{4}{*}{max\_new\_tokens} & InstPrompt & 256 \\
& & iCL & 256\\
& & CoT & 512\\
& & SumAsk & 256,64\\
& & 2-Chain & 256,64\\
 \bottomrule
\end{tabular}
}
 \label{tab:hyperparametersInference}
\end{table}

\begin{table}[htp!]
 \caption{\color{black}LoRA configuration and training parameters used across all fine-tuning experiments.}
 \color{black}
\centering
{\footnotesize
 \begin{tabular}{lll}
 \toprule
 \textbf{Parameter} & \textbf{HF parameter} & \textbf{Value}\\
 \hline
 \hline
 LoRA rank & \texttt{r} & 8\\
 LoRA alpha & \texttt{lora\_alpha} & 32\\
 LoRA dropout & \texttt{lora\_dropout} & 0.05\\
 LoRA target modules & \texttt{target\_modules} & ["q\_proj", "v\_proj"]\\
\hline
 Validation Metric & \texttt{metric\_for\_best\_model}& eval\_loss\\
 Train batch size & \texttt{per\_device\_train\_batch\_size} & 8\\
 Validation batch size& \texttt{per\_device\_eval\_batch\_size} & 8\\
 Gradient Accumulation Steps & \texttt{gradient\_accumulation\_steps} & 4\\
Evaluation Strategy & \texttt{evaluation\_strategy} & steps\\
Evaluation Steps & \texttt{eval\_steps} & 10\\
Early Stopping Patience & \texttt{early\_stopping\_patience} & 3 \\
 Learning Rate& \texttt{learning\_rate} & 2e-4\\
 Optimizer & \texttt{optim} & paged\_adamw\_8bit \\
16-bit Precision Training & \texttt{fp16} & True\\
 
 \hline
\end{tabular}
}
 \label{tab:hyperparameters}
\end{table}

\ifdraft{
  \listoffixmes
}{}

\begin{figure}[!htbp]
\begin{tcolorbox}[width=\textwidth]
\begin{tcolorbox}[fontupper=\ttfamily\footnotesize,fonttitle=\ttfamily\small,colback=yellow!5!white,colframe=yellow!50!black,colbacktitle=yellow!75!black,title=\textbf{Input}]
Text:Bacterial Meningitis/ Fever :Again, the patient presented with obtundation and an inflammatory CSF after URI symptoms.She was admitted to the[**Hospital Unit Name 153**] for a short course and improved dramatically on Ceftriaxone, Vancomycin and Acyclovir.

Entities: E1:Fever , E2:Acyclovir

Gold Standard Label: 7

\end{tcolorbox}
\begin{tcolorbox}[fontupper=\ttfamily\footnotesize,fonttitle=\ttfamily\small,colback=yellow!5!white,colframe=darkgreen,colbacktitle=mediumgreen,title=\textbf{Model Response MistralOrca}]
The relation holding between Fever and Acyclovir is that Acyclovir is a medication used to treat Fever.The closest relation that best describes it is: Acyclovir prevents Fever.The label associated to this closest relation is 5. The answer is: LABEL 5.
\end{tcolorbox}

\begin{tcolorbox}[fontupper=\ttfamily\footnotesize,fonttitle=\ttfamily\small,colback=yellow!5!white,colframe=darkgreen,colbacktitle=mediumgreen,title=\textbf{Model Response DeepSeek-Qwen-Distill}]
Response CoT: Okay, so I need to figure out the relationship between `Fever' and `Acyclovir' in the given text. Let me read the text carefully.The text says: \textbf{[...]} So, the patient had fever and other symptoms, and was treated with three medications: Ceftriaxone, Vancomycin, and Acyclovir. The improvement was dramatic. Now, the entities are `Fever' and `Acyclovir'. I need to see how they relate. Looking at the options: 0: Fever causes Acyclovir. That doesn't make sense because Acyclovir is a medication, not a symptom or condition. [...] 5: Acyclovir prevents Fever. That could be possible. If the fever is a symptom of the underlying infection, then treating with Acyclovir might prevent further fever episodes or reduce fever. [...] 7: Acyclovir hinders Fever. This is similar to option 5. If Acyclovir is treating the cause of the fever, then it would hinder or reduce the fever. [...]. So, the patient had fever, and was treated with Acyclovir, which likely helped reduce the fever. So, Acyclovir hinders Fever. That would be option 7.Wait, but another way to look at it: the fever was a symptom, and the treatment (Acyclovir) was given, which helped the patient improve. So, the fever was a result of the infection, and the medication helped, so the fever went down. So, Acyclovir hinders Fever. Alternatively, maybe the fever was a symptom that indicated the need for Acyclovir, but that's more about enabling, which is option 2. But I'm not sure if that's the correct interpretation. Wait, the question is about the relation between Fever and Acyclovir. So, Fever is a symptom, and Acyclovir is a treatment. So, does Acyclovir hinder Fever? Yes, because it's used to treat the underlying cause, which would reduce the fever.So, the correct label would be 7.</think>The patient had fever as a symptom and was treated with Acyclovir, which likely helped reduce the fever. Therefore, Acyclovir hinders Fever.LABEL: 7
\end{tcolorbox}
\end{tcolorbox}

\caption{Comparing \textit{Mistral Orca} and \textit{DeepSeek-Qwen-Distill} model responses for a sample zero-shot CoT prompt. The target label to be extracted is 7 (\textit{E2 hinders E1}). Notice as \textit{DeepSeek-Qwen-Distill} evaluates, among others (not shown for simplicity), Label 5 option, but eventually opts for Label 7.} 
 \label{fig:errorAnalysis}
\end{figure}

\begin{figure*}[!ht]
\centering
\color{black}
\includegraphics[width=0.6\textwidth, keepaspectratio]{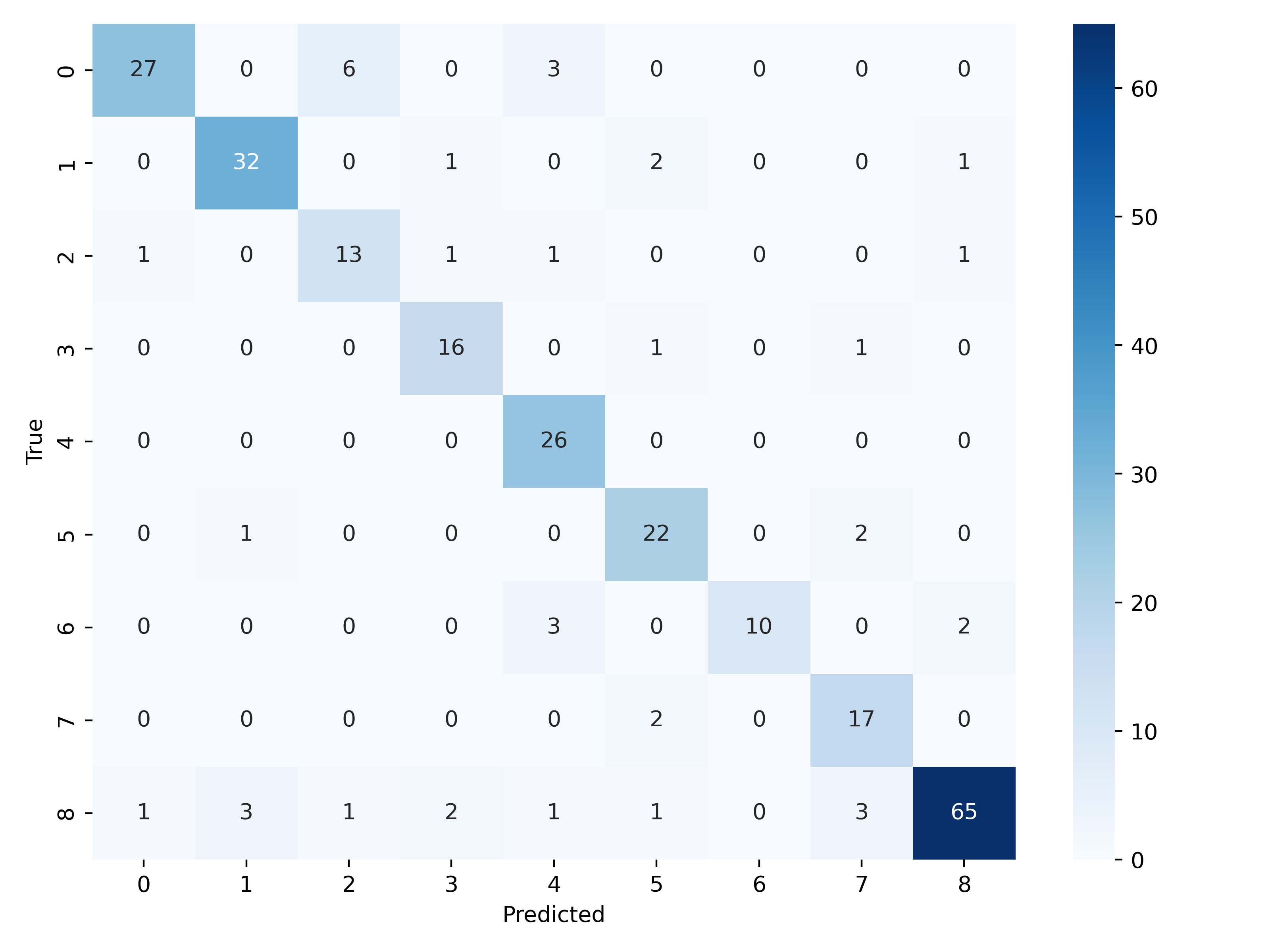}
\caption{Confusion matrix across the nine causal relation classes for the best-performing model, the instruction fine-tuned \textit{MedLlama}.}
\label{fig:confusionMatrix}
\end{figure*}

\end{document}